%% file: slmvar_arxiv_2.tex
\documentclass[11pt]{article}
\usepackage{natbib}
\usepackage{amsmath}
\usepackage{amsfonts}
\usepackage{amsopn}
\usepackage{ifthen}
\usepackage{algorithm}
\usepackage{algorithmic}
\usepackage{bbm}
\usepackage{bm}
\usepackage{graphicx}
\usepackage{url}

\newcommand{\figureres}{JPG}

\input{mymacros.tex} 
\input{mymacros2.tex}

\newtheorem{theorem}{Theorem}

\title{Large Scale Variational Inference and Experimental Design for
  Sparse Generalized Linear Models}

\author{Matthias W. Seeger\thanks{Saarland University and Max Planck Institute
        for Informatics, Campus E1.4, 66123 Saarbr\"{u}cken, Germany
        ({\tt mseeger@mmci.uni-saarland.de}).}
        \and Hannes Nickisch \thanks{Max Planck Institute for Biological
        Cybernetics, Spemannstra{\ss}e 38, 72076 T\"{u}bingen, Germany
        ({\tt hn@tuebingen.mpg.de}).}}

\begin{document}

\maketitle


\begin{abstract}%
Many problems of low-level computer vision and image processing, such as
denoising, deconvolution, tomographic reconstruction or super-resolution,
can be addressed by maximizing the posterior distribution of a
sparse linear model (SLM). We show how higher-order Bayesian decision-making
problems, such as optimizing image acquisition in magnetic resonance scanners,
can be addressed by querying the SLM posterior covariance, unrelated to the
density's mode. We propose a scalable algorithmic framework, with which SLM
posteriors over full, high-resolution images can be approximated for the first
time, solving a variational optimization problem which is convex iff
posterior mode finding is convex. These methods successfully drive the
optimization of sampling trajectories for real-world magnetic resonance imaging
through Bayesian experimental design, which has not been attempted before.
Our methodology provides new insight into similarities and differences
between sparse reconstruction and approximate Bayesian inference, and has
important implications for compressive sensing of real-world images. Parts of
this work appeared at conferences \cite{Seeger:08a,Nickisch:09}.
\end{abstract}

\pagestyle{plain}
\markboth{M.~Seeger and H.~Nickisch}{Large Scale Sparse Bayesian Inference}

\renewcommand{\labelenumi}{(\arabic{enumi})}

\section{Introduction}
\label{sec:intro}

Natural images have a sparse low-level statistical signature, represented in
the prior distribution of a sparse linear model (SLM). Imaging problems such
as reconstruction, denoising or deconvolution can successfully be solved
by maximizing its posterior density (maximum {\it a posteriori} (MAP)
estimation), a convex program for certain SLMs, for which efficient solvers
are available. The success of these techniques in modern imaging
practice has somewhat shrouded their limited scope as Bayesian techniques: of
all the information in the posterior distribution, they make use of the
density's mode only.

Consider the problem of {\em optimizing image acquisition}, our major
motivation in this work. Magnetic resonance images are reconstructed from
Fourier samples. With scan time proportional to the number of samples, a central
question to ask is which sampling designs of minimum size still lead to MAP
reconstructions of diagnostically useful image quality? This is not a direct
reconstruction problem, the focus is on improving measurement designs to
better support subsequent reconstruction. Goal-directed acquisition
optimization cannot sensibly be addressed by MAP estimation, yet we show
how to successfully drive it by Bayesian posterior information beyond the
mode. Advanced decision-making of this kind needs uncertainty quantification
(posterior covariance) rather than point estimation, requiring us to step out
of the {\em sparse reconstruction} scenario and approximate {\em sparse
Bayesian inference} instead.

The Bayesian inference relaxation we focus on is not new \cite{Girolami:01,
Palmer:06,Jaakkola:97a}, yet when it comes to problem characterization or
efficient algorithms, previous inference work lags far behind standards
established for MAP reconstruction. Our contributions range from theoretical
characterizations over novel scalable solvers to applications not previously
attempted. The inference relaxation is shown to be a convex optimization
problem if and only if this holds for MAP estimation (\secref{inference}), a
property not previously established for this or any other SLM inference
approximation.
Moreover, we develop novel {\em scalable} double loop algorithms to solve the
variational problem orders of magnitude faster than previous methods we are
aware of (\secref{algorithm}). These algorithms expose an important link
between variational Bayesian inference and sparse MAP reconstruction, reducing
the former to calling variants of the latter few times, interleaved by Gaussian
covariance (or PCA) approximations (\secref{estim-vars}). By way of this
reduction, the massive recent interest in MAP estimation can play a role for
variational Bayesian inference just as well. To complement these similarities
and clarify confusion in the literature, we discuss computational and
statistical differences of sparse estimation and Bayesian inference in detail
(\secref{map-estim}).

The ultimate motivation for novel developments presented here is sequential
Bayesian experimental design (\secref{design}), applied to acquisition
optimization for medical imaging. We present a powerful variant of adaptive
compressive sensing, which succeeds on real-world image data where theoretical
proposals for non-adaptive compressive sensing \cite{Donoho:06,Candes:06,
Donoho:03} fail (\secref{compsens}). Among our experimental results is part of
a first successful study for Bayesian sampling optimization of magnetic resonance
imaging, learned and evaluated on real-world image data
(\secref{exp-mri}).

An implementation of the algorithms presented here is publicly available, as part
of the {\tt glm-ie} toolbox (\secref{glm-ie}). It can be downloaded from
\url{mloss.org/software/view/269/}.

\section{Sparse Bayesian Inference. Variational Approximations}
\label{sec:model}

In a sparse linear model (SLM), the image $\vu{}\in \R^n$ of $n$ pixels is
unknown,
and $m$ linear measurements $\vy{}\in \R^m$ are given, where $m\ll n$ in
many situations of practical interest.
\begin{equation}\label{eq:lin-likehood}
  \vy{} = \mxx{} \vu{} + \veps{},\quad \veps{}\sim N(\vzero,\sigma^2 \Id),
\end{equation}
where $\mxx{}\in \R^{m\times n}$ is the {\em design matrix}, and $\veps{}$
is Gaussian noise of variance $\sigma^2$, implying the Gaussian likelihood
$P(\vy{}|\vu{})=N(\vy{}|\mxx{}\vu{},\sigma^2\Id)$. Natural images are
characterized by histograms of simple filter responses (derivatives,
wavelet coefficients) exhibiting {\em super-Gaussian} (or {\em sparse})
form: most coefficients are close to zero, while a small fraction have
significant sizes \cite{Simoncelli:99,Seeger:07d} (a precise definition of
super-Gaussianity is given in \secref{var-bounds}). Accordingly, SLMs have
super-Gaussian prior distributions $P(\vu{})$. The MAP estimator
$\hat{\vu{}}_{\text{MAP}} := \argmax_{\vu{}}[\log P(\vy{}|\vu{}) +
\log P(\vu{})]$ can outperform maximum likelihood
$\hat{\vu{}}_{\text{ML}} := \argmax_{\vu{}} \log P(\vy{}|\vu{})$, when $\vu{}$
represents an image.

In this paper, we focus on priors of the form $P(\vu{})\propto
\prod_{i=1}^q t_i(s_i)$, where $\vs{}=\mxb{}\vu{}$. The potential functions
$t_i(\cdot)$ are positive and bounded. The operator $\mxb{}\in\R^{q\times n}$
may contain derivative filters or a wavelet transform. Both $\mxx{}$ and
$\mxb{}$ have to be structured or
sparse, in order for any SLM algorithm to be scalable. {\em Laplace} (or
double exponential) potentials are sparsity-enforcing:
\begin{equation}\label{eq:laplace}
  t_i(s_i) = e^{-\tau_i|s_i|},\quad \tau_i>0.
\end{equation}
For this particular prior and the Gaussian likelihood \eqp{lin-likehood},
MAP estimation corresponds to a quadratic program, known as {\em LASSO}
\cite{Tibshirani:96} for $\mxb{}=\Id$. Note that $\log t_i(s_i)$ is concave.
In general, if {\em log-concavity} holds for all model potentials, MAP
estimation is a convex problem.
Another example for sparsity potentials are {\em Student's t}:
\begin{equation}\label{eq:student-t}
  t_i(s_i) =  (1 + (\tau_i/\nu) s_i^2)^{-(\nu+1)/2},\quad
  \tau_i,\nu > 0.
\end{equation}
For these, $\log t_i(s_i)$ is not concave, and MAP estimation is not (in
general) a convex program. Note that $-\log t_i(s_i)$ is also known
as Lorentzian penalty function.


\subsection{Variational Lower Bounds}
\label{sec:var-bounds}

Bayesian inference amounts to computing moments of the posterior distribution
\begin{equation}\label{eq:posterior}
\begin{split}
  P(\vu{}|\vy{}) & = Z^{-1} N(\vy{}|\mxx{}\vu{},\sigma^2\Id)
  \prod\nolimits_{i=1}^q t_i(s_i),\quad \vs{}=\mxb{}\vu{}, \\
  Z & = \int N(\vy{}|\mxx{}\vu{},\sigma^2\Id) \prod\nolimits_{i=1}^q t_i(s_i)\,
  d\vu{}.
\end{split}
\end{equation}
This is not analytically tractable in general for sparse linear models, due to
two reasons coming together: $P(\vu{}|\vy{})$ is highly coupled ($\mxx{}$ is
not blockdiagonal) and non-Gaussian. We focus on {\em variational}
approximations here, rooted in statistical physics. The {\em log partition
function} $\log Z$ (also known as log evidence or log marginal likelihood)
is the prime target for variational methods \cite{Wainwright:08}. Formally,
the potentials $t_i(s_i)$ are replaced by Gaussian terms of parametrized width,
the posterior $P(\vu{}|\vy{})$ by a Gaussian approximation $Q(\vu{}|\vy{})$.
The width parameters are adjusted by fitting $Q(\vu{}|\vy{})$ to
$P(\vu{}|\vy{})$, in what amounts to the variational optimization problem.

\begin{figure}[ht!]
\begin{center}
  \includegraphics[width=\columnwidth]{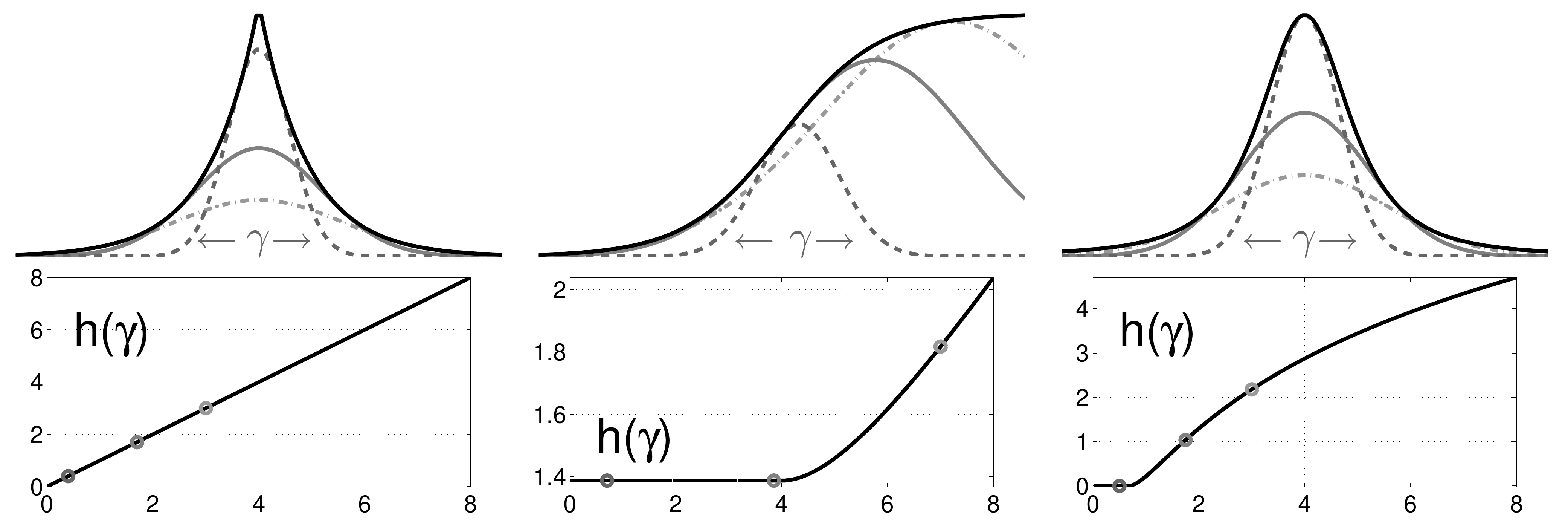}
\end{center}
\caption{\label{fig:super-gauss}
  Super-Gaussian distributions admit tight Gaussian-form lower
  bounds of any width $\gamma$. Left: Laplace \eqp{laplace}; middle:
  Bernoulli \eqp{bernoulli}; right: Student's t \eqp{student-t}.}
\end{figure}

Our variational relaxation exploits the fact that all potentials $t_i(s_i)$ are
{\em (strongly) super-Gaussian}: there exists a $b_i\in\R$ such that
$\tilde{g}_i(s_i) := \log t_i(s_i) - b_i s_i$ is even, and convex and
decreasing
as a function of $x_i := s_i^2$ \cite{Palmer:06}. We write
$g_i(x_i) := \tilde{g}_i(x_i^{1/2})$, $x_i\ge 0$ in the sequel. This implies
that
\begin{equation}\label{eq:super-gauss}
  t_i(s_i) = \max_{\gamma_i\ge 0} e^{b_i s_i - s_i^2/(2\gamma_i) -h_i(\gamma_i)/2},
  \quad h_i(\gamma_i) := \max_{x_i\ge 0} (-x_i/\gamma_i - 2 g_i(x_i)).
\end{equation}
A super-Gaussian $t_i(s_i)$ has tight Gaussian-form lower bounds of all
widths (see \figref{super-gauss}). We replace $t_i(s_i)$ by these
lower bounds in order to step from $P(\vu{}|\vy{})$ to the family of
approximations $Q(\vu{}|\vy{})\propto P(\vy{}|\vu{}) e^{\vb{}^T\vs{} -
\frac{1}2 \vs{}^T(\diag\vgamma{})^{-1}\vs{}}$, where $\vgamma{} = (\gamma_i)$.

To establish \eqp{super-gauss}, note that the extended-value function
$g_i(x_i)$ (assigning $g_i(x_i)=\infty$ for $x_i<0$) is a closed proper convex
function, thus can be represented by Fenchel duality
\cite[Sect.~12]{Rockafellar:70}:
$g_i(x_i) = \sup_{\lambda_i} x_i\lambda_i - g_i^*(\lambda_i)$, where the
conjugate function $g_i^*(\lambda_i) = \sup_{x_i} x_i\lambda_i - g_i(x_i)$ is
closed convex as well. For $x_i\ge 0$ and $\lambda_i>0$, we have that
$x_i\lambda_i - g_i(x_i) \ge x_i\lambda_i - g_i(0)\to \infty$ with $x_i\to
\infty$, which implies that $g_i^*(\lambda_i)=\infty$ for $\lambda_i>0$.
Also, $g_i^*(0) = -\lim_{x_i\to\infty} g_i(x_i)$, so that $-g_i^*(0) < g_i(x_i)$
for any $x_i<\infty$. Therefore, for any $x_i\in [0,\infty)$, $g_i(x_i) =
\sup_{\lambda_i<0} x_i\lambda_i - g_i^*(\lambda_i)$. Reparameterizing
$\gamma_i := -1/(2\lambda_i)$, we have that $g_i(x_i) = \max_{\gamma_i\ge 0}
-x_i/(2\gamma_i) - g_i^*(-1/(2\gamma_i))$ (note that in order to accommodate
$g_i(0)$ in general, we have to allow for $\gamma_i=0$). Finally,
$h_i(\gamma_i) := 2 g_i^*(-1/(2\gamma_i))$.

The class of super-Gaussian potentials is large. All scale mixtures
(mixtures of zero-mean Gaussians $t_i(s_i) =
\Ex_{\gamma_i}[N(s_i | 0,\gamma_i)]$) are super-Gaussian, and
$h_i(\gamma_i)$ can be written in terms of the density for $\gamma_i$
\cite{Palmer:06}. Both Laplace \eqp{laplace} and Student's t potentials
\eqp{student-t} are super-Gaussian, with $h_i(\gamma_i)$ given in
\appref{impl-details}.
Bernoulli potentials, used as binary classification likelihoods,
\begin{equation}\label{eq:bernoulli}
  t_i(s_i) = \left( 1 + e^{-y_i \tau_i s_i} \right)^{-1},\quad
  y_i\in \{\pm 1\},\; \tau_i>0
\end{equation}
are super-Gaussian, with $b_i=y_i\tau_i/2$
\cite[Sect.~3.B]{Jaakkola:97a}. While the corresponding $h_i(\gamma_i)$
cannot be determined analytically, this is not required in our algorithms,
which can be implemented based on $g_i(x_i)$ and its derivatives only.
Finally, if the $t_i^{(l)}(s_i)$ are super-Gaussian, so is the positive mixture
$\sum_{l=1}^L \alpha_l t_i^{(l)}(s_i)$, $\alpha_l>0$,
because the {\em logsumexp} function $\vx{}\mapsto \log\vone^T\exp(\vx{})$
\cite[Sect.~3.1.5]{Boyd:02} is strictly convex on $\R^L$ and increasing in each
argument, and the log-mixture is the concatenation of {\em logsumexp}
with $(\log t_i^{(l)}(s_i) + \log\alpha_l)_l$, the latter convex and
decreasing for $x_i=s_i^2>0$ in each component \cite[Sect.~3.2.4]{Boyd:02}.

A natural criterion for fitting $Q(\vu{}|\vy{})$ to $P(\vu{}|\vy{})$ is
obtained by plugging these bounds into the partition function $Z$ of
\eqp{posterior}:
\begin{equation}\label{eq:lower-first}
  Z \ge e^{-h(\vgamma{})/2}\int N(\vy{}|\mxx{}\vu{},\sigma^2\Id)
  e^{\vb{}^T\vs{} - \frac{1}2 \vs{}^T\mxgamma{}^{-1}\vs{}}\, d\vu{},\;
  h(\vgamma{}) := \sum\nolimits_{i=1}^q h_i(\gamma_i),
\end{equation}
where $\mxgamma{} := \diag\vgamma{}$ and $\vs{}=\mxb{}\vu{}$.
The right hand side is a Gaussian integral and can be evaluated easily.
The variational problem is to maximize this lower bound w.r.t.\ the variational
parameters $\vgamma{}\succeq\vzero$ ($\gamma_i\ge 0$ for all $i=\rng{q}$), with
the aim of tightening the approximation to $\log Z$. This criterion can
be interpreted as a divergence function: if the family of all $Q(\vu{}|\vy{})$
contained the true posterior $P(\vu{}|\vy{})$, the latter would maximize the
bound.

This relaxation has frequently been used before
\cite{Girolami:01,Palmer:06,Jaakkola:97a} on inference problems of moderate
size. In the following, we provide results that extend its scope to large
scale imaging problems of interest here. In the next section, we characterize
the convexity of the underlying optimization problem precisely. In
\secref{algorithm}, we provide a new class of algorithms for solving this
problem orders of magnitude faster than previously used techniques.

\section{Convexity Properties of Variational Inference}
\label{sec:inference}

In this section, we characterize the variational optimization problem of
maximizing the right hand side of \eqp{lower-first}. We show that it is
a convex problem if and only if all potentials $t_i(s_i)$ are
log-concave, which is equivalent to MAP estimation being convex for
the same model.

We start by converting the lower bound \eqp{lower-first} into a more amenable
form. The Gaussian posterior $Q(\vu{}|\vy{})$ has the covariance matrix
\begin{equation}\label{eq:gauss-post}
  \Cov_Q[\vu{}|\vy{}] = \mxa{}^{-1},\quad \mxa{} := \sigma^{-2}\mxx{}^T
  \mxx{} + \mxb{}^T\mxgamma{}^{-1}\mxb{},\; \mxgamma{} = \diag\vgamma{}.
\end{equation}
We have that
\[
  \int P(\vy{} | \vu{}) e^{\vb{}^T\vs{} - \frac{1}2 \vs{}^T\mxgamma{}^{-1}\vs{}}\,
  d\vu{} = |2\pi\mxa{}^{-1}|^{1/2}\max_{\vu{}} P(\vy{} | \vu{})
  e^{\vb{}^T\vs{} - \frac{1}2 \vs{}^T\mxgamma{}^{-1}\vs{}},\quad \vs{}=\mxb{}\vu{},
\]
since $\max_{\vu{}}Q(\vu{}|\vy{}) = |2\pi\mxa{}^{-1}|^{-1/2}
\int Q(\vu{}|\vy{})\, d\vu{}$. We find that
$Z\ge C_1 e^{-\phi(\vgamma{})/2}$, where
\begin{equation}\label{eq:phi-crit}
  \phi(\vgamma{}) := \log|\mxa{}| + h(\vgamma{}) + \min_{\vu{}}
  R(\vu{},\vgamma{}),\quad R := \sigma^{-2}\|\vy{} - \mxx{}\vu{}\|^2
  + \vs{}^T\mxgamma{}^{-1}\vs{} - 2\vb{}^T\vs{}
\end{equation}
and $C_1 = (2\pi)^{(n-m)/2}\sigma^{-m}$. The variational problem is
$\min_{\vgamma{}\succeq\vzero}\phi(\vgamma{})$, and the Gaussian posterior
approximation is $Q(\vu{} | \vy{})$ with the final parameters $\vgamma{}$
plugged in. We will also write $\phi(\vu{},\vgamma{}) := \log|\mxa{}|
+ h(\vgamma{}) + R(\vu{},\vgamma{})$, so that $\phi(\vgamma{}) =
\min_{\vu{}}\phi(\vu{},\vgamma{})$.

It is instructive to compare this variational inference problem with
maximum a posteriori (MAP) estimation:
\begin{equation}\label{eq:map-estim}
\begin{split}
  \min_{\vu{}} -2\log P(\vu{}|\vy{}) & = \min_{\vu{}}
  \sigma^{-2}\|\vy{} - \mxx{}\vu{}\|^2 -2\sum\nolimits_i \log t_i(s_i) + C_2 \\
  & = \min_{\vu{},\vgamma{}\succeq\vzero} h(\vgamma{}) + R(\vu{},\vgamma{})
  + C_2,
\end{split}
\end{equation}
where $C_2$ is a constant. The difference between these problems
rests on the $\log|\mxa{}|$ term, present in $\phi(\vgamma{})$ yet absent
in MAP. Ultimately, this observation is the key to the characterization
provided in this section and to the scalable solvers developed in the
subsequent section. Its full relevance will be clarified in \secref{map-estim}.

\subsection{Convexity Results}
\label{sec:inf-convex-res}

In this section, we prove that $\phi(\vgamma{})$ is convex if all potentials
$t_i(s_i)$ are log-concave, with this condition being necessary in general.
We address each term in \eqp{phi-crit} separately.

\begin{theorem}\label{th:ldeta-convex}
Let $\mxx{}\in \R^{m\times n}$, $\mxb{}\in \R^{q\times n}$ be arbitrary
matrices, and
\[
  \tilde{\mxa{}}(\vd{}) := \sigma^{-2}\mxx{}^T\mxx{} +
  \mxb{}^T(\diag\vd{})\mxb{},
  \quad \vd{}\succ\vzero,
\]
so that $\tilde{\mxa{}}(\vd{})$ is positive definite for all
$\vd{}\succ\vzero$.
\begin{enumerate}
\item\label{th1-1}
  Let $f_i(\gamma_i)$ be twice continuously differentiable
  functions into $\R_+$, so that $\log f_i(\gamma_i)$ are convex for all $i$
  and $\gamma_i$. Then, $\vgamma{}\mapsto \log|\tilde{\mxa{}}(f(\vgamma{}))|$
  is convex. Especially, $\vgamma{}\mapsto \log|\mxa{}|$ is convex.
\item\label{th1-2}
  Let $f_i(\pi_i)$ be concave functions into $\R_+$.
  Then, $\vpi{}\mapsto \log|\tilde{\mxa{}}(f(\vpi{}))|$ is concave.
  Especially, $\vgamma{}^{-1}\mapsto \log|\mxa{}|$ is concave.
\item\label{th1-3}
  Let $f_i(\gamma_i)$ be concave functions into $\R_+$.
  Then, $\vgamma{}\mapsto \vone^T(\log f(\vgamma{}))
  + \log|\tilde{\mxa{}}(f(\vgamma{})^{-1})|$ is concave. Especially,
  $\vgamma{}\mapsto \vone^T(\log\vgamma{}) + \log|\mxa{}|$ is concave.
\item\label{th1-4}
  Let $Q(\vu{}|\vy{})$ be the approximate posterior with
  covariance matrix given by \eqp{gauss-post}. Then, for all $i$:
  \[
    \Var_Q[s_i|\vy{}] = \vdelta{i}^T\mxb{}\mxa{}^{-1}\mxb{}^T
    \vdelta{i}\le \gamma_i.
  \]
\end{enumerate}
\end{theorem}

A proof is provided in \appref{ldeta-convex}. Instantiating part (\ref{th1-1})
with
$f_i(\gamma_i) = \gamma_i^{-1}$, we see that $\vgamma{}\mapsto\log|\mxa{}|$ is
convex. Other valid examples are $f_i(\gamma_i) = \gamma_i^{-\beta_i}$,
$\beta_i>0$. For $f_i(\gamma_i) = e^{\gamma_i}$, we obtain the convexity of
$\vgamma{}\mapsto \log|\tilde{\mxa{}}(\exp(\vgamma{}))|$, generalizing the
{\em logsumexp} function to matrix values. Part (\ref{th1-2}) and part
(\ref{th1-3}) will be required in \secref{ldeta-bounds}. Finally,
part (\ref{th1-4}) gives a precise characterization of $\gamma_i$ as
sparsity parameter, regulating the variance of $s_i$.

\begin{theorem}\label{th:crit-convex}
The function
\[
  \vgamma{}\mapsto \phi(\vgamma{}) - h(\vgamma{}) = \log|\mxa{}| +
  \min_{\vu{}} \left( \sigma^{-2}\|\vy{} - \mxx{}\vu{}\|^2
  + \vs{}^T\mxgamma{}^{-1}\vs{} - 2\vb{}^T\vs{} \right)
\]
is convex for $\vgamma{}\succ\vzero$, where $\vs{}=\mxb{}\vu{}$.
\end{theorem}
{\em Proof}. The convexity of $\log|\mxa{}|$ has been shown in
\thref{ldeta-convex}(\ref{th1-1}).
$\sigma^{-2}\|\vy{} - \mxx{}\vu{}\|^2 - 2\vb{}^T\vs{}$ is convex in $\vu{}$,
and $(\vu{},\vgamma{})\mapsto \vs{}^T\mxgamma{}^{-1}\vs{}$ is {\em jointly}
convex, since the quadratic-over-linear function
$(s_i,\gamma_i)\mapsto s_i^2/\gamma_i$ is jointly convex for $\gamma_i>0$
\cite[Sect.~3.1.5]{Boyd:02}. Therefore, $\min_{\vu{}} R(\vu{},\vgamma{})$
is convex for $\vgamma{}\succ\vzero$
\cite[Sect.~3.2.5]{Boyd:02}. This completes the proof.

To put this result into context, note that
\[
  \phi(\vgamma{}) - h(\vgamma{}) = -2\log\int P(\vy{}|\vu{}) e^{\vb{}^T\vs{}
  - \frac{1}2 \vs{}^T\mxgamma{}^{-1}\vs{}}\, d\vu{},\quad \vs{}=\mxb{}\vu{}
\]
is the {\em negative} log partition function of a Gaussian with natural
parameters $\vgamma{}^{-1}$: it is well known that $\vgamma{}^{-1}\mapsto
\phi(\vgamma{}) - h(\vgamma{})$ is a {\em concave} function
\cite{Wainwright:08}. However, $\vgamma{}^{-1}\mapsto h(\vgamma{})$ is
{\em convex} for a model with super-Gaussian potentials (recall that
$h_i(\gamma_i) = 2 g_i^*(-1/(2\gamma_i))$, where $g_i^*(\cdot)$ is convex as
dual function of $g_i(\cdot)$), which means that in general $\vgamma{}^{-1}
\mapsto \phi(\gamma)$ need not be convex or concave. The {\em convexity} of
this negative log partition function w.r.t.\ $\vgamma{}$ seems specific to the
Gaussian case.

Given \thref{crit-convex}, if all $h_i(\gamma_i)$ are convex, the whole
variational problem $\min_{\vgamma{}\succeq\vzero}\phi$ is convex. With the
following theorem, we characterize this case precisely.

\begin{theorem}\label{th:convex-inf}
Consider a model with Gaussian likelihood \eqp{lin-likehood} and a prior
$P(\vu{})\propto \prod_{i=1}^q t_i(s_i)$, $\vs{}=\mxb{}\vu{}$, so that all
$t_i(s_i)$ are strongly super-Gaussian, meaning that $\tilde{g}_i(s_i) =
\log t_i(s_i) - b_i s_i$ is even, and $g_i(x_i) = \tilde{g}_i(x_i^{1/2})$ is
strictly convex and decreasing for $x_i>0$.
\begin{enumerate}
\item\label{th2-1}
  If $\tilde{g}_i(s_i)$ is concave and twice continuously
  differentiable for $s_i>0$, then $h_i(\gamma_i)$ is convex. On the
  other hand, if $\tilde{g}_i''(s_i)>0$ for some $s_i>0$, then $h_i(\gamma_i)$
  is not convex at some $\gamma_i>0$.
\item\label{th2-2}
  If all $\tilde{g}_i(s_i)$ are concave and twice continuously
  differentiable for $s_i>0$, then the variational problem
  $\min_{\vgamma{}\succeq\vzero}\phi$ is a convex optimization problem. On the
  other hand, if $\tilde{g}_i''(s_i)>0$ for some $i$ and $s_i>0$, then
  $h(\vgamma{})$ is not convex, and there exist some $\mxx{}$, $\mxb{}$, and
  $\vy{}$ such that $\phi(\vgamma{})$ is not a convex function.
\end{enumerate}
\end{theorem}
The proof is given in \appref{convex-inf}. Our theorem provides a complete
characterization of convexity for the variational inference relaxation
of \secref{model}, which is {\em the same} as for MAP estimation.
Log-concavity of all potentials is sufficient, and necessary in general, for
the
convexity of either. We are not aware of a comparable equivalence having been
established for any other nontrivial approximate inference method for
continuous variable models.

%
We close this section with some examples. For Laplacians \eqp{laplace},
$h_i(\gamma_i) = \tau_i^2 \gamma_i$ (see \appref{impl-details}). For SLMs with
these potentials,
MAP estimation is a convex quadratic program. Our result implies that
variational inference is a convex problem as well, albeit with a differentiable
criterion. Bernoulli potentials \eqp{bernoulli} are log-concave. MAP estimation
for generalized linear models with these potentials is
known as penalized logistic regression, a convex problem typically solved by the
iteratively reweighted least squares (IRLS) algorithm. Variational inference
for this model is also a convex problem, and our algorithms introduced in
\secref{algorithm} make use of IRLS as well.
Finally, Student's t potentials \eqp{student-t} are not log-concave, and
$h_i(\gamma_i)$ is neither convex nor concave (see \appref{impl-details}).
Neither MAP estimation nor variational inference are convex in this case.

Convexity of an algorithm is desirable for many reasons. No restarting is
needed to avoid local minima. Typically, the result is robust to small
perturbations of the data. These stability properties become all the more
important in the context of sequential experimental design (see
\secref{design}), or when Bayesian model selection\footnote{
  Model selection (or hyperparameter learning) is not discussed in this paper.
  It can be implemented easily by maximizing the lower bound
  $-\phi(\vgamma{})/2 + \log C_1\le \log Z$ w.r.t.\ hyperparameters.}
is used. However, the convexity of $\phi(\vgamma{})$ does not
necessarily imply that the minimum point can be found efficiently. In the next
section, we propose a class of algorithms that solve the variational problem
for very large instances, by decoupling the criterion \eqp{phi-crit} in a novel
way.

\section{Scalable Inference Algorithms}\label{sec:algorithm}

In this section, we propose novel algorithms for solving the variational
inference problem $\min_{\vgamma{}}\phi$ in a scalable way. Our algorithms
can be used whether $\phi(\vgamma{})$ is convex or not, they are guaranteed
to converge to a stationary point. All efforts are reduced to well known,
scalable algorithms of signal reconstruction and numerical mathematics, with
little extra technology required, and no additional heuristics or step size
parameters to be tuned.

We begin with the special case of log-concave potentials $t_i(s_i)$, such
as Laplace \eqp{laplace} or Bernoulli \eqp{bernoulli}, extending our framework
to full generality in \secref{algo-general}. The
variational inference problem is convex in this case (\thref{convex-inf}).
Previous algorithms for solving $\min_{\vgamma{}}\phi(\vgamma{})$
\cite{Girolami:01,Palmer:06} are of the {\em coordinate descent} type,
minimizing $\phi$ w.r.t.\ one $\gamma_i$ at a time. Unfortunately, such
algorithms cannot be scaled up to imaging problems of interest here. An
update of $\gamma_i$ depends on the marginal posterior $Q(s_i|\vy{})$,
whose computation requires the solution of a linear system with matrix
$\mxa{}\in\R^{n\times n}$. At the projected scale, neither $\mxa{}$ nor a
decomposition thereof can be maintained, systems have to be solved iteratively.
Now, each of the $q$ potentials has to be visited at least once, typically
several times. With $q$, $n$, and $m$ in the hundred thousands, it is
certainly infeasible to solve $O(q)$ linear systems. In contrast, the
algorithms we develop here often converge after less than hundred systems have
been solved. We could also feed $\phi(\vgamma{})$ and its gradient
$\nabla_{\vgamma{}}\phi$ into an off-the-shelf gradient-based optimizer. However,
as already noted in \secref{inference}, $\phi(\vgamma{})$ is the sum of a
standard penalized least squares (MAP) part and a highly coupled,
computationally difficult term. The algorithms we propose take account of
this decomposition, decoupling the troublesome term in inner loop standard
form problems which can be solved by any of a large number of specialized
algorithms not applicable to $\min_{\vgamma{}}\phi(\vgamma{})$. The expensive
part of $\nabla_{\vgamma{}}\phi$ has to be computed only a few times for our
algorithms to converge.

We make use of a powerful idea known as {\em double loop} or
{\em concave-convex} algorithms. Special cases of such algorithms are
frequently used in machine learning, computer vision, and statistics: the
expectation-maximization (EM) algorithm \cite{Dempster:77}, variational mean
field Bayesian inference \cite{Attias:99}, or CCCP for discrete approximate
inference \cite{Yuille:03}, among many others. The idea is to tangentially
upper bound $\phi$ by decoupled functions $\phi_{\vz{}}$ which are much
simpler to minimize than $\phi$ itself: algorithms iterate between refitting
$\phi_{\vz{}}$ to $\phi$ and minimizing $\phi_{\vz{}}$. For example, in the EM
algorithm for maximizing a log marginal likelihood, these stages correspond to
``E step'' and ``M step'': while the criterion could well be minimized
directly (at the expense of one ``E step'' per criterion evaluation),
``M step'' minimizations are much easier to do.

As noted in \secref{inference}, if the variational criterion \eqp{phi-crit}
lacked the $\log|\mxa{}|$ part, it would correspond to a penalized least
squares MAP objective \eqp{map-estim}, and simple efficient algorithms would
apply. As discussed in \secref{estim-vars}, evaluating
$\log|\mxa{}|$ or its gradient are computationally challenging. Crucially,
this term satisfies a concavity property. As shown in \secref{ldeta-bounds},
Fenchel duality implies that $\log|\mxa{}| \le
\vz{1}^T(\vgamma{}^{-1}) - g_1^*(\vz{1})$. For any fixed $\vz{1}\succ\vzero$,
the upper bound is tangentially tight, convex in $\vgamma{}$, and decouples
additively. If $\log|\mxa{}|$ is replaced by this upper bound, the resulting
objective
$\phi_{\vz{1}}(\vu{},\vgamma{}) := \vz{1}^T(\vgamma{}^{-1}) + h(\vgamma{})
+ R(\vu{},\vgamma{}) - g_1^*(\vz{1})$
is of the same decoupled penalized least squares form than a MAP criterion
\eqp{map-estim}. This decomposition suggests a double loop algorithm for
solving $\min_{\vgamma{}}\phi(\vgamma{})$. In {\em inner loop minimizations},
we solve $\min_{\vu{},\vgamma{}\succeq\vzero}\phi_{\vz{1}}$ for fixed
$\vz{1}\succ\vzero$, and in interjacent {\em outer loop updates}, we refit
$\vz{1}\leftarrow \argmin \phi_{\vz{1}}(\vu{},\vgamma{})$.

The MAP estimation objective \eqp{map-estim} and
$\phi_{\vz{1}}(\vu{},\vgamma{})$ have a similar form. Specifically,
recall that $-2 g_i(x_i) = \min_{\gamma_i\ge 0} x_i/\gamma_i + h_i(\gamma_i)$,
where $g_i(x_i) = \tilde{g}_i(x_i^{1/2})$ and $\tilde{g}_i(s_i) = \log t_i(s_i)
- b_i s_i$. The inner loop problem is
\begin{equation}\label{eq:inner-logconcave}
  \min_{\vu{},\vgamma{}\succeq\vzero}\phi_{\vz{1}}(\vu{},\vgamma{}) =
  \min_{\vu{}} \sigma^{-2}\|\vy{}-\mxx{}\vu{}\|^2 - 2
  \sum\nolimits_{i=1}^q \left( g_i(z_{1,i}+s_i^2) + b_i s_i \right),
\end{equation}
where $\vs{}=\mxb{}\vu{}$.
This is a smoothed version of the MAP estimation problem, which would be
obtained for $z_{1,i}=0$. However, $z_{1,i}>0$ in our approximate inference
algorithm at all times (see \secref{ldeta-bounds}). Upon inner loop convergence
to $\vu{*}$, $\gamma_{*,i} = -1/[2 (d g_i/d x_i)|_{x_i=z_{1,i}+s_{*,i}^2}]$, where
$\vs{*}=\mxb{}\vu{*}$. Note
that in order to run the algorithm, the analytic form of $h_i(\gamma_i)$ need
not be known. For Laplace potentials \eqp{laplace}, the inner loop penalizer
is $2\sum_i \tau_i\sqrt{z_{1,i}+s_i^2}$, and $\gamma_{*,i} =
\sqrt{z_{1,i}+s_{*,i}^2}/\tau_i$.

Importantly, the inner loop problem \eqp{inner-logconcave} is of the same
simple penalized least squares form than MAP estimation, and any of the wide
range of recent efficient solvers can be plugged into our method. For example,
the {\em iteratively reweighted least squares}
(IRLS) algorithm \cite{Green:84}, a variant of the Newton-Raphson method, can
be used (details are given in \secref{inner-loop}). Each Newton step requires
the solution of a linear system with a matrix of the same form as
$\mxa{}$ \eqp{gauss-post}, the convergence rate of IRLS is quadratic. It
follows from the derivation of \eqp{phi-crit} that once an inner loop has
converged to $(\vu{*},\vgamma{*})$, the minimizer $\vu{*}$ is the mean
of the approximate posterior $Q(\vu{}|\vy{})$ for $\vgamma{*}$.

The rationale behind our algorithms lies in {\em decoupling} the variational
criterion $\phi$ via a Fenchel duality upper bound, thereby matching
algorithmic scheduling to the computational complexity structure of $\phi$.
To appreciate this point, note that in an off-the-shelf optimizer applied to
$\min_{\vgamma{}\succeq\vzero}\phi(\vgamma{})$, both $\phi(\vgamma{})$ and the
gradient $\nabla_{\vgamma{}}\phi$ have to be computed frequently. In this
respect, the $\log|\mxa{}|$ coupling term proves by far more computationally
challenging than the rest (see \secref{estim-vars}). This obvious computational
difference between parts of $\phi(\vgamma{})$ is not exploited in standard
gradient based algorithms: they require all of $\nabla_{\vgamma{}}\phi$ in each
iteration, all of $\phi(\vgamma{})$ in every single line search step. As
discussed in \secref{estim-vars}, computing $\log|\mxa{}|$ to high accuracy
is not feasible for models of interest here, and most off-the-shelf
optimizers with fast convergence rates are very hard to run with such
approximately computed
criteria. In our algorithm, the critical part is recognized and decoupled,
resulting inner loop problems can be solved by robust and efficient standard
code, requiring a minimal effort of adaptation. Only at the start of each
outer loop step, $\phi_{\vz{1}}$ has to be refitted: $\vz{1}\leftarrow
\nabla_{\vgamma{}^{-1}}\log|\mxa{}|$ (see \secref{ldeta-bounds}), the
computationally critical part of $\nabla_{\vgamma{}}\phi$ is required there.
Fenchel duality bounding\footnote{
  Note that Fenchel duality bounding is also used in difference-of-convex
  programming, a general framework to address non-convex, typically
  non-smooth optimization problems in a double loop fashion. In our application,
  $\phi(\vgamma{})$ is smooth in general and convex in many applications
  (see \secref{inference}): our reasons for applying bound minimization are
  different.}
is used to minimize the number of these costly steps (further advantages are
noted at the end of \secref{estim-vars}). Resulting double loop algorithms
are simple to implement based on efficient penalized least squares
reconstruction code, taking full advantage of the very well-researched state
of the art for this setup.

\subsection{The General Case}\label{sec:algo-general}

In this section, we generalize the double loop algorithm along two
directions. First, if potentials $\log t_i(s_i)$ are not log-concave, the
inner loop problems \eqp{inner-logconcave} are not convex in general
(\thref{convex-inf}), yet a simple variant can be used to remedy this
defect. Second, as detailed in \secref{ldeta-bounds}, there are different
ways of decoupling $\log|\mxa{}|$, giving rise to different algorithms.
In this section, we concentrate on developing these variants, their practical
differences and implications thereof are elaborated on in
\secref{map-estim}.

If $t_i(s_i)$ is not log-concave, then $h_i(\gamma_i)$ is not convex in general
(\thref{convex-inf}). In this case, we can write $h_i(\gamma_i) = 
h_{\cap,i}(\gamma_i) + h_{\cup,i}(\gamma_i)$, where $h_{\cap,i}$ is concave
and nondecreasing, $h_{\cup,i}$ is convex. Such a decomposition is not unique,
and has to be chosen for each $h_i$ at hand. With hindsight, $h_{\cap,i}$
should be chosen as small as possible (for example, $h_{\cap,i}\equiv 0$ if
$t_i(s_i)$ is
log-concave, the case treated above), and if IRLS is to be used for inner
loop minimizations (see \secref{inner-loop}), $h_{\cup,i}$ should be twice
continuously differentiable. For Student's t potentials \eqp{student-t}, such
a decomposition is given in \appref{impl-details}.
We define $h_{\cap}(\vgamma{}) = \sum_i h_{\cap,i}(\gamma_i)$,
$h_{\cup}(\vgamma{}) = \sum_i h_{\cup,i}(\gamma_i)$, and modify outer loop
updates by applying a second Fenchel duality bounding operation:
$h_{\cap}(\vgamma{})\le \tvz{2}^T\vgamma{} - \tilde{g}_2^*(\tvz{2})$, resulting
in a variant of the inner loop criterion \eqp{inner-logconcave}. If
$h_{\cap,i}$ is differentiable, the outer loop update is $\tilde{z}_2\leftarrow
h_{\cap,i}'(\gamma_i)$, otherwise any element from the subgradient can be
chosen (note that $\tilde{z}_2\ge 0$, as $h_{\cap,i}$ is nondecreasing).
Moreover, as
shown in \secref{ldeta-bounds}, Fenchel duality can be employed in order to
bound $\log|\mxa{}|$ in two different ways, one employed above, the other
being $\log|\mxa{}|\le \vz{2}^T\vgamma{} - \vone^T(\log\vgamma{}) -
g_2^*(\vz{2})$, $\vz{2}\succeq\vzero$. Combining these bounds (by adding
$\tvz{2}$ to $\vz{2}$), we obtain
\[
  \phi(\vgamma{},\vu{})\le \phi_{\vz{}}(\vu{},\vgamma{}) :=
  \vz{1}^T(\vgamma{}^{-1}) + \vz{2}^T\vgamma{} -
  \vz{3}^T(\log\vgamma{}) + h_{\cup}(\vgamma{}) + R(\vu{},\vgamma{}) -
  g^*(\vz{}),
\]
where $z_{3,i}\in\{0,1\}$, and $g^*(\vz{})$ collects the offsets of all
Fenchel duality bounds. Note that $\vz{j}\succeq\vzero$ for $j=1,2,3$, and
for each $i$, either $z_{1,i}>0$ and $z_{3,i}=0$, or $z_{1,i}=0$ and
$z_{2,i}>0, z_{3,i}=1$. We have that
\begin{equation}\label{eq:inner-crit}
\begin{split}
  \phi_{\vz{}}(\vu{}) & := \min_{\vgamma{}\succeq\vzero}\phi_{\vz{}}(\vu{},
  \vgamma{}) = \sigma^{-2}\|\vy{}-\mxx{}\vu{}\|^2
  + 2 \sum\nolimits_{i=1}^q h_i^*(s_i) - 2\vb{}^T\vs{}, \\
  h_i^*(s_i) & := \frac{1}2 \min_{\gamma_i\ge 0}\; (z_{1,i}+s_i^2)/\gamma_i
  + z_{2,i}\gamma_i - z_{3,i}\log\gamma_i + h_{\cup,i}(\gamma_i),\quad
  \vs{}=\mxb{}\vu{}.
\end{split}
\end{equation}
Note that $h_i^*(s_i)$ is convex as minimum (over $\gamma_i\ge 0$) of a
jointly convex argument \cite[Sect.~3.2.5]{Boyd:02}. The inner loop
minimization problem
$\min_{\vu{}}(\min_{\vgamma{}}\phi_{\vz{}})$ is of penalized least squares form
and can be solved with the same array of efficient algorithms applicable to
the special case \eqp{inner-logconcave}. An application of the second order
IRLS method is detailed in \secref{inner-loop}. A schema for the full
variational inference algorithm is given in \algref{double-loop}.

\begin{algorithm}[ht]
\begin{algorithmic}
\REPEAT
  \IF{first outer loop iteration}{
    \STATE{Initialize bound $\phi_{\vz{}}$. $\vu{}=\vzero$.}
  }
  \ELSE{
    \STATE{Outer loop update: Refit upper bound $\phi_{\vz{}}$ to $\phi$
      (tangent at $\vgamma{}$).}
    \STATE{Requires marginal variances $\hat{\vz{}}=
      \diag^{-1}(\mxb{}\mxa{}^{-1}\mxb{}^T)$ (\secref{estim-vars}).}
    \STATE{Initialize $\vu{}=\vu{*}$ (previous solution).}
  }
  \ENDIF
  \REPEAT
    \STATE{Newton (IRLS) iteration to minimize
      $\min_{\vgamma{}\succeq\vzero}\phi_{\vz{}}$
      \eqp{inner-crit} w.r.t.\ $\vu{}$.}
    \STATE{Entails solving a linear system (by LCG) and line search
      (\secref{inner-loop}).}
  \UNTIL{$\vu{*}=\argmin_{\vu{}}(\min_{\vgamma{}\succeq\vzero}\phi_{\vz{}})$
    converged}
  \STATE{Update $\vgamma{}=\argmin_{\vgamma{}\succeq\vzero}
    \phi_{\vz{}}(\vu{*},\cdot)$.}
\UNTIL{outer loop converged}
\end{algorithmic}
\caption{\label{alg:double-loop} Double loop variational inference algorithm}
\end{algorithm}

The algorithms are specialized to the $t_i(s_i)$ through $h_i^*(s_i)$
and its derivatives. The important special case of log-concave $t_i(s_i)$
has been detailed above. For Student's t potentials \eqp{student-t}, a
decomposition is detailed in \appref{impl-details}. In this case, the
overall problem $\min_{\vgamma{}\succeq\vzero}\phi(\vgamma{})$ is not convex,
yet our double loop algorithm iterates over standard-form convex inner
loop problems. Finally, for log-concave $t_i(s_i)$ and $z_{2,i}\ne 0$ (type B
bounding, \secref{ldeta-bounds}), our algorithm can be implemented generically
as detailed in \appref{gen-inner}.

We close this section establishing some characteristics of these algorithms.
First, we found it useful to initialize them with constant $\vz{1}$ and/or
$\vz{2}$ of small size, and with $\vu{}=\vzero$. Moreover, each subsequent
inner loop minimization is started with $\vu{}=\vu{*}$ from the last round.
The development of our algorithms is inspired by the sparse estimation
method of \cite{Wipf:08}, relationships to which are discussed in
\secref{map-estim}. Our algorithms are globally convergent, a stationary
point of $\phi(\vgamma{})$ is found from any starting point $\vgamma{}\succ
\vzero$ (recall from \thref{convex-inf} that for log-concave potentials,
this stationary point is a global solution). This is seen as detailed in
\cite{Wipf:08}. Intuitively, at the beginning of each outer loop iteration,
$\phi_{\vz{}}$ and $\phi$ have the same tangent plane at $\vgamma{}$, so that
each inner loop minimization decreases $\phi$ significantly unless
$\nabla_{\vgamma{}}\phi=\vzero$. Note that this convergence proof requires that
outer loop updates are done exactly, this point is elaborated on at the end
of \secref{estim-vars}.

Our variational {\em inference} algorithms differ from previous
methods\footnote{
  This comment holds for approximate {\em inference} methods. For sparse
  {\em estimation}, large scale algorithms are available (see
  \secref{map-estim}).}
in that orders of magnitude larger models can successfully be addressed.
They apply to the particular variational relaxation introduced in
\secref{inference}, whose relationship to other inference approximations is
detailed in \cite{Seeger:09c}. While most previous relaxations attain
scalability through many factorization assumptions
concerning the approximate posterior, $Q(\vu{}|\vy{})$ in our method is
fully coupled, sharing its conditional independence graph with the true
posterior $P(\vu{}|\vy{})$. A high-level view on our algorithms, discussed
in \secref{estim-vars}, is that we replace a priori independence
(factorization) assumptions by less damaging low rank approximations, tuned at
runtime to the posterior shape.

\subsection{Bounding $\log|\mxa{}|$}\label{sec:ldeta-bounds}

We need to upper bound $\log|\mxa{}|$ by a term which is convex and
decoupling in $\vgamma{}$. This can be done in two different ways using
Fenchel duality, giving rise to bounds with different characteristics.
Details for the development here are given in \appref{bound-a}.

Recall our assumption that $\mxa{}\succ\mxzero$ for each
$\vgamma{}\succ\vzero$. If $\vpi{} = \vgamma{}^{-1}$, then $\vpi{}\mapsto
\log|\tmxa{}(\vpi{})| = \log|\mxa{}|$ is concave for $\vpi{}\succ\vzero$
(\thref{ldeta-convex}(\ref{th1-2}) with $f\equiv\text{id}$). Moreover,
$\log|\tmxa{}(\vpi{})|$ is increasing and unbounded in each component of
$\vpi{}$ (\thref{sparse-infer}).
Fenchel duality \cite[Sect.~12]{Rockafellar:70} implies that
$\log|\tmxa{}(\vpi{})| = \min_{\vz{1}\succ\vzero} \vz{1}^T\vpi{} -
g_1^*(\vz{1})$ for $\vpi{}\succ\vzero$, thus $\log|\mxa{}| =
\min_{\vz{1}\succ\vzero} \vz{1}^T(\vgamma{}^{-1}) - g_1^*(\vz{1})$ for
$\vgamma{}\succ\vzero$.
Therefore, $\log|\mxa{}|\le \vz{1}^T(\vgamma{}^{-1}) - g_1^*(\vz{1})$. For
fixed $\vgamma{}\succ\vzero$, this is an equality for
\[
  \vz{1,*} = \nabla_{\vgamma{}^{-1}}\log|\mxa{}| = \hvz{} :=
  (\Var_Q[s_i|\vy{}]) =
  \diag^{-1}\left(\mxb{} \mxa{}^{-1}\mxb{}^T\right)\succ\vzero,
\]
and $g_1^*(\vz{1,*}) = \vz{1,*}^T(\vgamma{}^{-1}) - \log|\mxa{}|$. This is
called bounding type A in the sequel.

On the other hand, $\vgamma{}\mapsto \vone^T(\log\vgamma{}) + \log|\mxa{}|$
is concave for $\vgamma{}\succ\vzero$ (\thref{ldeta-convex}(\ref{th1-3}) with
$f\equiv\text{id}$). Employing
Fenchel duality once more, we have that $\log|\mxa{}|\le
\vz{2}^T\vgamma{} - \vone^T(\log\vgamma{}) - g_2^*(\vz{2})$, $\vz{2}\succeq
\vzero$. For any fixed $\vgamma{}$, equality is attained at
$\vz{2,*} = \vgamma{}^{-1}\circ(\vone - \vgamma{}^{-1}\circ\hat{\vz{}})$,
and $g_2^*(\vz{2,*}) = \vz{2,*}^T\vgamma{} -\log|\mxa{}|
-\vone^T(\log\vgamma{})$ at this point. This is referred to as bounding type
B.

\begin{figure}[ht!]
\begin{center}
  \includegraphics[width=0.8\columnwidth]{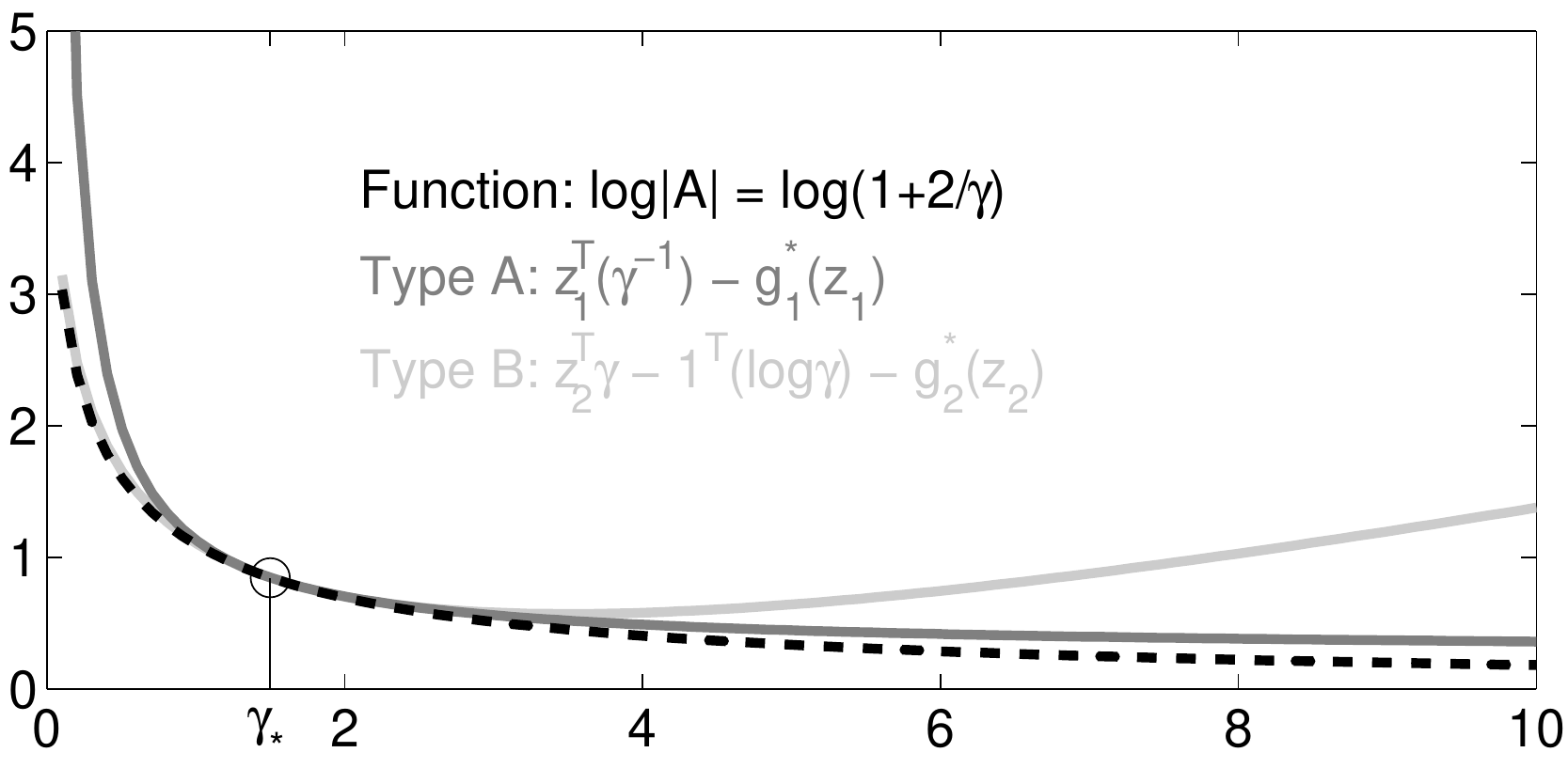}
\end{center}
\caption{\label{fig:bound-types}
  Comparison of type A and B upper bounds on $\log(1+2/\gamma)$.}
\end{figure}

In general, type A bounding is tighter for $\gamma_i$ away from zero,
while type B bounding is tighter for $\gamma_i$ close to zero (see
\figref{bound-types}), implications of this point are discussed in
\secref{map-estim}. Whatever bounding type we use,
refitting the corresponding upper bound to $\log|\mxa{}|$ requires the
computation of $\hvz{} = (\Var_Q[s_i|\vy{}])$: all marginal {\em variances} of
the Gaussian distribution $Q(\vs{}|\vy{})$. In general, computing Gaussian
marginal variances is a hard numerical problem, which is discussed in more
detail in \secref{estim-vars}.

\subsection{The Inner Loop Optimization}\label{sec:inner-loop}

The inner loop minimization problem is given by \eqp{inner-crit}, its special
case \eqp{inner-logconcave} for log-concave potentials and $\log|\mxa{}|$
bounding type A is given by $h_i^*(s_i) = -g_i(z_{1,i}+s_i^2)$. This problem
is of standard penalized least squares form, and a large number of recent
algorithms \cite{Goldstein:09,Bioucas:07,Wright:09} can be applied with little
customization efforts. In this section, we provide details about how to
apply the {\em iteratively reweighted least squares} (IRLS) algorithm
\cite{Green:84}, a special case of the Newton-Raphson method.

We describe a single IRLS step here, starting from $\vu{}$.
Let $\vr{}:=\mxx{}\vu{}-\vy{}$ denote the residual vector. If $\theta_i :=
(h_i^*)'(s_i) - b_i$, $\rho_i := (h_i^*)''(s_i)$, then
\[
  \nabla_{\vu{}}(\phi_{\vz{}}/2) = \sigma^{-2}\mxx{}^T\vr{}+\mxb{}^T\vth{},
  \quad \nabla\nabla_{\vu{}}(\phi_{\vz{}}/2) = \sigma^{-2}\mxx{}^T\mxx{} +
  \mxb{}^T(\diag \vrho{})\mxb{}.
\]
Note that $\rho_i\ge 0$ by the convexity of $h_i^*$. The Newton search
direction is
\[
  \vd{} := -(\sigma^{-2}\mxx{}^T\mxx{}+\mxb{}^T(\diag \vrho{})\mxb{})^{-1}
  (\sigma^{-2}\mxx{}^T\vr{} + \mxb{}^T\vth{}).
\]
The computation of $\vd{}$ requires us to solve a system with a matrix of the
same form as $\mxa{}$, a reweighted least squares problem otherwise used to
compute the means in a Gaussian model of the structure of $Q(\vu{}|\vy{})$. We
solve these systems approximately by the (preconditioned)
{\em linear conjugate gradients} (LCG) algorithm \cite{Golub:96}. The cost
per iteration of LCG is dominated by matrix-vector multiplications (MVMs)
with $\mxx{}^T\mxx{}$, $\mxb{}$, and
$\mxb{}^T$. A line search along $\vd{}$ can be run in negligible time. If
$f(t) := \phi_{\vz{}}(\vu{} + t\vd{})/2$, then
$f'(t) = \sigma^{-2}((\mxx{}\vd{})^T\vr{} + t\|\mxx{} \vd{}\|^2) +
(\mxb{}\vd{})^T\vth[t]{}$, where $\vth[t]{}$ is the gradient at
$\vs[t]{} = \vs{} + t \mxb{} \vd{}$. With $(\mxx{}\vd{})^T\vr{}$,
$\|\mxx{}\vd{}\|^2$,
and $\mxb{}\vd{}$ precomputed, $f(t)$ and $f'(t)$ can be evaluated in $O(q)$
without any further MVMs. The line search is started with $t_0 = 1$.
Finally, once $\vu{*} = \argmin_{\vu{}}\phi_{\vz{}}(\vu{})$ is found,
$\vgamma{}$ is explicitly updated as $\argmin\phi_{\vz{}}(\vu{*},\cdot)$.
Note that at this point, $\vu{*}=\Ex_Q[\vu{}|\vy{}]$, which follows from the
derivation at the beginning of \secref{inference}.

\subsection{Estimation of Gaussian Variances}\label{sec:estim-vars}

Variational inference does require marginal variances $\hvz{} =
\diag^{-1}(\mxb{}\mxa{}^{-1}\mxb{}^T) = (\Var_Q[s_i|\vy{}])$ of the Gaussian
$Q(\vs{}|\vy{})$ (see \secref{ldeta-bounds}), which are much harder to
approximate than means. In this section, we discuss a general method for
(co)variance approximation. Empirically, the performance of our double loop
algorithms is remarkably robust in the light of substantial overall variance
approximation errors, some insights into this finding are given below.

Marginal posterior variances have to be computed in any approximate
Bayesian inference method, while they are not required in typical sparse
point estimation techniques (see \secref{map-estim}). Our double loop
algorithms reduce approximate inference to point estimation and
Gaussian (co)variance approximation. Not only do they expose the latter as
missing link between sparse estimation and variational inference, their
main rationale is that Gaussian variances have to be computed a few times only,
while off-the-shelf variational optimizers query them for every single
criterion evaluation.

Marginal variance approximations have been proposed for sparsely connected
Gaussian Markov random fields (MRFs), iterating over
embedded spanning tree models \cite{Wainwright:01} or exploiting rapid
correlations decay in models with homogeneous prior \cite{Malioutov:06}.
In applications of interest here, $\mxa{}$ is neither sparse
nor has useful graphical model structure. Committing to a low-rank
approximation of the covariance $\mxa{}^{-1}$ \cite{Malioutov:06,Schneider:01},
an optimal choice in terms of preserving variances is principal components
analysis (PCA), based on the smallest eigenvalues/-vectors of $\mxa{}$ (the
largest of $\mxa{}^{-1}$). The {\em Lanczos algorithm} \cite{Golub:96}
provides a scalable approximation to PCA and was employed for variance
estimation in
\cite{Schneider:01}. After $k$ iterations, we have an orthonormal basis
$\mxq{k}\in\R^{n\times k}$, within which extremal eigenvectors of $\mxa{}$
are rapidly well approximated (due to the nearly {\em linear} spectral decay
of typical $\mxa{}$ matrices (\figref{exp-vars}, upper panel), both largest
and smallest eigenvalues are obtained). As $\mxq{k}^T\mxa{}\mxq{k}=\mxt{k}$ is
tridiagonal, the Lanczos variance approximation $\hvz{k} =
\diag^{-1}(\mxb{}\mxq{k}\mxt{k}^{-1}\mxq{k}^T\mxb{}^T)$ can be computed
efficiently. Importantly, $\hat{z}_{k,i}\le \hat{z}_{k+1,i}\le \hat{z}_i$ for
all $k$ and $i$. Namely, if $\mxq{k}=[\vq{1}\dots\vq{k}]$, and $\mxt{k}$ has
main diagonal $(\alpha_l)$,
subdiagonal $(\beta_l)$, let $(e_l)$ and $(d_l)$ be the
main and subdiagonal of the bidiagonal Cholesky factor $\mxl{k}$ of
$\mxt{k}$. Then, $d_{k-1} = \beta_{k-1} / e_{k-1}$, $e_k = (\alpha_k -
d_{k-1}^2)^{1/2}$, with $d_0=0$. If $\mxv{k} := \mxb{}\mxq{k}\mxl{k}^{-T}$, we
have $\mxv{k}=[\vv{1}\dots\vv{k}]$, $\vv{k} = (\mxb{}\vq{k} -
d_{k-1}\vv{k-1})/e_k$. Finally, $\hvz{k} = \hvz{k-1} + \vv{k}^2$ (with
$\hvz{0}=\vzero$).

Unfortunately, the Lanczos algorithm is much harder to run in practice than
LCG, and its cost grows superlinearly in $k$.
A promising variant of selectively reorthogonalized Lanczos \cite{Parlett:79}
is given in \cite{Bekas:08}, where contributions from undesired parts of the
spectrum ($\mxa{}$'s largest eigenvalues in our case) are filtered out by
replacing $\mxa{}$ with polynomials of itself. Recently, randomized PCA
approaches have become popular \cite{Halko:09}, although their relevance for
variance approximation is unclear. Nevertheless, for large scale problems of
interest, standard Lanczos can be run for $k\ll n$ iterations only, at
which point most of the $\hat{z}_{k,i}$ are severely underestimated
(see \secref{exp-estim-vars}). Since Gaussian variances are essential for
variational Bayesian inference, yet scalable, uniformly accurate variance
estimators are not known, {\em robustness} to variance approximations errors
is critical for any large scale approximate inference algorithm.

What do the Lanczos variance approximation errors imply for our double loop
algorithms? First, the global convergence proof of \secref{algo-general}
requires exact variances $\hvz{}$, thus may be compromised if $\hvz{k}$ is
used instead. This problem is analyzed in \cite{Seeger:10a}: the convergence
proof remains valid with the PCA approximation, which however is different
from the Lanczos\footnote{
  While Lanczos can be used to compute the PCA approximation (fixed number
  $L$ of smallest eigenvalues/-vectors of $\mxa{}$), this is rather wasteful.}
approximation. Empirically, we did not encounter convergence problems so far.

Surprisingly, while $\hvz{k}$ is much smaller than $\hvz{}$ in practice, there
is little indication of substantial negative impact on performance. This
important robustness property is analyzed in \secref{exp-estim-vars} for an
SLM with Laplace potentials.
The underestimation bias has systematic structure (\figref{exp-vars},
middle and lower panel): moderately small $\hat{z}_i$ are damped most strongly,
while large $\hat{z}_i$ are approximated accurately. This happens because
the largest coefficients $\hat{z}_i$ depend most strongly on the largest
covariance eigenvectors, which are shaped in early Lanczos iterations. This
particular error structure has statistical consequences.
Recalling the inner loop penalty for Laplacians \eqp{laplace}: 
$h_i^*(s_i) = \tau_i(\hat{z}_i+s_i^2)^{1/2}$, the smaller
$\hat{z}_i$, the stronger it enforces sparsity. If $\hat{z}_i$ is
underestimated, the penalty on $s_i$ is stronger than intended, yet this
strengthening does not happen uniformly. Coefficients $s_i$ deemed most
relevant with exact variance computation (largest $\hat{z}_i$) are least
affected (as $\hat{z}_{k,i}\approx\hat{z}_i$ for those), while already
subdominant ones (smaller $\hat{z}_i$) are suppressed even stronger (as
$\hat{z}_{k,i}\ll\hat{z}_i$). At least in our experience so far (with
sparse linear models), this selective variance underestimation effect seems
benign or even somewhat beneficial.


\subsection{Extension to Group Potentials}
\label{sec:group-pots}

There is substantial recent interest in methods incorporating sparse
{\em group penalization}, meaning that a number of latent coefficients
(such as the column of a matrix, or the incoming edge weights for a graph)
are penalized jointly \cite{Yuan:06,Tropp:06}. Our algorithms are easily
generalized to models with potentials of the form $t_i(\|\vs{i}\|)$, $\vs{i}$
a subvector of $\vs{}$, $\|\cdot\|$ the Euclidean norm, if $t_i(\cdot)$ is
even and super-Gaussian.
Such {\em group potentials} are frequently used in practice. The isotropic
total variation penalty is the sum of $\|(dx_i,dy_i)\|$, $dx_i$, $dy_i$
differences along different axes, which corresponds to group Laplace
potentials. In our MRI application (\secref{exp-mri}), we deal with
complex-valued $\vu{}$ and $\vs{}$. Each entry is treated as element in
$\R^2$, and potentials are placed on $|s_i| = \|(\Re s_i,\Im s_i)\|$.
Note that with $t_i$ on $\|\vs{i}\|$, the single parameter $\gamma_i$ is shared
by the coefficients $\vs{i}$.

The generalization of our algorithms to group potentials is almost automatic.
For example, if all $\vs{i}$ have the same dimensionality, $\mxgamma{}^{-1}$
is replaced by $\mxgamma{}^{-1}\otimes\Id$ in the definition of $\mxa{}$, and
$\hat{\vz{}}$ is replaced by $(\Id\otimes\vone{}^T)
\diag^{-1}(\mxb{}\mxa{}^{-1}\mxb{}^T)$ in \secref{ldeta-bounds}. Moreover,
$x_i=s_i^2$ is replaced by $x_i=\|\vs{i}\|^2$, whereas the definition of
$g_i(x_i)$ remains the same. Apart from these simple replacements, only IRLS
inner loop iterations have to be modified (at no extra cost), as is detailed
in \appref{norm-sites}.

\subsection{Publicly Available Code: The \texttt{glm-ie} Toolbox}
\label{sec:glm-ie}

Algorithms and techniques presented in this paper are implemented\footnote{
  Our experiments in \secref{exper} use different {\tt C++} and Fortran code,
  which differs from \texttt{glm-ie} mainly by being somewhat faster on large
  problems.}
as part of the {\em generalized linear model inference and estimation} toolbox
(\texttt{glm-ie}), maintained as \texttt{mloss.org} project at
\url{mloss.org/software/view/269/}. The code runs with both {\tt Matlab~7} and
the free {\tt Octave~3.2}. It comprises algorithms for MAP (penalized least
squares) estimation and variational inference in generalized linear models
(\secref{algorithm}), along with Lanczos code for Gaussian variances
(\secref{estim-vars}).

Its generic design allows for a range of applications, as illustrated by a
number of example programs included in the package.
Many super-Gaussian potentials $t_i(s_i)$ are included, others can
easily be added by the user. In particular, the toolbox contains a range of
solvers for MAP and inner loop problems, from IRLS (or truncated Newton, see
\secref{inner-loop}) over conjugate gradients to Quasi-Newton, as well as a
range of commonly used operators to construct $\mxx{}$ and $\mxb{}$ matrices.

\section{Sparse Estimation and Sparse Bayesian Inference}
\label{sec:map-estim}

In this section, we contrast approximate Bayesian inference with point
estimation for sparse linear models (SLMs): {\em sparse Bayesian inference}
versus {\em sparse estimation}. These problem classes serve distinct goals
and come with different algorithmic characteristics, yet are frequently
confused in the literature. Briefly, the goal in sparse estimation is to
eliminate variables not needed for the task at hand, while sparse inference
aims at quantifying uncertainty in decisions and dependencies between
components. While variable elimination is a boon for efficient computation,
it cannot be relied upon in sparse inference. Sensible uncertainty estimates
like posterior covariance, at the heart of decision-making problems such as
Bayesian experimental design, are eliminated alongside.

We restrict ourselves to super-Gaussian SLM problems in terms of variables
$\vu{}$ and $\vgamma{}\succeq\vzero$, relating the sparse Bayesian inference
relaxation $\min_{\vgamma{}\succeq\vzero}\phi(\vgamma{})$ with two
sparse estimation principles: maximum a posteriori (MAP) reconstruction
\eqp{map-estim} and automatic relevance determination (ARD) \cite{Wipf:08}, a
sparse reconstruction method which inspired our algorithms. We begin by
establishing a key difference between these settings. Recall from
\thref{ldeta-convex}(\ref{th1-4}) that $\gamma_i=0$ implies\footnote{
  While the proof of \thref{ldeta-convex}(\ref{th1-4}) holds for
  $\vgamma{}\succ\vzero$, $\Var_Q[s_i|\vy{}]$ is a continuous function of
  $\vgamma{}$.}
$\Var_Q[s_i|\vy{}]=0$: $s_i$ is {\em eliminated}, fixed at zero with absolute
certainty. Exact sparsity in $\vgamma{}$ does not happen for
{\em inference}, while sparse {\em estimation} methods are characterized by
fixing many $\gamma_i$ to zero.

\begin{theorem}\label{th:sparse-infer}
Let $\mxx{}\in \R^{m\times n}$, $\mxb{}\in \R^{q\times n}$ be matrices such
that $\tmxa{}(\vpi{}) = \sigma^{-2}\mxx{}^T\mxx{} +
\mxb{}^T(\diag\vpi{})\mxb{}\succ\mxzero$ for each $\vpi{}\succ\vzero$,
and no row of $\mxb{}$ is equal to $\vzero^T$.
\begin{itemize}
\item
  The function $\log|\tmxa{}(\vpi{})|$ is increasing in each component
  $\pi_i$, and unbounded above. For any sequence $\vpi{t}$ with
  $\|\vpi{t}\|\to\infty$ ($t\to\infty$) and $\vpi{t}\succeq\eps\vone$ for
  some $\eps>0$, we have that
  $\log|\tmxa{}(\vpi{t})|\to\infty$ ($t\to\infty$).
\item
  Assume that $\log P(\vu{}|\vy{})$ is bounded above as
  function of $\vu{}$.
  Recall the variational criterion $\phi(\vgamma{})$ from
  \eqp{phi-crit}. For any bounded sequence $\vgamma{t}$ with
  $(\vgamma{t})_i\to 0$ ($t\to\infty$) for some $i\in\srng{q}$, we have
  that $\phi(\vgamma{t})\to\infty$. \\
  In particular, any local minimum point $\vgamma{*}$ of the variational
  inference problem $\min_{\vgamma{}\succeq\vzero}\phi(\vgamma{})$ must have
  positive components: $\vgamma{*}\succ\vzero$.
\end{itemize}
\end{theorem}

A proof is given in \appref{sparse-infer}. $\log|\mxa{}|$ acts as barrier
function for $\vgamma{}\succ\vzero$. Any local minimum point $\vgamma{*}$
of \eqp{phi-crit} is positive throughout, and $\Var_Q[s_i|\vy{}]>0$ for all
$i=\rng{q}$. Coefficient elimination does not happen in variational Bayesian
inference.

Consider MAP estimation \eqp{map-estim} with even super-Gaussian potentials
$t_i(s_i)$. Following \cite{Rao:03}, a sufficient condition for sparsity is
that $-\log t_i(s_i)$ is concave for $s_i>0$. In this case, if $\rank\mxx{}=m$
and $\rank\mxb{}=n$, then any local MAP solution $\vu{*}$ is exactly sparse:
no more than $m$ coefficients of $\vs{*}=\mxb{}\vu{*}$ are nonzero. Examples
are $t_i(s_i) = e^{-\tau_i|s_i|^p}$, $p\in(0,1]$, including Laplace potentials
($p=1$). Moreover, $\gamma_{*.i}=0$ whenever $s_{*,i}=0$ in this case (see
\appref{sparse-infer}). Local minimum points of SLM MAP estimation are
substantially exactly sparse, with matching sparsity patterns of
$\vs{*}=\mxb{}\vu{*}$ and $\vgamma{*}$.

\begin{figure}[ht!]
\begin{center}
\begin{tabular}{cc}
  \includegraphics[width=0.8\columnwidth]{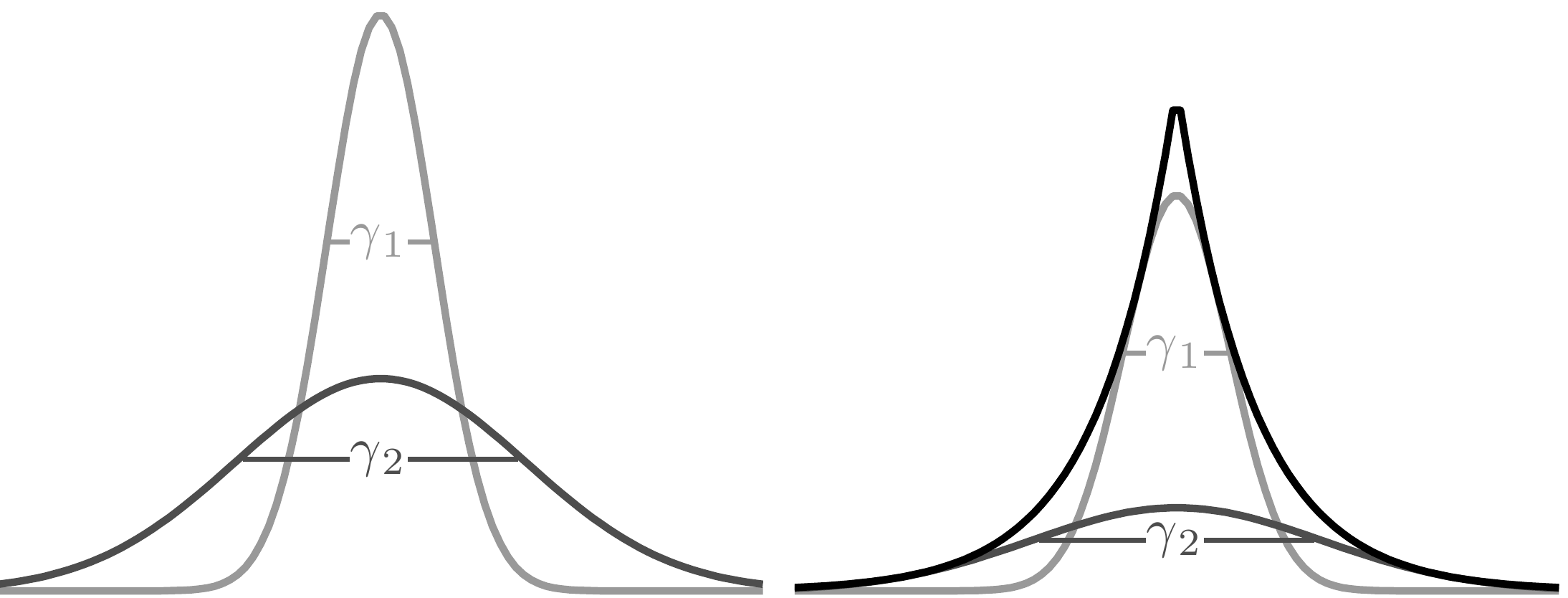}
\end{tabular}
\end{center}
\caption{\label{fig:estim-infer}
  Different roles of Gaussian functions (width $\gamma_i$) in sparse estimation
  versus sparse inference. Left: Sparse estimation (ARD).
  Gaussian functions are normalized, there is incentive to drive
  $\gamma_i\to 0$.
  Right: Variational inference for Laplace potentials \eqp{laplace}.
  Gaussian functions are lower bounds of $t_i(s_i)$, their mass vanishes
  as $\gamma_i\to 0$. No incentive to eliminate $\gamma_i$.}
\end{figure}

A powerful sparse estimation method, {\em automatic relevance determination}
(ARD) \cite{Wipf:08}, has inspired our approximate inference algorithms
developed above. The ARD criterion $\phi_{\text{ARD}}$ is \eqp{phi-crit}
with $h(\vgamma{})=\vone^T(\log\vgamma{})$, obtained as zero-temperature limit
($\nu\to 0$) of variational inference with Student's t
potentials \eqp{student-t}. The function $h_i(\gamma_i)$ is given in
\appref{impl-details}, and $h_i(\gamma_i)\to \log\gamma_i$ ($\nu\to 0$) if
additive constants independent of $\gamma_i$ are dropped.\footnote{
  Note that the term dropped ($C_i$ in \appref{impl-details}) becomes
  unbounded as $\nu\to 0$. Removing it is essential to obtain a well-defined
  problem.}
ARD can also be seen as marginal likelihood maximization:
$\phi_{\text{ARD}}(\vgamma{}) = -2\log\int P(\vy{}|\vu{})
N(\vs{}|\vzero,\mxgamma{})\, d\vu{}$ up to an additive constant. Sparsity
penalization is implied by the fact that the prior $N(\vs{}|\vzero,\mxgamma{})$
is {\em normalized} (see \figref{estim-infer}, left).
The ARD problem is not convex. A provably convergent double loop
ARD algorithm is obtained by employing bounding type B (\secref{ldeta-bounds}),
along similar lines to \secref{algo-general} we obtain
\[
  \min_{\vgamma{}\succeq\vzero}\phi_{\text{ARD}}(\vgamma{}) =
  \min_{\vz{2}\succeq\vzero}\left( \min_{\vu{}}
  \sigma^{-2}\|\vy{}-\mxx{}\vu{}\|^2 + 2\sum\nolimits_{i=1}^q z_{2,i}^{1/2}
  |s_i| \right) - g_2^*(\vz{2}).
\]
The inner problem is $\ell_1$ penalized least squares estimation, a reweighted
variant of MAP reconstruction for Laplace potentials. Its solutions
$\vs{*}=\mxb{}\vu{*}$ are exactly sparse, along with corresponding $\vgamma{*}$
(since $\gamma_{*,i} = z_{2,i}^{-1/2}|s_{*,i}|$). ARD is enforcing sparsity more
aggressively than Laplace ($\ell_1$) MAP reconstruction \cite{Wipf:10}. The
$\log|\mxa{}|$ barrier function
is counterbalanced by $h(\vgamma{}) = \vone^T(\log\vgamma{}) =
\log|\mxgamma{}|$. If $\mxb{}=\Id$, then
\[
  \log|\mxa{}| + \log|\mxgamma{}| = \log|\Id +
  \sigma^{-2}\mxx{}\mxgamma{}\mxx{}^T| \to 0\; (\vgamma{}\to\vzero).
\]
The conceptual difference between ARD and our variational inference relaxation
is illustrated in \figref{estim-infer}. In sparse inference,
Gaussian functions $e^{-s_i^2/(2\gamma_i) - h_i(\gamma_i)/2}$ lower bound
$t_i(s_i)$. Their mass vanishes as $\gamma_i\to 0$, driving $\phi(\vgamma{})
\to\infty$. For ARD, Gaussian functions $N(s_i|0,\gamma_i)$ are normalized,
and $\gamma_i\to 0$ is encouraged.

At this point, the roles of different bounding types introduced in
\secref{ldeta-bounds} become transparent. $\log|\mxa{}|$ is a barrier
function for $\vgamma{}\succ\vzero$ (\thref{sparse-infer}), as is its
type A bound $\vz{1}^T(\vgamma{}^{-1}) - g_1^*(\vz{1})$, $\vz{1}\succ\vzero$
(see \figref{bound-types}). On the other hand, $\log|\mxa{}| +
\vone^T(\log\vgamma{})$ is bounded below, as is its type B bound
$\vz{2}^T\vgamma{} - g_2^*(\vz{2})$. These facts suggest that type A bounding
should be preferred for variational inference, while type B bounding is best
suited for sparse estimation. Indeed, experiments in \secref{exp-classes}
show that for approximate inference with Laplace potentials, type A bounding
is by far the better choice, while for ARD, type B bounding leads to the very
efficient algorithm just sketched.

Sparse estimation methods eliminate a substantial fraction of $\vgamma{}$'s
coefficients, variational inference methods do not zero any of them.
This difference has important computational and statistical implications.
First, exact sparsity in $\vgamma{}$ is computationally beneficial. In this
regime, even coordinate descent algorithms can be scaled up to large problems
\cite{Tipping:03}. Within the ARD sparse estimation algorithm, variances
$\hvz{}\leftarrow\diag^{-1}(\mxb{}\mxa{}^{-1}\mxb{}^T)$ have to be computed,
but since $\hvz{}$ is as sparse as $\vgamma{}$, this is not a hard
problem.
Variational inference methods have to cope without exact sparsity. The double
loop algorithms of \secref{algorithm} are scalable nevertheless, reducing to
numerical techniques whose performance does not depend on the 
sparsity of $\vgamma{}$.

While exact sparsity in $\vgamma{}$ implies computational simplifications, it
also rules out proper model-based uncertainty quantification.\footnote{
  Uncertainty quantification may also be obtained by running sparse
  estimation many times in a bootstrapping fashion \cite{Meinshausen:09}.
  While such procedures cure some robustness issues of MAP estimation,
  they are probably too costly to run in order to drive experimental design,
  where dependencies between variables are of interest.}
If $\gamma_i=0$, then $\Var_Q[s_i|\vy{}]=0$. If $Q(\vu{}|\vy{})$ is understood
as representation of uncertainty, it asserts that there is {\em no
posterior variance} in $s_i$ at all: $s_i$ is eliminated with absolute
certainty, along with all correlations between $s_i$ and other $s_j$.
Sparsity in $\vgamma{}$ is computationally useful only if most $\gamma_i=0$.
$Q(\vu{}|\vy{})$, a degenerate distribution with mass only in the
subspace corresponding to surviving coefficients, cannot sensibly be
regarded as approximation to a Bayesian posterior. As zero is just zero, even
basic queries such as a confidence ranking over eliminated coefficients
cannot be based on a degenerate $Q(\vu{}|\vy{})$.

In particular, Bayesian experimental design (\secref{design}) cannot sensibly
be driven by underlying sparse {\em estimation} technology, while it excels
for a range of real-world scenarios (see \secref{exp-mri}) when based on a
sparse {\em inference} method \cite{Steinke:07,Seeger:07d,Seeger:08,
Seeger:08c}. The former approach is taken in \cite{Ji:07}, employing the
{\em sparse Bayesian learning} estimator \cite{Tipping:01} to drive inference
queries. Their approach fails badly on real-world image data \cite{Seeger:08}.
Started with few initial measurements, it identifies a very small subspace of
non-eliminated coefficients (as expected for sparse estimation fed with little
data), which it essentially remains locked in ever after. To sensibly score
a candidate $\mxx{*}$, we have to reason about what happens to {\em all}
coefficients, which is not possible based on a ``posterior'' $Q(\vu{}|\vy{})$
ruling out most of them with full certainty.

Finally, even if the goal is point reconstruction from given data, the
sparse inference posterior mean $\Ex_Q[\vu{}|\vy{}]$ (obtained as byproduct
$\vu{*}$ in the double loop algorithm of \secref{algorithm}) can be an
important alternative to an exactly sparse estimator. For the former,
$\Ex_Q[\vs{}|\vy{}] = \mxb{}\vu{*}$ is not sparse in general, and the degree
to which coefficients are penalized (but not eliminated) is determined by
the choice of $t_i(s_i)$. To illustrate this point, we compare the mean
estimator for Laplace and Student's t potentials (different $\nu>2$) in
\secref{exp-student-t}. These results demonstrate that contrary to some
folklore in signal and image processing, {\em sparser is not necessarily
better} for point reconstruction of real-world images. Enforcing sparsity
too strongly leads to fine details being smoothed out, which is not acceptable
in medical imaging (fine features are often diagnostically most relevant) or
photography postprocessing (most users strongly dislike unnaturally hard edges
and oversmoothed areas).

Sparse {\em estimation} methodology has seen impressive advancements towards
what it is intended to do: solving a given overparameterized reconstruction
problem by eliminating nonessential variables. However, it is ill-suited for
addressing decision-making scenarios driven by Bayesian {\em inference}.
For the latter, a useful (nondegenerate) posterior approximation has to be
obtained without relying on computational benefits of exact sparsity. We
show how this can be done, by reducing variational inference to numerical
techniques (LCG and Lanczos) which can be scaled up to large problems without
exact variable sparsity.

\section{Bayesian Experimental Design}
\label{sec:design}

In this section, we show how to optimize the image acquisition matrix
$\mxx{}$ by way of Bayesian sequential experimental design (also known as
Bayesian active learning),
maximizing the expected amount of information gained. Unrelated to the output
of point reconstruction methods, information gain scores depend on the
posterior covariance matrix $\Cov[\vu{}|\vy{}]$ over full images $\vu{}$,
which within our large scale variational inference framework is approximated
by the Lanczos algorithm.

In each round, a part $\mxx{*}\in \R^{d\times n}$ is appended to the design
$\mxx{}$, a new (partial) measurement $\vy{*}$ to $\vy{}$. Candidates
$\{\mxx{*}\}$ are ranked by the {\em information gain}\footnote{
  $\Ent[P(\vu{})] = \Ex_P[-\log P(\vu{})]$ is the (differential) entropy,
  measuring the amount of uncertainty in $P(\vu{})$. For a Gaussian,
  $\Ent[N(\vmu{},\mxsigma{})] = \frac{1}2\log|2\pi e \mxsigma{}|$.}
score $\Ent[P(\vu{} | \vy{})] -
\Ex_{P(\vy{*}|\vy{})}[ \Ent[P(\vu{}|\vy{},\vy{*})] ]$, where $P(\vu{}|\vy{})$ and
$P(\vu{}|\vy{},\vy{*})$ are posteriors for $(\mxx{},\vy{})$ and
$(\mxx{}\cup\mxx{*},\vy{}\cup\vy{*})$ respectively, and $P(\vy{*}|\vy{}) =
\int N(\vy{*}|\mxx{*}\vu{},\sigma^2\Id) P(\vu{}|\vy{})\, d\vu{}$.
Replacing $P(\vu{}|\vy{})$ by its best Gaussian variational approximation
$Q(\vu{}|\vy{}) = N(\vu{*},\mxa{}^{-1})$ and $P(\vu{}|\vy{},\vy{*})$ by
$Q(\vu{}|\vy{},\vy{*})\propto N(\vy{*}|\mxx{*}\vu{},\sigma^2\Id)
Q(\vu{}|\vy{})$, we obtain an approximate information gain score
\begin{equation}\label{eq:ed-score}
  \Delta(\mxx{*}) := -\log|\mxa{}| + \log\left| \mxa{} +
  \sigma^{-2}\mxx{*}^T \mxx{*} \right| = \log\left|\Id +
  \sigma^{-2}\mxx{*}\mxa{}^{-1}\mxx{*}^T\right|.
\end{equation}
Note that $Q(\vu{}|\vy{},\vy{*})$ has the same variational parameters
$\vgamma{}$ as $Q(\vu{}|\vy{})$, which simplifies and robustifies score
computations. Refitting of $\vgamma{}$ is done at the end of each
round, once the score maximizer $\mxx{*}$ is appended along with a new
measurement $\vy{*}$.

With $N$ candidates of size $d$ to be scored, a naive computation of
\eqp{ed-score} would require $N\cdot d$ linear systems to be solved, which
is not tractable (for example, $N=240$, $d=512$ in \secref{exp-mri}).
We can make use of the Lanczos
approximation once more (see \secref{estim-vars}). If $\mxq{k}^T\mxa{}\mxq{k}
= \mxt{k} = \mxl{k}\mxl{k}^T$ ($\mxl{k}$ is bidiagonal, computed in $O(k)$),
let $\mxv{*} := \sigma^{-1}\mxx{*}\mxq{k}\mxl{k}^{-T}\in\R^{d\times k}$.
Then, $\Delta(\mxx{*})\approx \log|\Id+\mxv{*}\mxv{*}^T| =
\log|\Id+\mxv{*}^T\mxv{*}|$ (the latter is preferable if $k<d$), at a total
cost of $k$  matrix-vector multiplications (MVMs) with $\mxx{*}$ and
$O(\max\{k,d\}\cdot \min\{k,d\}^2)$.
Just as with marginal variances, Lanczos approximations of $\Delta(\mxx{*})$
are underestimates, nondecreasing in $k$. The impact of Lanczos approximation
errors on design decisions is analyzed in \cite{Seeger:10a}. While absolute
score values are much too small, decisions only depend on the
{\em ranking} among the highest-scoring candidates $\mxx{*}$, which often is
faithfully reproduced even for $k\ll n$. To understand this point, note that
$\Delta(\mxx{*})$ measures the alignment of $\mxx{*}$ with the directions of
largest variance in $Q(\vu{}|\vy{})$. For example, the single best
unit-norm filter $\vx{*}\in\R^n$ is given by the maximal eigenvector of
$\Cov_Q[\vu{}|\vy{}]=\mxa{}^{-1}$, which is obtained after few Lanczos
iterations.

In the context of Bayesian experimental design, the convexity of our
variational inference relaxation (with log-concave potentials) is an
important asset. In contrast to single image reconstruction, which can be
tuned by the user until a desired result is obtained, sequential acquisition
optimization is an autonomous process consisting of many individual steps
(a real-world example is given in \secref{exp-mri}), each of which requires
a variational refitting $Q(\vu{}|\vy{})\to Q(\vu{}|\vy{},\vy{*})$. Within our
framework, each of these has a unique solution which is found by a very
efficient algorithm. While we are not aware of Bayesian acquisition
optimization being realized at comparable scales with other inference
approximations, this would be difficult to do indeed. Different variational
approximations are non-convex problems coming with notorious local minima
issues. For Markov chain Monte Carlo
methods, there are not even reliable automatic tests of convergence. If
approximate
inference drives a multi-step automated scheme free of human expert
interventions, properties like convexity and robustness gain relevance normally
overlooked in the literature.


\subsection{Compressive Sensing of Natural Images}\label{sec:compsens}

The main application we address in \secref{exp-mri}, automatic acquisition
optimization for magnetic resonance imaging, is an advanced real-world instance
of {\em compressive sensing} (CS) \cite{Candes:06,Bruckstein:09}. Given that
real-world images come with low entropy super-Gaussian statistics, how can we
tractably reconstruct them from a sample below the Nyquist-Shannon limit? How
do small successful designs $\mxx{}$ for natural images look like? Recent
celebrated results about recovery properties of convex sparse estimators
\cite{Donoho:03,Candes:06,Bruckstein:09} have been interpreted as suggesting
that up from a certain size, successful designs $\mxx{}$ may simply be drawn
blindly at random. Technically speaking, these results are about highly exactly
sparse signals (see \secref{map-estim}), yet advancements for {\em image}
reconstruction are typically being implied \cite{Candes:06,Bruckstein:09}. In
contrast, Bayesian experimental design is an adaptive approach, optimizing
$\mxx{}$ based on real-world training images. Our work is of the latter kind,
as are \cite{Ji:07,Seeger:08,Haupt:08} for much smaller scales.

The question whether a design $\mxx{}$ is useful for measuring images, can
(and should) be resolved empirically. Indeed, it takes not more than some
reconstruction code and a range of realistic images (natural photographs, MR
images) to convince oneself that MAP estimation from a subset of Fourier
coefficients drawn uniformly at random (say, at $1/4$ Nyquist) leads to very
poor results. This failure of blindly drawn designs is well established by now
both for natural images and MR images \cite{Seeger:08,Seeger:08c,Lustig:07,
Chang:09}, and is not hard to motivate. In a nutshell, the assumptions which
current CS theory relies upon do not sensibly describe realistic images.
Marginal statistics of the latter are not exactly sparse, but exhibit a power
law (super-Gaussian) decay. More important, their sparsity is highly
structured, a fact which is ignored in assumptions made by current CS theory,
therefore not reflected in recovery conditions (such as incoherence) or
in designs $\mxx{}$ drawn uniformly at random. Such designs fail for a
number of reasons. First, they do not sample where the image energy is
\cite{Seeger:08,Chang:09}. A more subtle problem is the inherent variability
of {\em independent} sampling in Fourier space: large gaps occur with high
probability, which leads to serious MAP reconstruction errors. These points are
reinforced in \cite{Seeger:08,Seeger:08c}. The former study finds that for
good reconstruction quality of real-world images, the choice of $\mxx{}$ is far
more important than the type of reconstruction algorithm used.

In real-world imaging applications, adaptive approaches promise remedies for
these problems (other proposals in this direction are \cite{Ji:07} and
\cite{Haupt:08}, which however have not successfully been applied to
real-world images).
Instead of relying on simplistic signal assumptions, they learn a design
$\mxx{}$ from realistic image data. Bayesian experimental design provides
a general framework for adaptive design optimization, driven not by point
reconstruction, but by predicting information gain through posterior
covariance estimates.

\section{Experiments}
\label{sec:exper}

We begin with a set of experiments designed to explore aspects and variants of
our algorithms, and to understand approximation errors. Our main
application concerns the optimization of sampling trajectories in magnetic
resonance imaging (MRI) sequences, with the aim of obtaining useful images
faster than previously possible.

\subsection{Type A versus Type B Bounding for Laplace Potentials}
\label{sec:exp-classes}

\begin{figure}[ht]
\centering
  \includegraphics[width=0.48\linewidth]{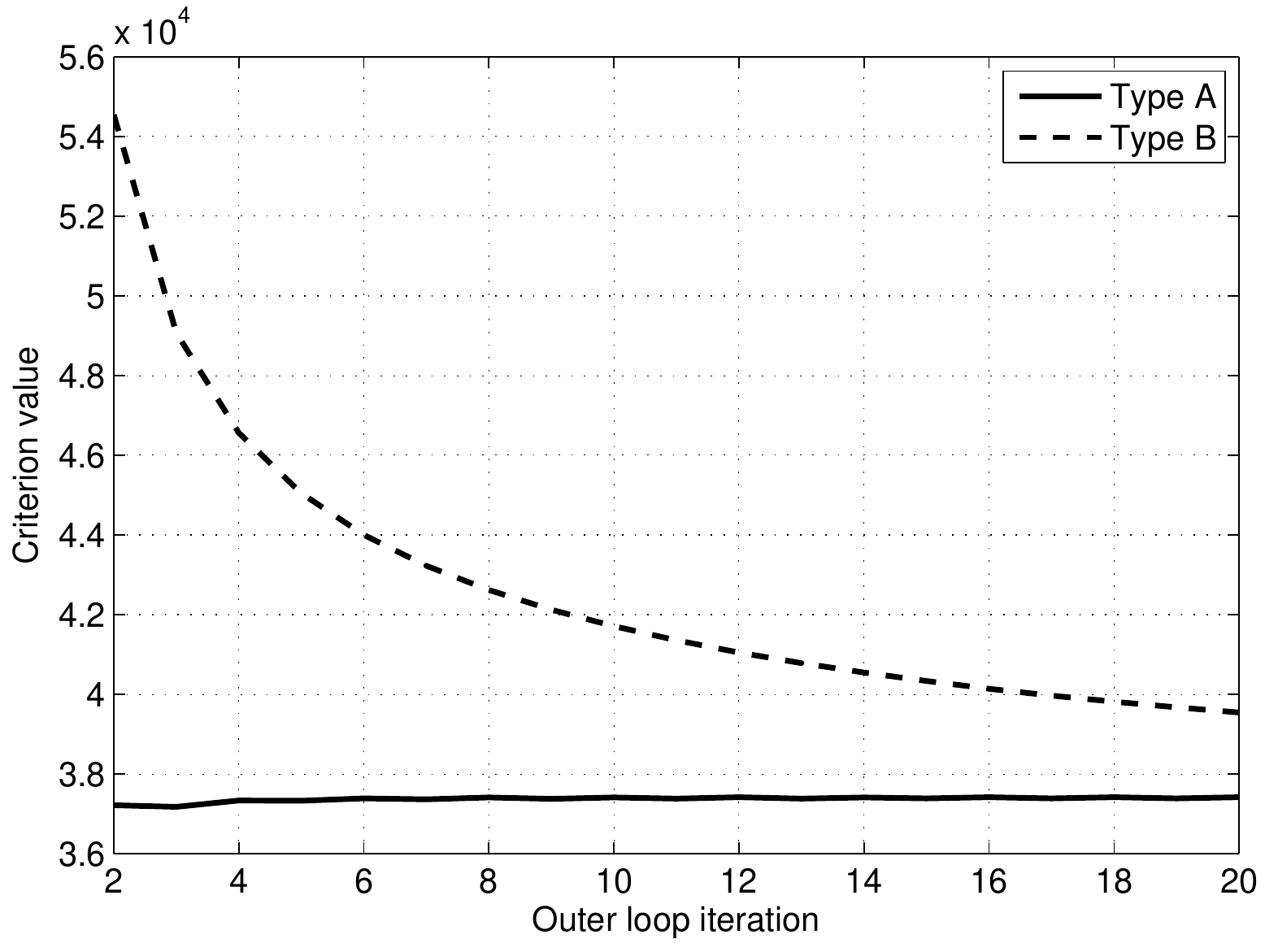} \hfill
  \includegraphics[width=0.48\linewidth]{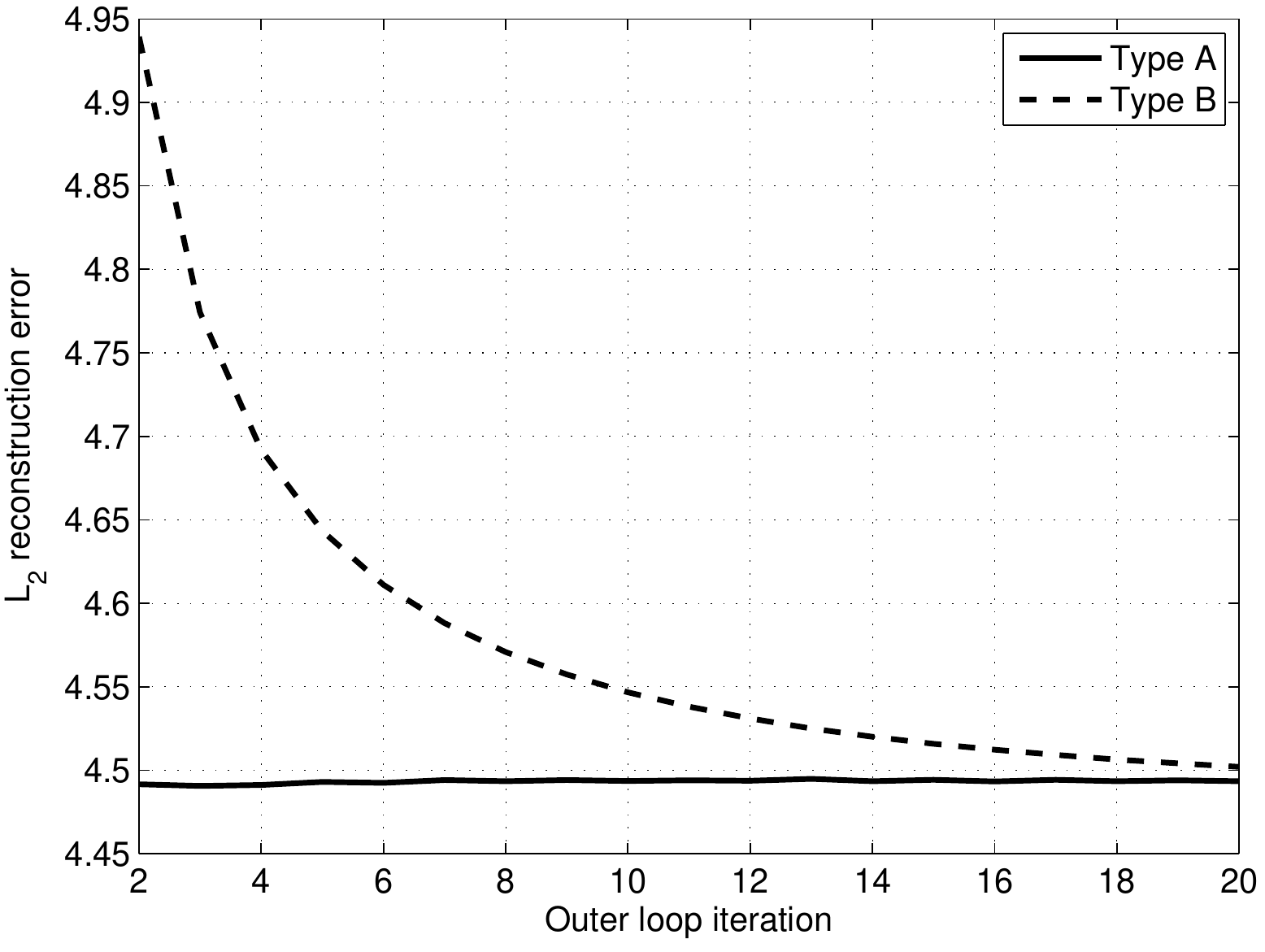}
\caption{\label{fig:exp-classes}\small
  Comparison of bounding types A, B for SLM with Laplace potentials.
  Shown are $\phi(\vgamma{})$ criterion values (left) and $\ell_2$
  errors of posterior mean estimates (not MAP, as in \secref{exp-mri}), at the
  end of each outer loop iteration (starting from second).
}
\end{figure}

Recall that the critical coupling term $\log|\mxa{}|$ in the variational
criterion $\phi(\vgamma{})$ can be upper bounded in two different ways, called
type A and type B in \secref{ldeta-bounds}. Type A is tight for moderate and
large $\gamma_i$, type B for small $\gamma_i$ (\secref{map-estim}). In this
section, we run our inference algorithm with type A and type B bounding
respectively, comparing the speed of convergence. The setup (SLM with Laplace
potentials) is as detailed in \secref{exp-mri}, with a design $\mxx{}$ of
64 phase encodes ($1/4$ Nyquist). Results are given in \figref{exp-classes},
averaged over 7 different slices from {\tt sg88} ($256\times 256$ pixels,
$n=131072$).

In this case, the bounding type strongly influences the algorithm's progress.
While two outer loop (OL) iterations suffice for convergence with type
A, convergence is not attained even after 20 OL steps with type B. More inner
loop (IL) steps are done for type A (30 in first OL iteration, 3--4
afterwards) than for type B (5--6 in first OL iteration, 3--4 afterwards). The
double loop strategy, to make substantial progress with far less expensive IL
updates, works out for type A, but not for type B bounding.
These results indicate that bounding type A should be preferred for SLM
variational {\em inference}, certainly with Laplace potentials. Another
indication comes from comparing IL penalties $h_i^*(s_i)$ respectively. For
type A, $h_i^*(s_i) = \tau_i(z_{1,i}+s_i^2)^{1/2}$ is sparsity-enforcing for
small $z_{1,i}$, retaining an important property of $\phi(\vgamma{})$, while
for type B, $h_i^*(s_i)$ does not enforce sparsity at all (see
\appref{impl-details}).

\subsection{Student's t Potentials}
\label{sec:exp-student-t}

\begin{figure}[ht]
\centering
  \includegraphics[width=0.48\linewidth]{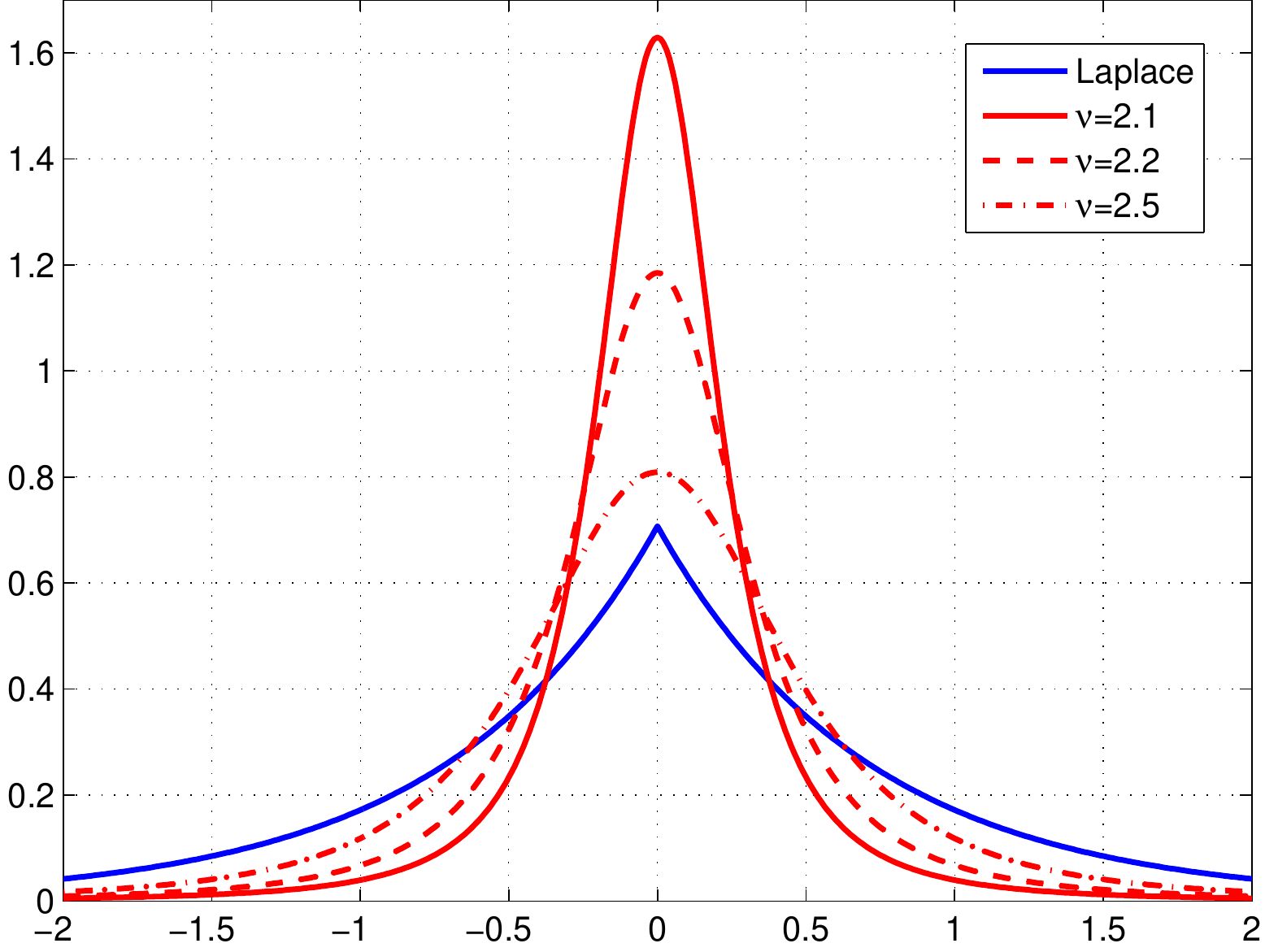} \hfill
  \includegraphics[width=0.48\linewidth]{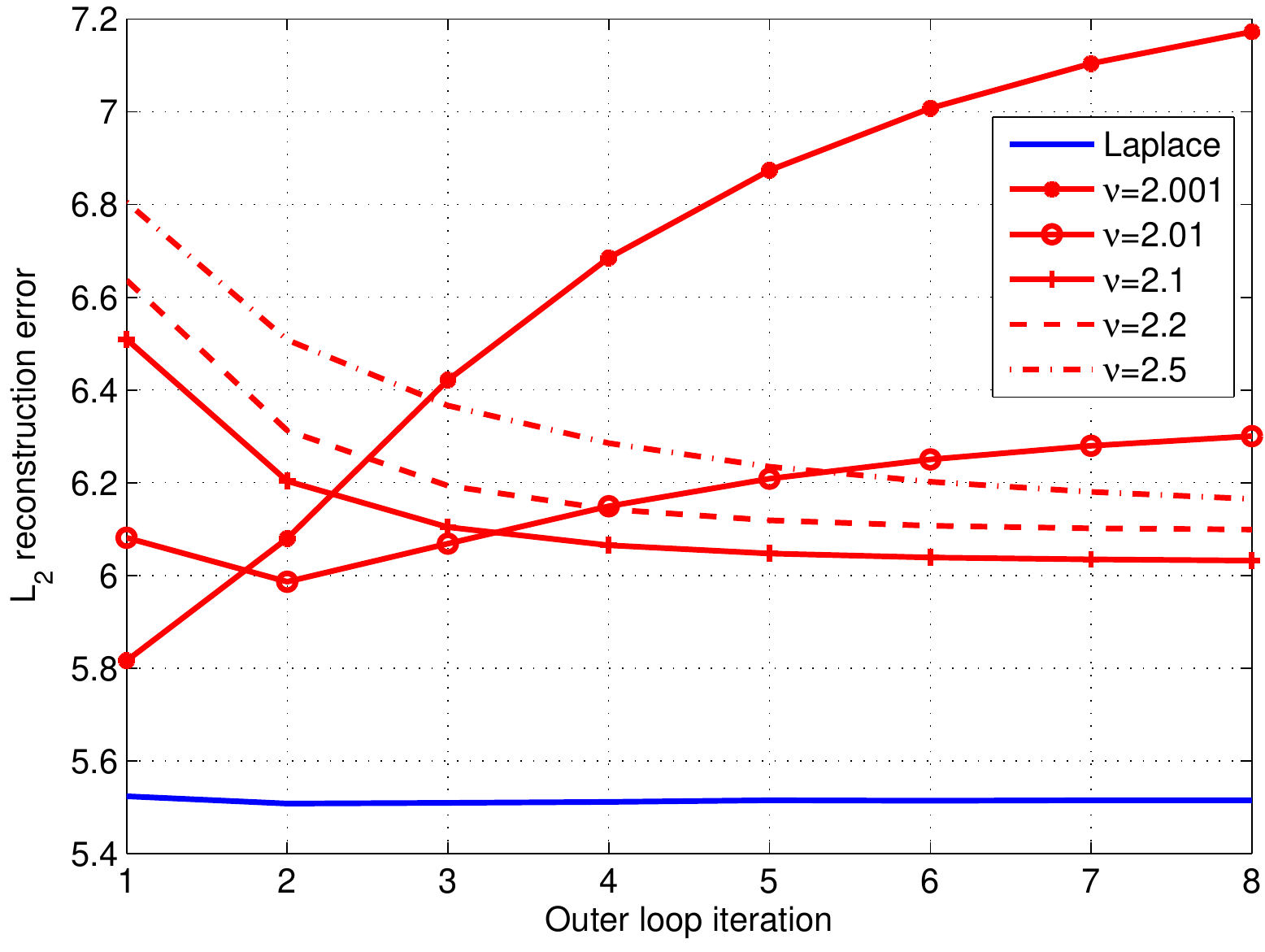} \\
  \includegraphics[width=0.23\linewidth]{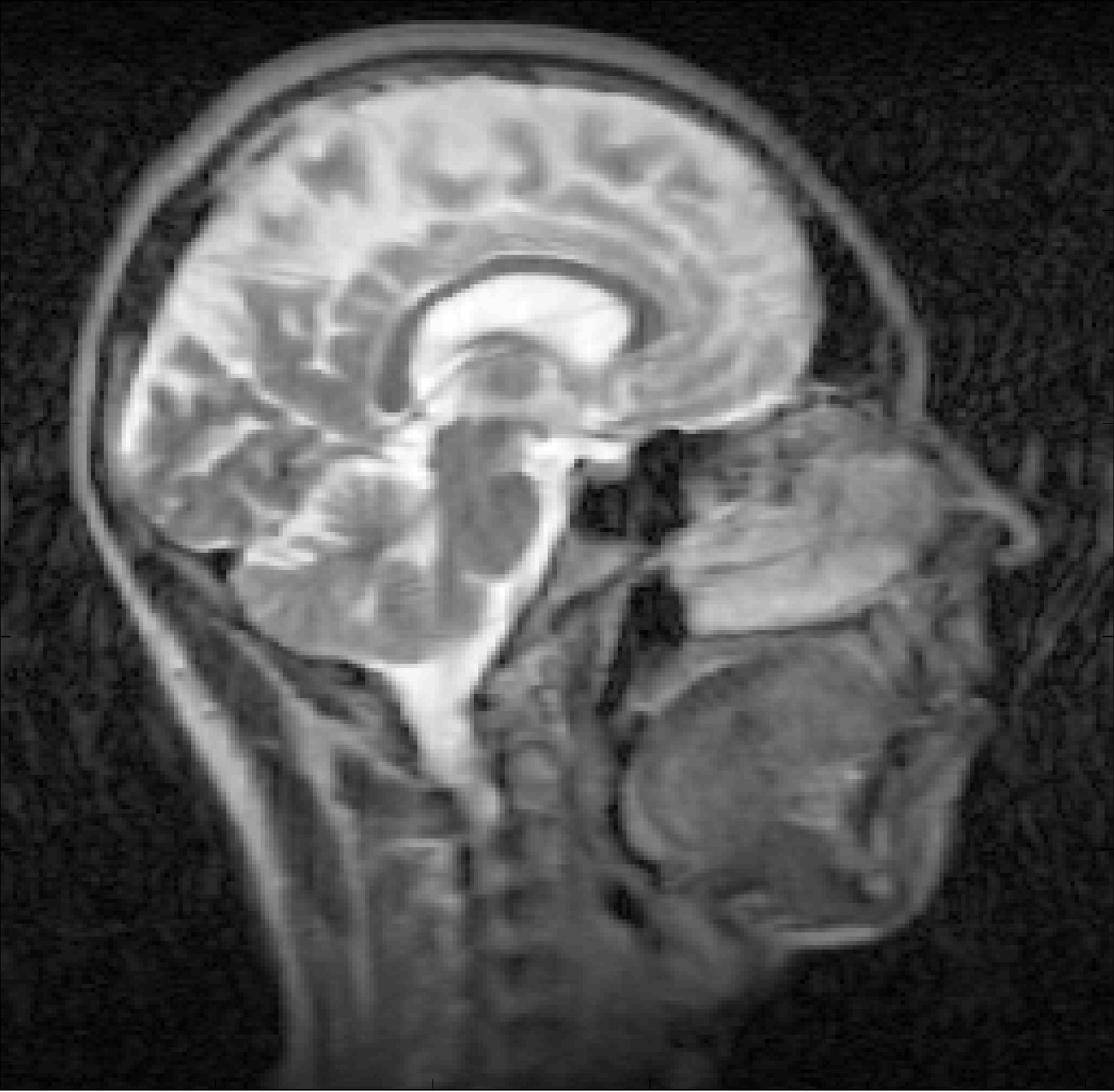} \hfill
  \includegraphics[width=0.23\linewidth]{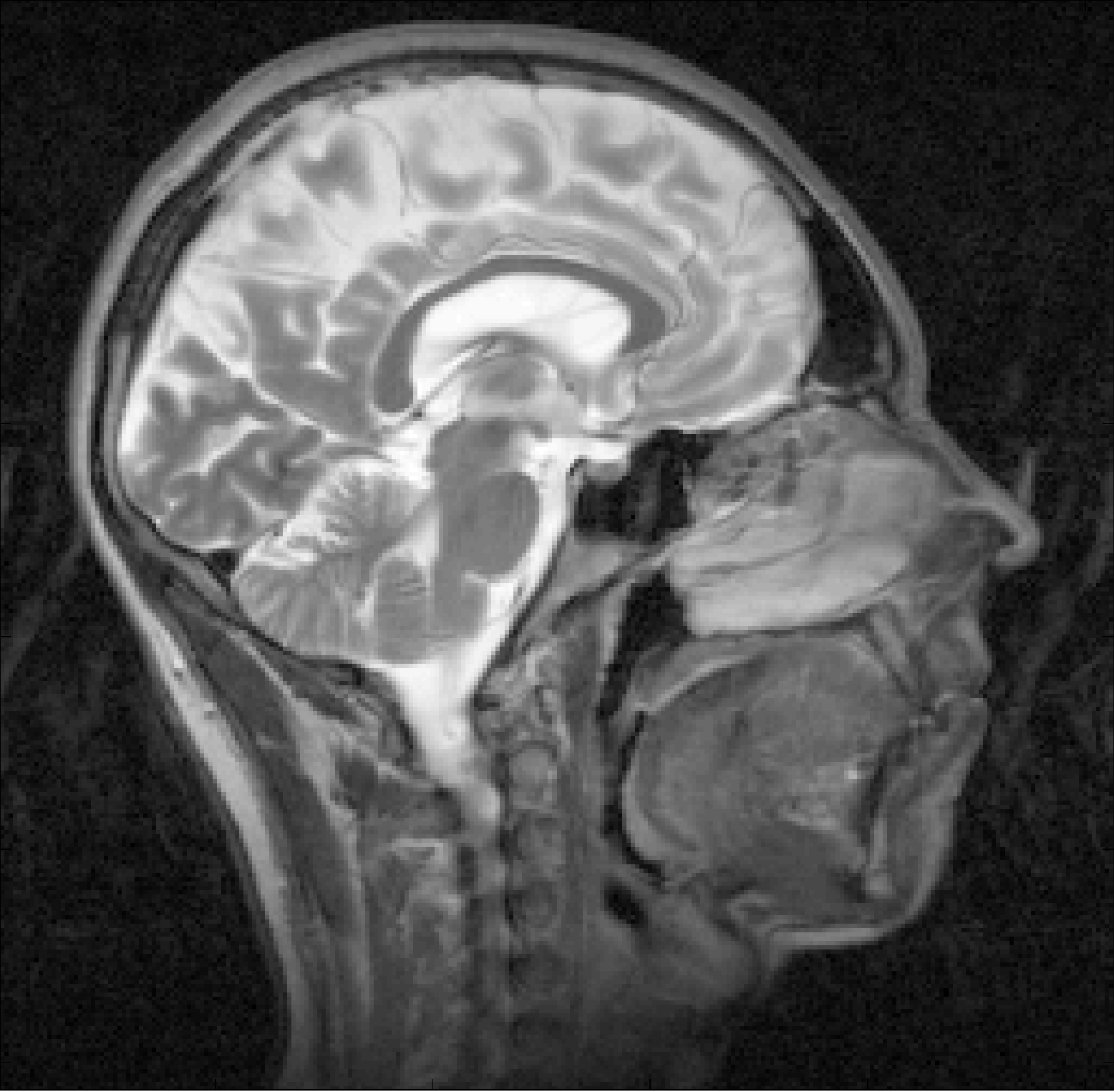} \hfill
  \includegraphics[width=0.23\linewidth]{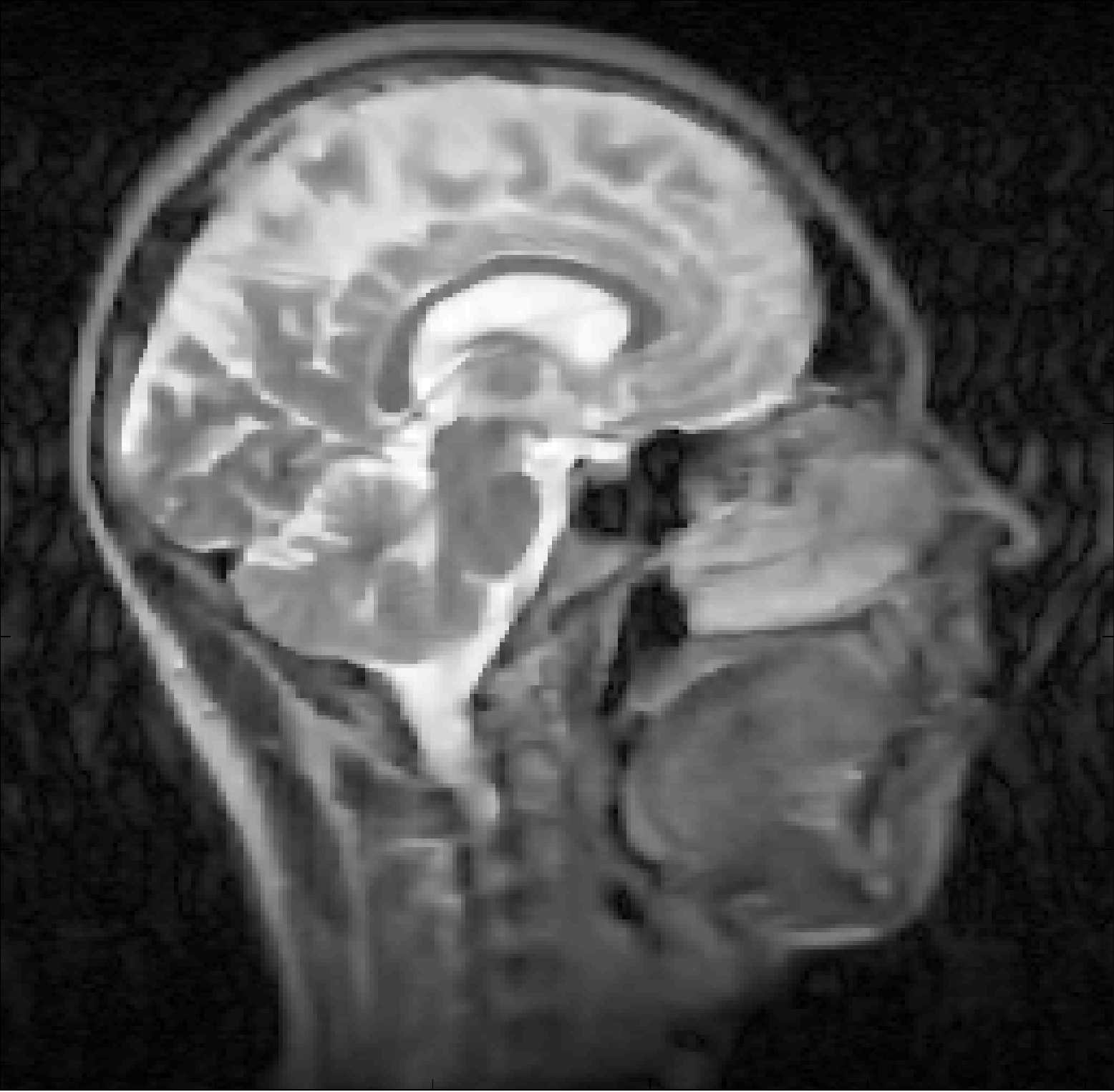} \hfill
  \includegraphics[width=0.23\linewidth]{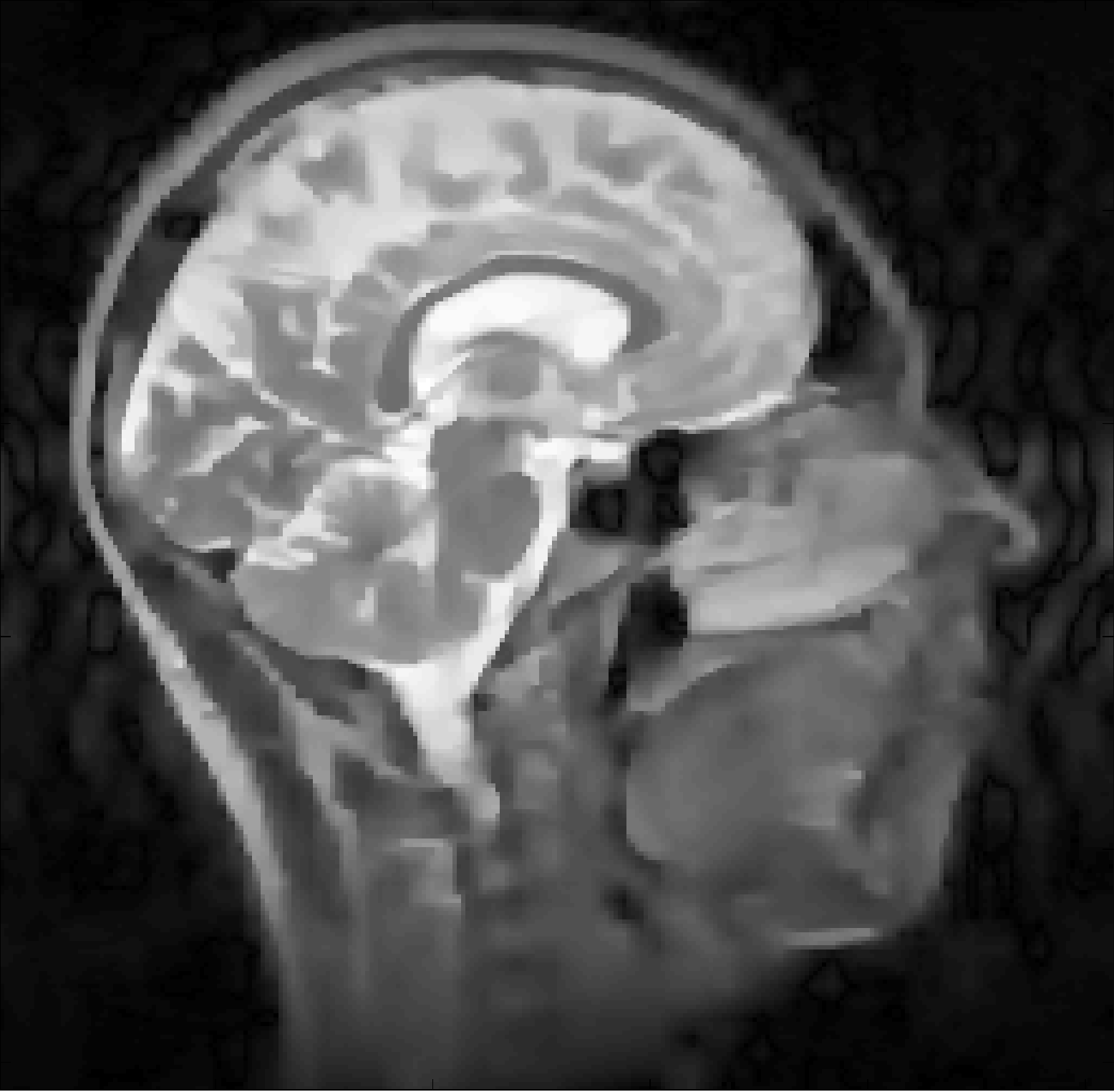}
\caption{\label{fig:exp-t}\small
  Comparison of SLM with Student's t and Laplace potentials (type A bounding).
  Shown are potential density functions (upper left), $\ell_2$
  errors of posterior mean estimates (upper right), over 8 outer loop
  iterations. Lower row: Posterior mean reconstructions $(|u_{*,i}|)$ after
  8 OL iterations: $\nu=2.1$; ground truth; $\nu=2.01$; $\nu=2.001$.}
\end{figure}

In this section, we compare SLM variational inference with Student's t
\eqp{student-t} potentials to the Laplace setup of \secref{exp-classes}.
Student's t potentials are not log-concave, so neither MAP estimation nor
variational inference are convex problems. Student's t potentials enforce
sparsity more strongly than Laplacians do, which is often claimed to be more
useful for image reconstruction. Their parameters are $\nu$ (degrees of
freedom; regulating sparsity) and $\alpha=\nu/\tau$ (scale). We compare
Laplace and Student's t potentials of same variance (the latter has a variance
for $\nu>2$ only): $\alpha_a=2(\nu-2)/\tau_a^2$, where $\tau_a$ is the
Laplace parameter, $\alpha_r$, $\alpha_i$ respectively. The model setup is the
same as in \secref{exp-classes}, using slice 8 of {\tt sg88} only. Result are
given in \figref{exp-t}.

Compared to the Laplace setup, reconstruction errors for Student's t SLMs are
worse across all values of $\nu$. While $\nu=2.1$ outperforms larger values,
the reconstruction error {\em grows} with iterations for $\nu=2.01$,
$\nu=2.001$. This is not a problem of sluggish convergence: $\phi(\vgamma{})$
decreases rapidly\footnote{
  For Student's t potentials (as opposed to Laplacians), type A and type
  B bounding behave very similar in these experiments.}
in this case. A glance at the mean reconstructions $(|u_{*,i}|)$
(\figref{exp-t}, lower row) indicates what happens. For $\nu=2.01,2.001$, image
sparsity is clearly enforced {\em too strongly}, leading to fine features being
smoothed out. The reconstruction for $\nu=2.001$ is merely a caricature of the
real image complexity, and rather useless as the output of a medical imaging
procedure. When it comes to {\em real-world image} reconstruction, more sparsity
does not necessarily lead to better results.

\subsection{Inaccurate Lanczos Variance Estimates}
\label{sec:exp-estim-vars}

\begin{figure}[ht]
\centering
  \raisebox{1ex}{\includegraphics[width=0.25\linewidth]{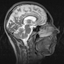}} \hfill
  \includegraphics[width=0.36\linewidth]{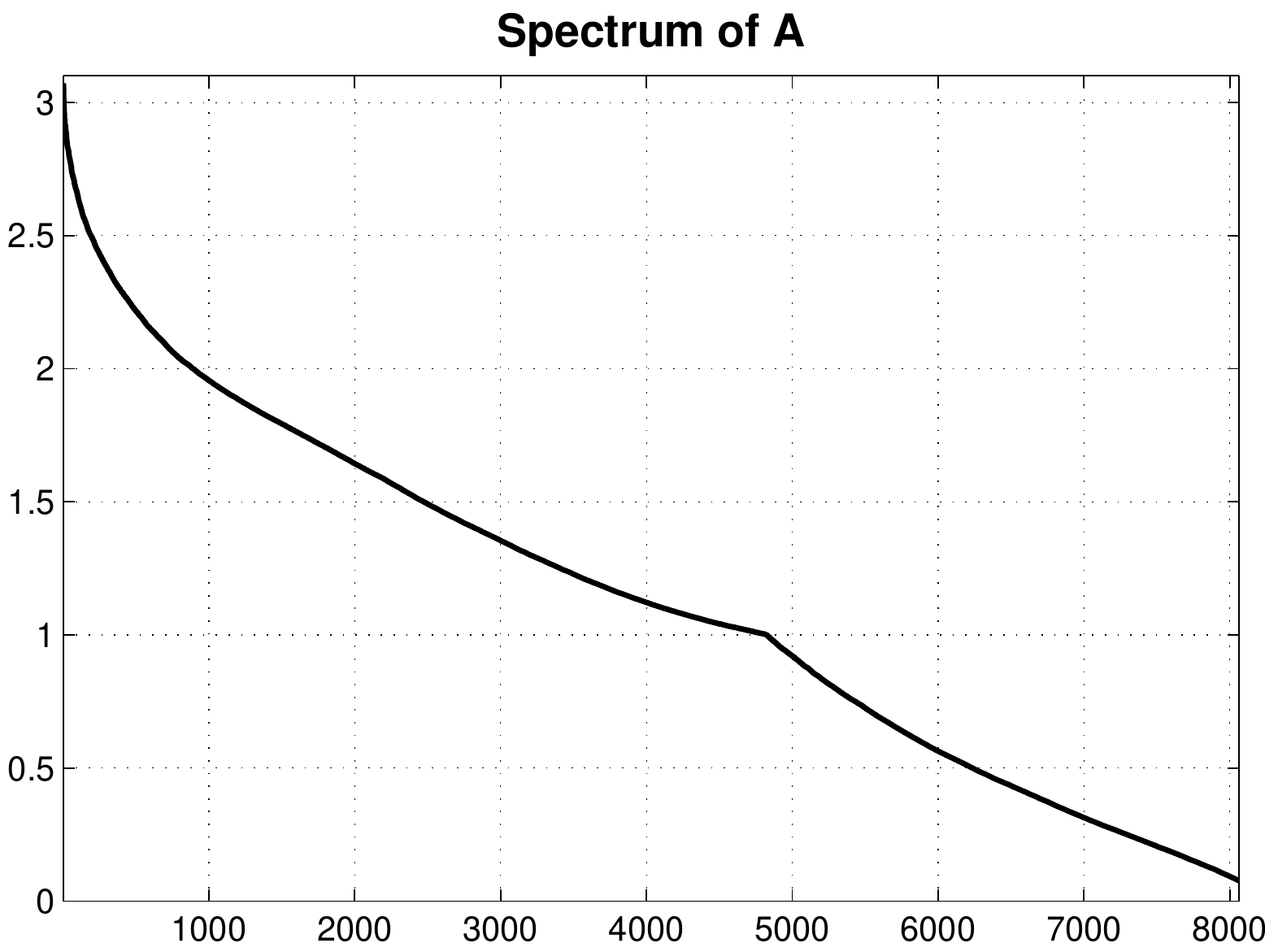} \hfill
  \includegraphics[width=0.36\linewidth]{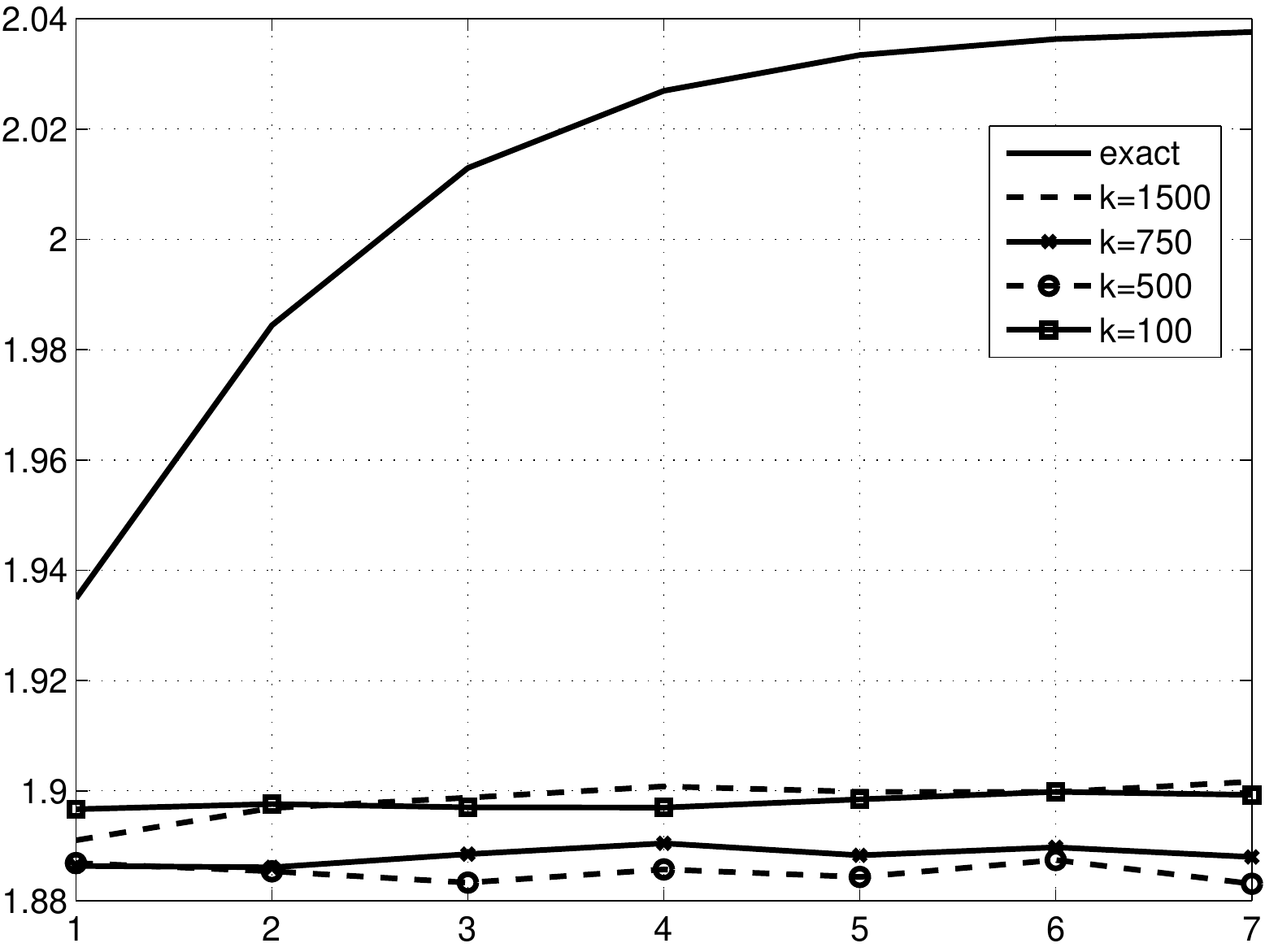} \\
  \ifthenelse{\equal{\figureres}{JPG}}{
    \includegraphics[width=0.24\linewidth]{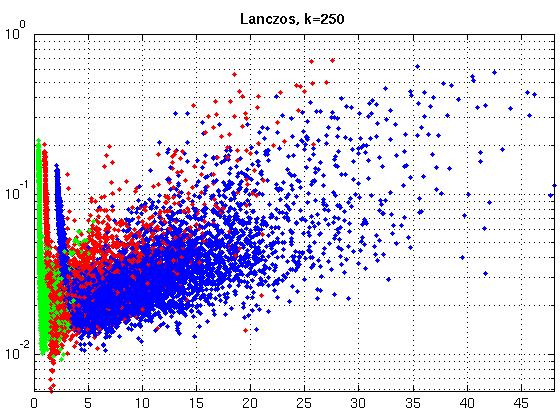} \hfill
    \includegraphics[width=0.24\linewidth]{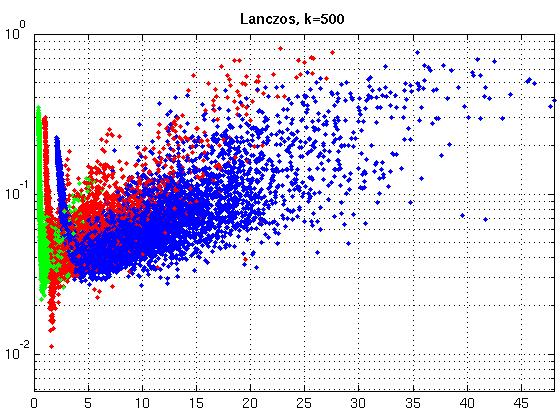} \hfill
    \includegraphics[width=0.24\linewidth]{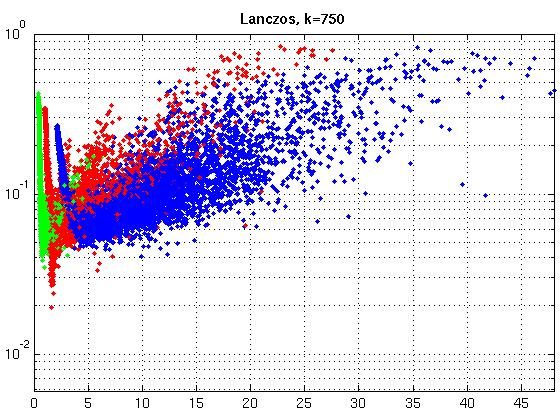} \hfill
    \includegraphics[width=0.24\linewidth]{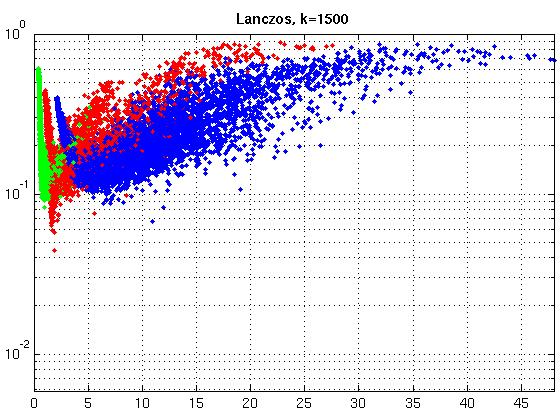} \\
    \includegraphics[width=0.325\linewidth]{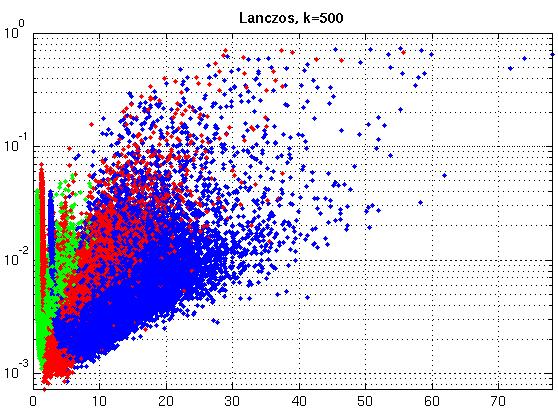} \hfill
    \includegraphics[width=0.325\linewidth]{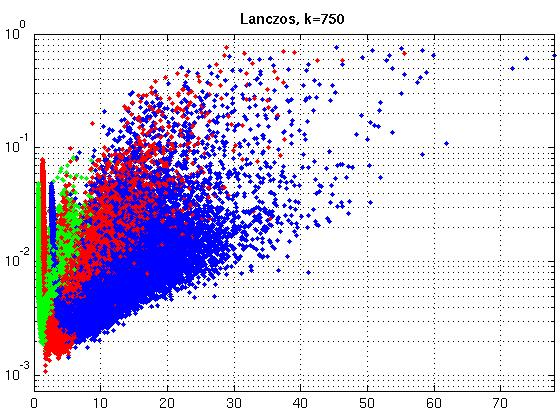} \hfill
    \includegraphics[width=0.325\linewidth]{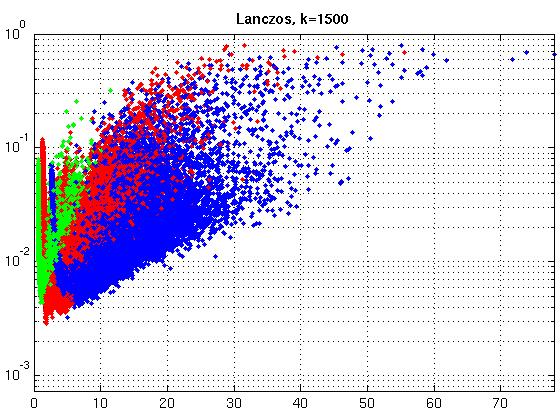}
  }{
    \includegraphics[width=0.24\linewidth]{img/slmvar-lanc-cart-zerr250} \hfill
    \includegraphics[width=0.24\linewidth]{img/slmvar-lanc-cart-zerr500} \hfill
    \includegraphics[width=0.24\linewidth]{img/slmvar-lanc-cart-zerr750} \hfill
    \includegraphics[width=0.24\linewidth]{img/slmvar-lanc-cart-zerr1500} \\
    \includegraphics[width=0.325\linewidth]{img/slmvar-lanc-cart-lgzerr500} \hfill
    \includegraphics[width=0.325\linewidth]{img/slmvar-lanc-cart-lgzerr750} \hfill
    \includegraphics[width=0.325\linewidth]{img/slmvar-lanc-cart-lgzerr1500}
  }
\caption{\label{fig:exp-vars}\small
  Lanczos approximations of Gaussian variances, at beginning of second OL
  iteration. For $64\times 64$ data (upper left), spectral decay of inverse
  covariance matrix $\mxa{}$ is roughly linear (upper middle). $\ell_2$
  reconstruction error of posterior mean estimate after subsequent OL
  iterations, for exact variance computation vs.\ $k=250,500,750,1500$ Lanczos
  steps (upper right).
  Middle row: Relative accuracy $\hat{z}_i\mapsto \hat{z}_{k,i}/\hat{z}_i$ at
  beginning of second OL iteration, separately for ``a'' potentials (on wavelet
  coefficients; red), ``r'' potentials (on derivatives; blue), and ``i''
  potentials (on $\Im(\vu{})$; green), see \secref{exp-mri}.
  Lower row: Relative accuracy $\hat{z}_i\mapsto \hat{z}_{k,i}/\hat{z}_i$ at
  beginning of second OL iteration, for full size setup ($256\times 256$),
  $k=500,750,1500$ (ground truth $\hat{z}_i$ determined by separate LCG runs).
}
\end{figure}

The difficulty of large scale Gaussian variance approximation is discussed in
\secref{estim-vars}. In this section, we analyse errors of the Lanczos variance
approximation we employ in our experiments. We downsampled our MRI data to
$64\times 64$, to allow for ground truth exact variance computations. The setup
is the same as above (Laplacians, type A bounding), with $\mxx{}$ consisting
of 30 phase encodes. Starting with a single common OL iteration, we compare
different ways of updating $\vz{1}$: exact variance computations
versus Lanczos approximations of different size $k$. Results are given in
\figref{exp-vars} (upper and middle row).

The spectrum of $\mxa{}$ at the beginning of the second OL iteration shows a
roughly linear decay. Lanczos approximation errors are rather large (middle
row). Interestingly, the algorithm does certainly not work better with exact
variance computations (judged by the development of posterior mean
reconstruction errors, upper right). We offer a heuristical explanation in
\secref{estim-vars}. A clear structure in the relative errors emerges from
the middle row: the largest (and also smallest) true values $\hat{z}_i$ are
approximated rather accurately, while smaller true entries are strongly
damped. The role of sparsity potentials $t_i(s_i)$, or of $\gamma_i$ within
the variational approximation, is to shrink coefficients selectively. The
structure of Lanczos variance errors serves to strengthen this effect.
We repeated the relative error estimation for the full-scale setup used in
the previous sections and below ($256\times 256$), ground truth values
$\hat{z}_i$ were obtained by separate conjugate gradients runs.
The results (shown in the lower row) exhibit the same structure, although
relative errors are larger in general.

Both our experiments and our heuristic explanation are given for sparse linear
model inference, we do not expect them to generalize to other models. Within
the same model and problem class, the impact of Lanczos approximations on
final design outcomes is analyzed in \cite{Seeger:10a}. As noted in
\secref{estim-vars}, understanding the real impact of Lanczos (or PCA)
approximations on approximate inference and decision-making is an important
topic for future research.

\subsection{Sampling Optimization for Magnetic Resonance Imaging}
\label{sec:exp-mri}

Magnetic resonance imaging \cite{Wright:97} is among the most important
medical imaging modalities. Without applying any harmful ionizing radiation,
a wide range of parameters, from basic anatomy to blood flow, brain
function or metabolite distribution, can be visualized. Image slices are
reconstructed from coefficients sampled along smooth trajectories in Fourier
space ({\em phase encodes}). In {\em Cartesian MRI}, phase encodes are dense
columns or rows in discrete Fourier space. The most serious limiting
factor\footnote{
  Patient movement (blood flow, heartbeat, thorax) is strongly detrimental to
  image quality, which necessitates uncomfortable measures such as breath-hold
  or fixation. In dynamic MRI, temporal resolution is limited by scan time.
}
is long scan time, which is proportional to the number of phase encodes
acquired. MRI is a prime candidate for compressive sensing (\secref{compsens})
in practice \cite{Lustig:07,Seeger:08c}: if images of diagnostic quality can be
reconstructed from an undersampled design, time is saved at no additional
hardware costs or risks to patients.

In this section, we address the problem of MRI sampling optimization:
which smallest subset of phase encodes results in MAP reconstructions of
useful quality? To be clear, we do not use approximate Bayesian technology
to improve reconstruction from fixed designs (see \secref{map-estim}), but
aim to optimize the design $\mxx{}$ itself, so to best support subsequent
standard MAP reconstruction on real-world images. As discussed in
\secref{map-estim} and \secref{compsens}, the focus for most work on
compressive sensing is on the reconstruction algorithm, the question of how
to choose $\mxx{}$ is typically not addressed (exceptions include
\cite{Ji:07,Haupt:08}).
We follow the {\em adaptive} Bayesian experimental design
scenario described in \secref{design}, where $\{\mxx{*}\}$ are phase encodes
(columns in Fourier space), $\vu{}$ the unknown (complex-valued) image.
Implementing
this proposal requires the approximation of dominating posterior covariance
directions for a large scale non-Gaussian SLM ($n=131072$), which to our
knowledge has not been
attempted before. Results shown below are part of a larger study
\cite{Seeger:08c} on human brain data acquired with a Siemens 3T scanner (TSE,
23 echos$/$exc, $120^\circ$ refocusing pulses, $1\times 1\times 4\,
\mathrm{mm}^3$ voxels, resolution $256\times 256$). Note that for Nyquist
dense acquisition, resolution is dictated by the number of phase encodes,
$256$ in this setting. We employ two datasets {\tt sg92} and
{\tt sg88} here (sagittal orientation, echo time $\approx$90ms).

\begin{figure}[ht!]
\includegraphics[width=\columnwidth]{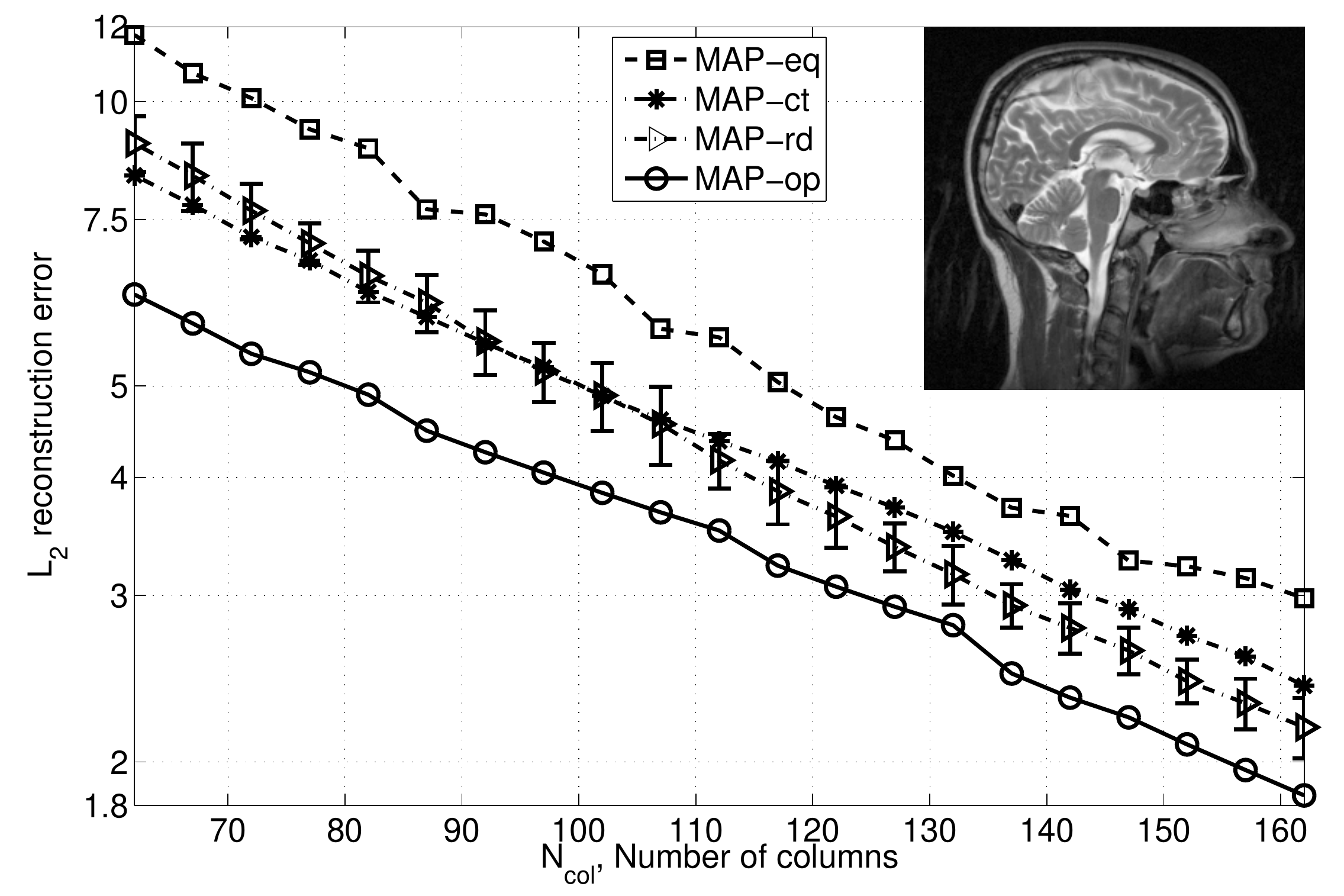}
\caption{\label{fig:mri-results}
  Results for Cartesian undersampling with different measurement designs,
  on sagittal slice (TSE, TE=92ms). All designs contain 32 central columns.
  Equispaced [eq]; low-pass [ct]; random with variable density [rd] (averaged
  over 10 repetitions); optimized by our Bayesian technique [op]. Shown are
  $\ell_2$ distances to $\boldsymbol{u}_{\text{true}}$ for MAP
  reconstruction with the Laplace SLM. Designs optimized on the same data.}
\end{figure}

We use a sparse linear model with Laplace potentials \eqp{laplace}. In MRI,
$\vu{}$, $\vy{}$, $\mxx{}$, and $\vs{}=\mxb{}\vu{}$ are naturally
complex-valued, and
we make use of the group potential extension discussed in \secref{group-pots}
(coding $\C$ as $\R^2$).
The vector $\vs{}$ is composed of multi-scale wavelet coefficients $\vs{a}$,
first derivatives (horizontal and vertical) $\vs{r}$, and the
imaginary
part $\vs{i}=\Im(\vu{})$. A matrix-vector multiplication (MVM) with $\mxx{}$
requires a fast Fourier
transform (FFT), while an MVM with $\mxb{}$ costs $O(n)$ only. Laplace
scale parameters were $\tau_a=0.07$, $\tau_r=0.04$, $\tau_i=0.1$). The
algorithms described above were run with $n=131072$, $q=261632$, candidate
size $d=512$, and $m=d\cdot N_{\text{col}}$, where $N_{\text{col}}$ is the
number of phase encodes in $\mxx{}$.
We compare different ways of constructing designs $\mxx{}$, all of which
start with the central 32 columns (lowest horizontal frequencies): Bayesian
sequential optimization, with all remaining 224 columns as candidates
({\tt op}); filling the grid from the center outwards ({\tt ct}; such low-pass
designs are typically used with {\em linear} MRI reconstruction); covering the
grid with equidistant columns ({\tt eq}); and drawing encodes at random
(without replacement), using the variable-density sampling approach of
\cite{Lustig:07} ({\tt rd}). The latter is motivated by compressive
sensing theory (see \secref{compsens}), yet is substantially refined compared
to naive i.i.d.\ sampling.\footnote{
  Results for drawing {\em phase encodes} uniformly at random are much worse
  than the alternatives show, even if started with the same central 32
  columns. Reconstructions become even worse when Fourier {\em coefficients}
  are drawn uniformly at random.
}
Results for sparse MAP reconstruction of the most difficult slice in
{\tt sg92} are shown in \figref{mri-results} (the error metric is
$\ell_2$ distance $\||\vu{*}|-|\vu{\text{true}}|\|$, where
$\vu{\text{true}}$ is the complete data reconstruction).

Obtained with the same standard sparse reconstruction method
(convex $\ell_1$ MAP estimation), results for fixed $N_{\text{col}}$ differ
``only'' in terms of the composition of $\mxx{}$ (recall that scan time grows
proportional to $N_{\text{col}}$). Designs chosen by our Bayesian technique
substantially outperform all other choices. These results, along with
\cite{Seeger:08,Seeger:08c}, are in stark contrast to claims that independent
random sampling is a good way to choose designs for sub-Nyquist reconstruction
{\em of real-world images}. The improvement of Bayesian optimized over
randomly drawn designs is larger for smaller $N_{\text{col}}$. In fact,
variable-density sampling does worse than conventional low-pass designs below
$1/2$ Nyquist. Similar findings are obtained
in \cite{Seeger:08} for different natural images. In the regime far
below the Nyquist limit, it is all the more important to judiciously optimize
the design, using criteria informed about realistic images in the first place.

A larger range of results is given in \cite{Seeger:08c}. Even at $1/4$
Nyquist, designs optimized by our method lead to images where most relevant
details are preserved. In \figref{mri-results}, testing and design optimization
is done on the same dataset. The generalization capability of our optimized
designs is tested in this larger study, applying them to a range of data from
different subjects, different contrasts, and different orientations,
achieving improvements on these test sets comparable to what is shown in
\figref{mri-results}.
Finally, we have concentrated on single image slice optimization in our
experiments. In realistic MRI experiments, a number of neighbouring slices
is acquired in an interleaved fashion. Strong statistical dependencies between
slices can be exploited, both in reconstruction and joint design optimization,
by combining our framework with structured graphical model message passing
\cite{Seeger:09a}.

\section{Discussion}
\label{sec:discuss}

In this paper, we introduce scalable algorithms for approximate {\em Bayesian
inference} in sparse linear models, complementing the large body of
work on {\em point estimation} for these models. If the Bayesian posterior
is not just taken for a criterion to be optimized, but as global picture of
uncertainty in a reconstruction problem, advanced decision-making problems
such as model calibration, feature relevance ranking or Bayesian experimental
design can be addressed.
We settle a long-standing question for continuous-variable variational
Bayesian inference, proving that the relaxation of interest here
\cite{Jaakkola:97a,Palmer:06,Girolami:01} has the same convexity profile than
MAP estimation. Our double loop algorithms are scalable by reduction
to common computational problems, penalized least squares optimization and
Gaussian covariance estimation (or principal components analysis). The large
and growing body of work for the latter, both in theory and algorithms, is put
to novel use in our methods. Moreover, the reductions offer valuable insight
into similarities and differences between sparse estimation and approximate
Bayesian inference, as do our focus on decision-making problems beyond point
reconstruction.

We apply our algorithms to the design optimization problem of improving
sampling trajectories for magnetic resonance imaging. To the best of our
knowledge, this has not been attempted before in the context of sparse
nonlinear reconstruction. Ours is the first approximate Bayesian framework for
adaptive compressive sensing that scales up to and succeeds on full
high-resolution real-world images. Results here are part of a larger MRI
study \cite{Seeger:08c}, where designs optimized by our Bayesian technique
are found to significantly and robustly improve sparse image reconstruction
on a wide range of test datasets, for measurements far below the Nyquist
limit.

In future work, we will address advanced joint design scenarios, such as MRI
sampling optimization for multiple image slices, 3D MRI, and parallel MRI
with array coils. Our technique can be sped up along many directions, from
algorithmic improvements (advanced algorithms for inner loop optimization,
modern Lanczos variants) down to parallel computation on graphics hardware.
An important future goal, currently out of reach, is supporting real-time
MRI applications by automatic on-line sampling optimization.

\subsubsection*{Acknowledgments}

The MRI application is joint work with Rolf Pohmann and Bernhard Sch\"{o}lkopf,
MPI for Biological Cybernetics, T\"ubingen. We thank Florian Steinke and David
Wipf for helpful discussions and input. Supported in part by the IST Programme
of the
European Community, under the PASCAL Network of Excellence, IST-2002-506778,
and the Excellence Initiative of the German research foundation (DFG).


\begin{appendix}

\section{Details and Proofs}
\label{app:proofs}
%

\subsection{Proof of \thref{ldeta-convex}}
\label{app:ldeta-convex}

In this section, we provide a proof of \thref{ldeta-convex}.
For notational convenience, we absorb $\sigma^{-2}$ into $\mxx{}^T\mxx{}$, by
replacing $\mxx{}$ by $\sigma^{-1}\mxx{}$.
We begin with part (\ref{th1-2}). It is well known that
$\vpi{}\mapsto \log|\tilde{\mxa{}}(\vpi{})|$ is concave and nondecreasing
for $\vpi{}\succ\vzero$ \cite[Sect.~3.1.5]{Boyd:02}. Both properties carry over
to the extended-value function.\footnote{
  In general, we extend convex continuous functions $f(\vpi{})$ on
  $\vpi{}\succ\vzero$ by $f(\vpi{}) = \lim_{\vd{}\searrow\vpi{}} f(\vd{})$,
  $\vpi{}\succeq\vzero$, and $f(\vpi{})=\infty$ elsewhere.}
The statement follows from the concatenation rules of
\cite[Sect.~3.2.4]{Boyd:02}.

We continue with part (\ref{th1-1}). Write $\tilde{\mxa{}} =
\tilde{\mxa{}}(f(\vgamma{}))$, $\psi_1 := \log|\tilde{\mxa{}}|$,
$\mxgamma{}=\diag\vgamma{}$, and $f(\mxgamma{})=\diag f(\vgamma{})$. First,
$\vgamma{}\mapsto \psi_1$ is the composition of twice continuously
differentiable mappings, thus inherits this property. Now, $d\psi_1 =
\trace\mxs{} f'(\mxgamma{}) (d\mxgamma{})$, where $\mxs{} :=
\mxb{}\tilde{\mxa{}}^{-1}\mxb{}^T$, moreover $d^2\psi_1 =
-\trace\mxs{} f'(\mxgamma{}) (d\mxgamma{})\mxs{} f'(\mxgamma{}) (d\mxgamma{})
+ \trace\mxs{} f''(\mxgamma{}) (d\mxgamma{})^2 = \trace (d\mxgamma{}) \mxs{}
(d\mxgamma{}) \mxe{1}$, where $\mxe{1}{} := f''(\mxgamma{}) -  f'(\mxgamma{})
\mxs{} f'(\mxgamma{})$. Since $\mxs{}\succeq\mxzero$, we have $\mxs{} =
\mxv{}\mxv{}^T$ for some matrix $\mxv{}$, and $d^2\psi_1 = \trace
((d\mxgamma{})\mxv{})^T \mxe{1}{} (d\mxgamma{})\mxv{}$. Now, if
$\mxe{1}{}\succeq\mxzero$, then for $\vgamma[t]{} = \vgamma{} +
t (\Delta\vgamma{})$, we have $\psi_1''(0) = \trace\mxn{}^T\mxe{1}{}\mxn{}\ge
0$ for any $\Delta\vgamma{}$, where $\mxn{} := (\diag\Delta\vgamma{})\mxv{}$,
so that $\psi_1$ is convex.

The log-convexity of $f_i(\gamma_i)$ implies that $f_i(\gamma_i)
f_i''(\gamma_i) \ge (f_i'(\gamma_i))^2$ for all $\gamma_i$, so that
\[
\begin{split}
  \mxe{1}{} & = f(\mxgamma{})^{-1} (f(\mxgamma{}) f''(\mxgamma{})) -
  f'(\mxgamma{}) \mxs{} f'(\mxgamma{}) \succeq f(\mxgamma{})^{-1}
  (f'(\mxgamma{}))^2 - f'(\mxgamma{}) \mxs{} f'(\mxgamma{}) \\
  & = f'(\mxgamma{}) \left( f(\mxgamma{})^{-1} - \mxs{} \right) f'(\mxgamma{}).
\end{split}
\]
Therefore, it remains to be shown that $f(\mxgamma{})^{-1}-\mxs{}\succeq
\mxzero$. We use the identity
\begin{equation}\label{eq:max-quad}
  \vv{}^T\mxm{}^{-1}\vv{} = \max_{\vx{}} 2\vv{}^T\vx{} - \vx{}^T\mxm{}\vx{},
\end{equation}
which holds whenever $\mxm{}\succ\mxzero$. For any $\vr{}\in \R^q$:
\[
  \vr{}^T\mxb{}\tilde{\mxa{}}^{-1}\mxb{}^T\vr{} = \max_{\vx{}} 2 \vr{}^T\mxb{}\vx{} -
  \vx{}^T \left( \mxx{}^T\mxx{} + \mxb{}^T f(\mxgamma{})\mxb{} \right)\vx{}
  \le \max_{\vk{}=\mxb{}\vx{}} 2\vr{}^T\vk{} - \vk{}^T f(\mxgamma{})\vk{},
\]
using \eqp{max-quad} and $\vx{}^T\mxx{}^T\mxx{}\vx{} = \|\mxx{}\vx{}\|^2\ge 0$.
Therefore,
$\vr{}^T\mxs{}\vr{} \le \max_{\vk{}} 2\vr{}^T\vk{} - \vk{}^T f(\mxgamma{})\vk{}
= \vr{}^T f(\mxgamma{})^{-1}\vr{}$,
using \eqp{max-quad} once more, which implies $f(\mxgamma{})^{-1}-\mxs{}\succeq
\mxzero$. This completes the proof of part (\ref{th1-1}). Since $\mxa{} =
\tmxa{}(\vgamma{}^{-1})$, we can employ this argument with $f_i(\gamma_i) =
\gamma_i^{-1}$ and $\vr{}=\vdelta{i}$ in order to establish part (\ref{th1-4}).

We continue with part (\ref{th1-3}). Write $\tilde{\mxa{}} =
\tilde{\mxa{}}(f(\vgamma{})^{-1})$ and $\psi_2 := \vone^T(\log f(\vgamma{})) +
\log|\tilde{\mxa{}}|$. Assume for now that $\mxx{}^T\mxx{}\succ\mxzero$.
Let $\mxb{} = (\mxb{<q}^T\, \tvb{})^T$ (so that $\tvb{}^T$ is the last row of
$\mxb{}$), and define $\tilde{\mxa{}}_{<q} = \mxx{}^T\mxx{} +
\mxb{<q}^T f(\mxgamma{<q})^{-1}\mxb{<q}$, where $f(\vgamma{<q}) =
(f_i(\gamma_i))_{i<q}\in \R^{q-1}_+$. We make use of the well-known determinant
identity $|\Id + \vv{}\vv{}^T| = 1 + \vv{}^T\vv{}$. Namely,
\begin{equation}\label{eq:det-ident}
\begin{split}
  & \log f_q(\gamma_q) + \log\left|\tmxa{<q} + f_q(\gamma_q)^{-1}
  \tvb{}\tvb{}^T\right| \\
  =\, & \log f_q(\gamma_q) + \log\left|\tmxa{<q}\right| + \log\left|
  \Id + f_q(\gamma_q)^{-1} (\tmxa{<q})^{-1}\tvb{}\tvb{}^T \right| \\
  =\, & \log\left|\tmxa{<q}\right| + \log f_q(\gamma_q) +
  \log\left( 1 + f_q(\gamma_q)^{-1}\tvb{}^T(\tmxa{<q})^{-1}\tvb{}
  \right) \\
  =\, & \log\left|\tmxa{<q}\right| +
  \log\left( f_q(\gamma_q) + \tvb{}^T(\tmxa{<q})^{-1}\tvb{} \right).
\end{split}
\end{equation}
Since the extended-value function $\log(\cdot)$ (assigning $-\infty$ to
arguments $\le 0$) is concave and nondecreasing, the concatenation rules of
\cite[Sect.~3.2.4]{Boyd:02} imply the concavity of the final term in
\eqp{det-ident} whenever $f_q(\gamma_q) + \tvb{}^T(\tmxa{<q})^{-1}\tvb{}$ is
concave.
We will use induction on $q$, the dimensionality of $\vgamma{}$. For $q=1$,
$\psi_2$ is given by \eqp{det-ident} with $\tmxa{<1} = \mxx{}^T\mxx{}$,
and its concavity follows from the concavity of $f_1(\gamma_1)$.
For $q>1$, \eqp{det-ident} implies
\[
  \psi_2 = \vone^T(\log f(\vgamma{<q})) + \log|\tilde{\mxa{}}_{<q}| +
  \log\left( f_q(\gamma_q) + \tvb{}^T(\tmxa{<q})^{-1}\tvb{} \right).
\]
Both the sum of the first two terms and $f_q(\gamma_q)$ are concave by
assumption, so that the concavity of $\psi_2$ is implied by the concavity of
$\vgamma{}\mapsto \tvb{}^T(\tmxa{<q})^{-1}\tvb{}$. Using
\eqp{max-quad}, we have
\[
  \tvb{}^T(\tmxa{<q})^{-1}\tvb{} = \max_{\vx{}} 2
  \tvb{}^T \vx{} - \vx{}^T\tilde{\mxa{}}_{<q}\vx{} = \max_{\vx{}}
  2\tvb{}^T\vx{} - \|\mxx{}\vx{}\|^2 - \vv{}^Tf(\mxgamma{<q})^{-1}\vv{}
\]
with $\vv{} := \mxb{<q}\vx{}$. Now, $(\vx{},\vf{})\mapsto 2\tvb{}^T\vx{} -
\|\mxx{}\vx{}\|^2 - \vv{}^T(\diag\vf{})^{-1}\vv{}$ is jointly concave for
$\vf{}\succ\vzero$ (see proof of \thref{crit-convex}), so that
$\kappa(\vf{}) := \tvb{}^T\tmxa{<q}(\vf{}^{-1})^{-1}\tvb{}$ is concave for
$\vf{}\succ\vzero$ \cite[Sect.~3.2.5]{Boyd:02} (recall that
$\tmxa{<q}(\vf{}^{-1}) = \mxx{}^T\mxx{} + \mxb{<q}^T(\diag\vf{}^{-1})\mxb{<q}$).
To finish the argument, we plug in $\vf{} := f(\vgamma{<q})$ and use the
concatenation theorems of \cite[Sect.~3.2.4]{Boyd:02}. What remains to be
shown in this context is that $\kappa(\vf{})$ is nondecreasing in each
argument. Pick any $i\in\srng{q}$, $\vf{}\succ\vzero$, and any $\Delta>0$.
Then,
\[
  \kappa(\vf{}+\Delta\vdelta{i}) = \tvb{}^T\left( \tmxa{<q}(\vf{}^{-1}) -
  \frac{\Delta}{f_i(f_i+\Delta)} \vb{i}\vb{i}^T \right)^{-1}\tvb{}
  \ge \tvb{}^T\tmxa{<q}(\vf{}^{-1})^{-1}\tvb{} = \kappa(\vf{}),
\]
where $\vb{i}=\mxb{}^T\vdelta{i}$.
This concludes the proof of part (\ref{th1-3}), under the assumption that
$\mxx{}^T\mxx{}$ is invertible.
If $\mxx{}^T\mxx{}$ is singular, define $\psi_2^{\eps}$ as above, but with
$\mxx{}^T\mxx{}\to \mxx{}^T\mxx{} + \eps\Id$. We saw that $\psi_2^{\eps}$ is
concave for any $\eps>0$. For any $\vgamma{}\succ\vzero$ such that
$\psi_2(\vgamma{})>-\infty$, $\psi_2^{\eps}$ converges uniformly to $\psi_2$ on
a closed environment of $\vgamma{}$ ($\psi_2$ and all $\psi_2^{\eps}$ are
continuous), so that $\psi_2$ is concave at $\vgamma{}$. This completes
the proof of part (\ref{th1-3}).

\subsection{Proof of \thref{convex-inf}}
\label{app:convex-inf}

In this section, we provide of proof of \thref{convex-inf}.
We begin with part (\ref{th2-1}), focussing on a single potential $i$ and
dropping its index. Since $b\ne 0$ is dealt with separately, assume that $t(s)$
is even: $\log t(s) = \tilde{g}(s) = g(s^2) = g(x)$, where $x=s^2$. If
$f(x,\gamma) := -x/\gamma - 2 g(x)$ and $\tilde{f}(s,\gamma) := -s^2/\gamma
- 2\tilde{g}(s)$, then $h(\gamma) = \max_{x\ge 0} f(x,\gamma)$, and
$f(x,\gamma) = \tilde{f}(x^{1/2},\gamma)$. It suffices to consider $s\ge 0$.
Denote $x_* = x_*(\gamma) := \argmax_{x\ge 0} f(x,\gamma)$ (unique, since
$g(x)$ is strictly convex). If $\gamma_0 := \sup\{\gamma\, |\, f(x,\gamma)\le
-2 g(0)\; \forall x\}$ ($\gamma_0=0$ for an empty set), then
\begin{itemize}
\item
  $x_*=0$, $h(\gamma)=-2 g(0)$ for $\gamma\in (0,\gamma_0]$
\item
  $x_*>0$, $h(\gamma)$ strictly increasing for $\gamma>\gamma_0$
\end{itemize}
Namely, if $\gamma_0<\gamma_1<\gamma_2$, then $x_*(\gamma_1)>0$ by definition
of $\gamma_0$, and $h(\gamma_1) = -x_*(\gamma_1)/\gamma_1 - 2 g(x_*(\gamma_1))
< -x_*(\gamma_1)/\gamma_2 - 2 g(x_*(\gamma_1)) \le -x_*(\gamma_2)/\gamma_2 -
2 g(x_*(\gamma_2)) = h(\gamma_2)$. Note that $\gamma_0>0$ iff
$\lim_{\eps\searrow 0} g'(\eps)$ is finite.
It suffices to show that $h$ is convex at all $\gamma>\gamma_0$, where
$x_*=s_*^2>0$.

We use the notation $\tilde{f}_s = \partial \tilde{f}/(\partial s)$, functions
are evaluated at $(x_*=s_*^2,\gamma)$ if nothing else is said. Now,
$\tilde{f}_s = -2 s_*/\gamma - 2 \tilde{g}_s(s_*) = 0$, so that
$\tilde{g}_s(s_*) = -s_*/\gamma$. Next, $g(x)$ is twice continuously
differentiable, and $x_*=s_*^2$ at $\gamma$. Therefore, $f_x =
\partial f/(\partial x)$ is continuously differentiable. Moreover,
$g_{x,x}(x)>0$ by the strict convexity of $g(x)$.
By the implicit function theorem, $x_*(\gamma)$ is continuously differentiable
at $\gamma$, and since $h(\gamma) = f(x_*(\gamma),\gamma)$, $h'(\gamma)$
exists. Moreover, $0 = (d/d\gamma) f_x(x_*(\gamma),\gamma) = f_{x,\gamma}
+ f_{x,x}\cdot(d x_*)/(d\gamma)$, so that $(d x_*)/(d\gamma) =
\gamma^{-2}/(2 g_{x,x}(x_*))>0$: $x_*(\gamma)$ is increasing.
From $f_x=0$, we have that $h'(\gamma) = f_{\gamma} = s_*^2/\gamma^2
= (\tilde{g}_s(s_*))^2$, since $\tilde{g}_s(s_*) = -s_*/\gamma$. Now,
$\tilde{g}_s(s)$ is nonincreasing by the concavity of $\tilde{g}(s)$, and
$\tilde{g}_s(s_*)<0$, so that
$s_*\mapsto h'(\gamma)$ is nondecreasing. Since $s_*^2=x_*$ is increasing in
$\gamma$, so is $s_*$. Therefore, $\gamma\mapsto h'(\gamma)$ is nondecreasing,
which means that $h(\gamma)$ is convex for $\gamma>\gamma_0$.

The concavity of $\tilde{g}(s)$ is necessary. Suppose that
$\tilde{g}_{s,s}(\tilde{s})>0$ for some $\tilde{s}>0$. If $\tilde{x} =
\tilde{s}^{1/2}$, $g(x)$ is differentiable at $\tilde{x}$, and if
$\tilde{\gamma} = -1/(2 g'(\tilde{x}))$, then $s_*(\tilde{\gamma})
= \tilde{s}$. But if $\tilde{g}_{s,s}(s_*)>0$ at $\tilde{\gamma}$, then
$s_*\mapsto h'(\gamma)$ is decreasing at $s_*=\tilde{s}$, and just as above
$\gamma\mapsto h'(\gamma)$ is decreasing at $\tilde{\gamma}$, so that $h$ is
not convex at $\tilde{\gamma}$. This concludes the proof of part (\ref{th2-1}).

Part (\ref{th2-2}) is a direct consequence of part (\ref{th2-1}) and
\thref{crit-convex}. For the final statement, suppose that $h_i'(\gamma_i)$ is
decreasing at $\gamma_i=\tilde{\gamma_i}$. Pick the other coefficients in
$\tvgamma{}\succ\vzero$ arbitrary, and choose $m=n=1$, $\vy{}=\vzero$,
$\mxx{}=X$, $\mxb{}=\vdelta{i}$, so that
$\phi(\vgamma{}) - h(\vgamma{}) = r(\gamma_i) := \log(1 + X^2\gamma_i) -
\log\gamma_i$,
ignoring additive constants. Consider $\tilde{\phi}(t) = \phi{}(\tvgamma{} +
t\vdelta{i})$. Since $r'(\tilde{\gamma}_i) = X^2/(1 + X^2\tilde{\gamma}_i)
- 1/\tilde{\gamma}_i\to 0$ for $X\to\infty$, $\tilde{\phi}'(t)$ is decreasing
at $t=0$ for large enough $X$, and $\phi$ is not convex at $\tvgamma{}$.

\subsection{Proof of \thref{sparse-infer}}
\label{app:sparse-infer}

In this section, we provides proofs related to \secref{map-estim}. We begin
with \thref{sparse-infer}.
For the first part, fix any $\vpi{}\succ\vzero$, any $i\in\srng{q}$, and any
$\Delta>0$. If $\vb{i}=\mxb{}^T\vdelta{i}\ne\vzero$, then using the determinant
identity previously employed in \appref{ldeta-convex}, we have
\[
\begin{split}
  \log|\tmxa{}(\vpi{}+\Delta\vdelta{i})| & = \log|\tmxa{}(\vpi{})| +
  \log\left| \Id + \Delta\tmxa{}(\vpi{})^{-1}\vb{i}\vb{i}^T \right| \\
  & = \log|\tmxa{}(\vpi{})| + \log(1 + \Delta\vb{i}^T\tmxa{}(\vpi{})^{-1}
  \vb{i}) > \log|\tmxa{}(\vpi{})|,
\end{split}
\]
since $\vb{i}^T\tmxa{}(\vpi{})^{-1}\vb{i}>0$ and $\log(1+x)>0$ for $x>0$.
Therefore, $\log|\tmxa{}(\vpi{})|$ is increasing in each component. Moreover,
we have that $\log|\tmxa{}(\vpi{}+\Delta\vdelta{i})|\to\infty$ ($\Delta\to
\infty$), since $\log(1+x)$ is unbounded above for $x\to\infty$. If
$\vpi{t}$ is a sequence with $\|\vpi{t}\|\to\infty$ and $\vpi{t}\succeq
\eps\vone$, there must be some
$i\in\srng{q}$ such that $(\vpi{t})_i\to\infty$. If
$\tvpi{t} := \eps\vone + ((\vpi{t})_i-\eps)\vdelta{i}\succeq\vpi{t}$, then
$\log|\tmxa{}(\vpi{t})|\ge \log|\tmxa{}(\tvpi{t})|\to \infty$ ($t\to\infty$).

For the second part, recall that
\[
\begin{split}
  & \phi(\vgamma{}) = \log|\mxa{}| + h(\vgamma{}) + \min_{\vu{}} \left\{
  R(\vu{},\vgamma{}) =
  \sigma^{-2}\|\vy{}-\mxx{}\vu{}\|^2 + \vs{}^T\mxgamma{}^{-1}\vs{}
  - 2\vb{}^T\vs{} \right\}, \\
  & -2\log P(\vu{}|\vy{}) = \min_{\vgamma{}\succeq\vzero}
  h(\vgamma{}) + R(\vu{},\vgamma{}) + C_2,\quad \vs{}=\mxb{}\vu{}
\end{split}
\]
for some constant $C_2$.
If $\vgamma{t}$ is a bounded sequence such that $(\vgamma{t})_i\to 0$
($t\to\infty$) for some $i\in\srng{q}$, then $\log|\mxa{}(\vgamma{t})| =
\log|\tmxa{}(\vgamma{t}^{-1})|\to\infty$. Suppose that $\phi(\vgamma{t})$
remains bounded above. Let $\vu{t} = \argmin_{\vu{}} R(\vu{},\vgamma{t})$.
Then, $\phi(\vgamma{t}) - \log|\mxa{}(\vgamma{t})| = h(\vgamma{t}) +
R(\vu{t},\vgamma{t})\to -\infty$, so that $-2\log P(\vu{t}|\vy{}) - C_2
\le h(\vgamma{t}) + R(\vu{t},\vgamma{t})\to -\infty$, in contradiction
to the boundedness of the log posterior. This concludes the proof.

Next, assume we run MAP estimation \eqp{map-estim} with even super-Gaussian
potentials $t_i(s_i)$, so that $|s_i|\mapsto -\log t_i(s_i) =
-\tilde{g}_i(s_i)$ is concave. As argued in \secref{map-estim}, any local
minimum point $\vs{*}=\mxb{}\vu{*}$ is exactly sparse. We show that the
corresponding $\vgamma{*}$ has the same sparsity pattern: $\gamma_{*,i}=0$
whenever $s_{*,i}=0$. Dropping the index,  since $\gamma_*\in
\argmin_{\gamma\ge 0} s_*^2/\gamma + h(\gamma)$, we have to show that
$h(\gamma) > h(0)$ for all
$\gamma>0$ (or, in terms of \appref{convex-inf}, that $\gamma_0=0$). Fix
$\gamma>0$, and recall that $h(\gamma) = \max_{s\ge 0}\{
\tilde{f}(s,\gamma) = -s^2/\gamma - 2\tilde{g}(s) \} \ge -2\tilde{g}(0) =
h(0)$. Now, $\tilde{f}(s,\gamma) = -2\tilde{g}(s) + O(s^2)$, $s\searrow 0$,
where $-2\tilde{g}(s)$ is concave, nondecreasing and not constant. Therefore,
$\lim_{s\searrow 0}\partial\tilde{f}/(\partial s)\in (0,\infty]$, and
$\tilde{f}(\tilde{s},\gamma) > \tilde{f}(0,\gamma)$ for some $\tilde{s}>0$,
so that $h(\gamma)\ge \tilde{f}(\tilde{s},\gamma) > h(0)$.

\subsection{Details for Bounding $\log|\mxa{}|$}
\label{app:bound-a}

In this section, we provide details concerning the $\log|\mxa{}|$ bounds
discussed in \secref{ldeta-bounds}. Recall that $\tmxa{}(\vpi{}) =
\sigma^{-2}\mxx{}^T\mxx{} + \mxb{}^T(\diag\vpi{})\mxb{}$ for
$\vpi{}\succ\vzero$. Define the extended-value extension
$g_1(\vpi{}) = \lim_{\vd{}\searrow\vpi{}}\log|\tmxa{}(\vd{})|$, $\vpi{}\succeq
\vzero$, $g_1(\vpi{}) = -\infty$ elsewhere (note that $\log|\tmxa{}(\vpi{})|$
is continuous). Since $g_1$ is lower semicontinuous, and concave for
$\vpi{}\succ\vzero$ (\thref{ldeta-convex}(\ref{th1-2})),
it is a closed proper concave function. Fenchel duality
\cite[Sect.~12]{Rockafellar:70} implies that
$g_1(\vpi{}) = \inf_{\vz{1}} \vz{1}^T\vpi{} - g_1^*(\vz{1})$, where
$g_1^*(\vz{1}) = \inf_{\vpi{}} \vz{1}^T\vpi{} - g_1(\vpi{})$ is closed concave
as well. As $g_1(\vpi{})$ is unbounded above as $\|\vpi{}\|\to\infty$
(\thref{sparse-infer}),
$\vz{1}^T\vpi{} - g_1(\vpi{})$ is unbounded below whenever $z_{1,i}\le 0$
for any $i$, and $g_1^*(\vz{1}) = -\infty$ in this case. Moreover, for
any $\vpi{}\succ\vzero$, the corresponding minimizer $\vz{1,*}$ is given
in \secref{ldeta-bounds}, so that $g_1(\vpi{}) = \min_{\vz{1}\succ\vzero}
\vz{1}^T\vpi{} - g_1^*(\vz{1})$.

Second, define the extended-value extension $g_2(\vgamma{}) =
\lim_{\vd{}\searrow\vgamma{}} \vone^T(\log\vd{}) + \log|\tmxa{}(\vd{})|$,
$\vgamma{}\succeq\vzero$, $g_2(\vgamma{}) = -\infty$ elsewhere (note that
$\vone^T(\log\vpi{}) + \log|\tmxa{}(\vpi{})|$ is continuous). Since $g_2$ is
lower semicontinuous, and concave for $\vgamma{}\succ\vzero$
(\thref{ldeta-convex}(\ref{th1-3})), it is a closed proper concave function.
Fenchel duality
\cite[Sect.~12]{Rockafellar:70} implies that $g_2(\vgamma{}) = \inf_{\vz{2}}
\vz{2}^T\vgamma{} - g_2^*(\vz{2})$, where $g_2^*(\vz{2}) = \inf_{\vgamma{}}
\vz{2}^T\vgamma{} - g_2(\vgamma{})$ is closed concave as well. Since
$\vz{2}^T\vgamma{} - g_2(\vgamma{})$ is unbounded below
whenever $z_{2,i}< 0$ for any $i$, we see that $g_2^*(\vz{2}) = -\infty$ in
this case. For any $\vgamma{}\succ\vzero$, the corresponding minimizer
$\vz{2,*}$ is given in \secref{ldeta-bounds}, so that $g_2(\vgamma{}) =
\min_{\vz{2}\succeq\vzero} \vz{2}^T\vgamma{} - g_2^*(\vz{2})$.

\subsection{Implicit Computation of $h_i$ and $h_i^*$}
\label{app:gen-inner}

Recall from \secref{algorithm} and \secref{inner-loop} that our algorithms
can be run whenever $h_i^*(s_i)$ and its derivatives can be evaluated.
For log-concave potentials, these evaluations can be done generically, even if
no closed form for $h_i(\gamma_i)$ is available. We focus on a single potential
$i$ and drop its index. As noted in \secref{algorithm}, if $z_2=z_3=0$, then
$h^*(s) = -g(z_1+s^2)$. With $p:=z_1+s^2$, we have that $\theta = -2 g'(p) s
- b$, $\rho = -4 g''(p) s^2 - 2 g'(p)$. With a view to \appref{norm-sites},
$\tilde{\theta} = -2 g'(p)$, $p=z_1+\|\vs{}\|^2$, and $\kappa =
2[g''(p)]^{1/2}$.

If $z_2\ne 0$ (and $t(s)$ log-concave), we have to employ scalar convex
minimization. We require
$h^*(s) = \frac{1}2\min_{\gamma} k(x,\gamma)$,
$k := (z_1+x)/\gamma + z_2\gamma - z_3\log\gamma + h(\gamma)$,
$x=s^2$, as well as $\theta=(h^*)'(s)$ and $\rho=(h^*)''(s)$.
Let $\gamma_* = \argmin k(x,\gamma)$. Assuming for now that $h$ and
its derivatives are available, $\gamma_*$ is found by univariate Newton
minimization,
where $\gamma^2 k_{\gamma} = -(z_1+x) - z_3\gamma + \gamma^2 (z_2+h'(\gamma))$,
$\gamma^3 k_{\gamma,\gamma} = 2(z_1+x) + \gamma z_3 + \gamma^3 h''(\gamma)$.
Now, $k_{\gamma}=0$ (always evaluated at $(x,\gamma_*)$), so that
$\theta=(h^*)'(s) =s/\gamma_*$. Moreover, $0 = (d/d s)k_{\gamma}
= k_{s,\gamma} + k_{\gamma,\gamma}\cdot(d\gamma_*)/(d s)$, so that
$\rho=(h^*)''(s) = \gamma_*^{-1}( 1 - s \gamma_*^{-1}(d\gamma_*)/(d s) ) =
\gamma_*^{-1}( 1 - 2 x/(\gamma_*^3 k_{\gamma,\gamma}(x,\gamma_*)) )$. With a
view to \appref{norm-sites}, $\tilde{\theta}=1/\gamma_*$ and $\kappa=
[2/(\gamma_*^4 k_{\gamma,\gamma}(x,\gamma_*))]^{1/2}$ (note that
$\tilde{\theta}\ge \rho$).

By Fenchel duality, $h(\gamma) = -\min_x l(x,\gamma)$,
$l := x/\gamma + 2 g(x)$, where $g(x)$ is strictly convex and decreasing.
We need methods to evaluate $g(x)$ and its first and second derivative
(note that $g''(x)>0$). The minimizer $x_*=x_*(\gamma)$ is found by convex
minimization once more, started from the last recently found $x_*$ for this
potential. Note that $x_*=0$ iff $\gamma\le \gamma_0 := -1/(2 g'(0))$ (where
$\gamma_0=0$ if $g'(x)\to -\infty$ as $x\to 0$), which has to be checked
up front. Given $x_*$, we have that $\gamma h(\gamma) = -x_* - 2\gamma g(x_*)$.
Since $l_x=0$ for $\gamma>\gamma_0$ (always evaluated at $(x_*,\gamma)$), then
$\gamma^2 h'(\gamma) = -\gamma^2 l_{\gamma} = x_*$ (this holds even if
$l_x>0$ and $x_*=0$). Moreover, if $x_*>0$ (for $\gamma>\gamma_0$), then
$(d/d\gamma) l_x(x_*,\gamma)=0$, so that $(d x_*)/(d\gamma) =
\gamma^{-2}/(2 g''(x_*))$, and $\gamma^3 h''(\gamma) = (2\gamma g''(x_*))^{-1}
- 2 x_*$. If $x_*=0$ and $l_x>0$, then $x_*(\tilde{\gamma})=0$ for
$\tilde{\gamma}$ close to $\gamma$, so that $h''(\gamma)=0$. A critical
case is $x_*=0$ and $l_x=0$, which happens for $\gamma = \gamma_0>0$:
$h''(\gamma)$ does not exist at this point in general. This is not a problem
for our code, since we employ a robust Newton/bisection search for $\gamma_*$.
If $\gamma>\gamma_*$, but is very close, note that $(d x_*)/(d\gamma)\approx
\xi_0/\gamma$ with $\xi_0 := -g'(0)/g''(0)$, therefore $x_*(\gamma)\approx
\int_{\gamma_0}^\gamma \xi_0/t\, d t = \xi_0(\log\gamma - \log\gamma_0)$.
We use $\gamma^2 h'(\gamma) = x_*\approx \xi_0(\log\gamma - \log\gamma_0)$ and
$\gamma^3 h''(\gamma)\approx \xi_0 - 2 x_*$ in this case.

\subsection{Details for Specific Potentials}
\label{app:impl-details}

Our algorithms are configured by the dual functions $h_i(\gamma_i)$ for each
non-Gaussian $t_i(s_i)$, and the inner loops require $h_i^*(s_i)$ and
its derivatives (see \eqp{inner-crit}, and recall that for each $i$, either
$z_{1,i}>0$ and $z_{3,i}=0$, or $z_{1,i}=0$ and $z_{2,i}>0, z_{3,i}=1$). In this
section, we show how these are computed for the potentials used in this paper.
We use the notation of \appref{gen-inner}, focus on a single potential $i$ and
drop its index.

\subsubsection*{Laplace Potentials}

These are $t(s) = \exp(-\tau|s|)$, $\tau>0$, so that $g(x) = \tau x^{1/2}$.
We have that $h(\gamma) =
h_{\cup}(\gamma)=\tau^2\gamma$, so that $k(x,\gamma) = (z_1+x)/\gamma +
(z_2+\tau^2)\gamma - z_3\log\gamma$. The stationary equation for $\gamma_*$ is
$(z_2+\tau^2)\gamma^2 - z_3\gamma - (z_1+x) = 0$. If $z_3=0$ (bounding type A),
this is just a special case of \appref{gen-inner}. With $p:=z_1+x$,
$q:=(z_2+\tau^2)^{1/2}$, we have that $\gamma_* = p^{1/2}/q$,
$h^*(s) = q p^{1/2}$, and $\theta = (h^*)'(s)= q p^{-1/2} s$, $\rho =
(h^*)''(s)= q z_1 p^{-3/2}$. With a view to \appref{norm-sites},
$\tilde{\theta} = q p^{-1/2}$ and $\kappa = [q p^{-3/2}]^{1/2}$.

If $z_3=1$ (bounding type B), note that $z_1=0$, $z_2>0$. Let
$q:=2(z_2+\tau^2)$, $p:=(1+2 q x)^{1/2}$. Then, $\gamma_*=(p+1)/q$, and
$k(x,\gamma_*) = p - \log(p+1) + \log q$ after some algebra, so that
$h^*(s) = \frac{1}2 ( p - \log(p+1) + \log q )$. With
$d p/(d s) = 2 q p^{-1} s$, we have $\theta = q s/(p+1)$,
$\rho = q/(p(p+1))$. With a view to \appref{norm-sites}, $\tilde{\theta} =
q/(p+1)$. Using $p^2-1 = 2 x q$, some algebra gives $\kappa =
(2/p)^{1/2} q/(p+1) = (2/p)^{1/2}\tilde{\theta}$.

\subsubsection*{Student's $t$ Potentials}

These are $t(s) = (1 + (\tau/\nu) x)^{-(\nu+1)/2}$, $\nu>0$, $\tau>0$.
If $\alpha:=\nu/\tau$, the critical point of \appref{convex-inf} is
$\gamma_0:=\alpha/(\nu+1)$, and $h(\gamma) = [\alpha/\gamma + (\nu+1)
\log\gamma + C ]\Ind{\gamma\ge\gamma_0}$ with $C:= -(\nu+1)(\log\gamma_0+1)$.
$h(\gamma)$ is not convex. We choose a decomposition such that
$h_{\cup}(\gamma)$ is convex and twice continuously differentiable, ensuring
that $h^*(s)$ is continuously differentiable, and the inner loop optimization
runs smoothly. Since $h(\gamma)$ does not have a second derivative at
$\gamma_0$, neither has $h_{\cap}(\gamma)$.
\[
\begin{split}
  h_{\cap}(\gamma) & = \left\{ \begin{array}{ll}
    (\nu+1-z_3)\log\gamma & |\; \gamma\ge\gamma_0 \\
    (2(\nu+1)-z_3)\log\gamma - a(\gamma-\gamma_0) - b & |\; \gamma<\gamma_0
  \end{array}\right., \\
  h_{\cup}(\gamma) & = \left\{ \begin{array}{ll}
    \alpha/\gamma + C & |\; \gamma\ge\gamma_0 \\
    -2(\nu+1)\log\gamma + a(\gamma-\gamma_0) + b & |\; \gamma<\gamma_0
  \end{array}\right.,
\end{split}
\]
where $b := (\nu+1)\log\gamma_0$, $a := (\nu+1)/\gamma_0$. Here, the
$-z_3\log\gamma$ term of $k(x,\gamma)$ is folded into $h_{\cap}(\gamma)$.

We follow \appref{gen-inner} in determining $h^*(s)$ and its derivatives,
but solve for $\gamma_*$ directly. Note that $z_2>0$ even if $z_3=0$ (bounding
type A), due to the Fenchel bound on $h_{\cap}(\gamma)$. We minimize
$k(x,\gamma)$ for $\gamma\ge\gamma_0$, $\gamma<\gamma_0$ respectively and pick
the minimum. For $\gamma\ge\gamma_0$: $k(x,\gamma) = (z_1+\alpha+x)/\gamma +
z_2\gamma + C$, whose minimum point $\gamma_{*,1} := [(z_1+\alpha+x)/z_2]^{1/2}$
is a candidate if $\gamma_{*,1}\ge\gamma_0$, with $k(x,\gamma_{*,1}) =
2[z_2(z_1+\alpha+x)]^{1/2}+C$. For $\gamma<\gamma_0$: $k(x,\gamma) =
(z_1+x)/\gamma + (z_2+a)\gamma - 2(\nu+1)\log\gamma + b-a\gamma_0$, with
minimum point $\gamma_{*,2} := [\nu+1+((\nu+1)^2 +
(z_2+a)(z_1+x))^{1/2}]/(z_2+a) <\gamma_0$. If $z_2\le a$, then
$\gamma_{*,2}\ge\gamma_0$ (not a candidate). This can be tested up front.
If $c:=(z_2+a)(z_1+x)$, $d:= ((\nu+1)^2+c)^{1/2}\ge \nu+1$, then
$k(x,\gamma_{*,2}) = c/(\nu+1+d) + d + (\nu+1)[2\log(z_2+a) - 2\log(\nu+1+d) +
\log\gamma_0] = 2 d + (\nu+1)[2\log(z_2+a) - 2\log(\nu+1+d) + \log\gamma_0
- 1]$. Now, $\theta$ and $\rho$ are computed as in
\appref{gen-inner} ($h(\gamma)$ there is $h_{\cup}(\gamma)$ here, and
$z_3=0$, since this is folded into $h_{\cap}$ here), where $\gamma_*^3
h_{\cup}''(\gamma_*) = 2\alpha$ for $\gamma_*\ge\gamma_0$, and $\gamma_*^3
h_{\cup}''(\gamma_*) = 2(\nu+1)\gamma_*$ for $\gamma_*<\gamma_0$.

\subsubsection*{Bernoulli Potentials}

These are $t(s) = (1 + e^{-y\tau s})^{-1} = e^{y\tau s/2}
(2\cosh v)^{-1}$, $v := (y\tau/2) x^{1/2} = (y\tau/2)|s|$. They are
not even, $b = y\tau/2$. While $h(\gamma)$ is not known analytically, we can
plug in these expressions into the generic setup of \appref{gen-inner}. Namely,
$g(x) = -\log(\cosh v) - \log 2$, so that $g'(x)
= -C (\tanh v)/v$, $g''(x) = (C/2) x^{-1} ( (\tanh v)/v + \tanh^2 v - 1)$,
$C := (y\tau/2)^2/2$. For $x$ close to zero, we use $\tanh v = v - v^3/3 +
2 v^5/15 + O(v^7)$ for these computations. Moreover, $\gamma_0 = 1/(2 C)$ and
$\xi_0 = 3/(2 C)$.

\subsection{Group Potentials}
\label{app:norm-sites}

An extension of our framework to group potentials $t_i(\|\vs{i}\|)$ is
described in \secref{group-pots}. Recall the details about the IRLS algorithm
from \secref{inner-loop}. For group potentials, the inner Hessian is not
diagonal anymore, but of similarly simple form developed here.
$h_i^*(s_i)$ becomes $h_i^*(\|\vs{i}\|)$, and $x_i=\|\vs{i}\|^2$. If
$\theta_i$, $\rho_i$ are as in \secref{inner-loop}, $d s_i\to d\|\vs{i}\|$, and
$\tilde{\theta}_i := \theta_i/\|\vs{i}\|$, we have that
$\nabla_{\vs{i}} h_i^* = \tilde{\theta}_i\vs{i}$, since
$\nabla_{\vs{i}}\|\vs{i}\| = \vs{i}/\|\vs{i}\|$. Therefore, the gradient
$\vth{}$ is given by $\vth{i}=\tilde{\theta}_i\vs{i}$. Moreover,
\[
  \nabla\nabla_{\vs{i}} h_i^* = \tilde{\theta}_i\Id -
  (\tilde{\theta}_i-\rho_i)\|\vs{i}\|^{-2}\vs{i}\vs{i}^T.
\]
For simplicity of notation, assume that all $\vs{i}$ have the same
dimensionality. From \appref{gen-inner}, we see that
$\tilde{\theta}_i\ge\rho_i$. Let
$\kappa_i := (\tilde{\theta}_i-\rho_i)^{1/2}/\|\vs{i}\|$, and
$\hat{\vs{}} := ((\diag \vkappa{})\otimes\Id)\vs{}$. The Hessian w.r.t.\
$\vs{}$ is
\[
  \mxh[s]{} = (\diag\tilde{\vth{}})\otimes \Id - \sum_i \vw{i}\vw{i}^T,\quad
  \vw{i} = (\vdelta{i}\vdelta{i}^T\otimes \Id) \hat{\vs{}}.
\]
If $\vw{}$ is given by $w_i = \hat{\vs{}}_i^T \vv{i}$, then
$\mxh[s]{}\vv{} = ((\diag\tilde{\vth{}})\otimes \Id)\vv{} -
((\diag\vw{})\otimes\Id)\hat{\vs{}}$.
The system matrix for the Newton direction is $\sigma^{-2}\mxx{}^T\mxx{} +
\mxb{}^T\mxh[s]{}\mxb{}$. For numerical reasons, $\tilde{\theta}_i$ and
$\kappa_i$ should be computed directly, rather than via $\theta_i$, $\rho_i$.

If $\vs{i}\in\R^2$, we can avoid the subtraction in computing $\mxh[s]{}\vv{}$
and gain numerical stability. Namely,
$\nabla\nabla_{\vs{i}} h_i^* = \rho_i\Id + \kappa_i^2\left( \|\vs{i}\|^2\Id -
\vs{i}\vs{i}^T\right)$.
Since $\|\vs{i}\|^2\Id - \vs{i}\vs{i}^T = \mxm{}\vs{i} (\mxm{}\vs{i})^T$,
$\mxm{} = \vdelta{2}\vdelta{1}^T - \vdelta{1}\vdelta{2}^T$, if we redefine
$\hat{\vs{}} := ((\diag \vkappa{})\otimes (\vdelta{2}\vdelta{1}^T -
\vdelta{1}\vdelta{2}^T))\vs{}$, then
\[
  \mxh[s]{}\vv{} = ((\diag\vrho{})\otimes \Id)\vv{} + ((\diag\vw{})\otimes
  \Id)\hat{\vs{}},\quad \vw{} = \left( \hat{\vs{}}_i^T \vv{i} \right).
\]

\end{appendix}



\end{document}

%% file: mymacros.tex
\newcommand{\field}[1]{\mathbb{#1}}

\newcommand{\R}{\field{R}}
\newcommand{\C}{\field{C}}

\newcommand{\vect}[1]{\boldsymbol{#1}} 
\newcommand{\mat}[1]{\boldsymbol{#1}} 
\newcommand{\tvect}[1]{\tilde{\boldsymbol{#1}}} 
\newcommand{\tmat}[1]{\tilde{\boldsymbol{#1}}} 
\newcommand{\hvect}[1]{\hat{\boldsymbol{#1}}} 
\newcommand{\hmat}[1]{\hat{\boldsymbol{#1}}} 
\newcommand{\vzero}{\vect{0}}
\newcommand{\vone}{\vect{1}}
\newcommand{\mxzero}{\mat{0}}

\newcommand{\dummystring}{QWERTYU}
\newcommand{\vci}[3][\dummystr]{\ifthenelse{\equal{#1}{\dummystring}}{\vect{#2}_{#3}}{\vect{#2}_{#3}^{(#1)}}}
\newcommand{\mx}[3][\dummystr]{\ifthenelse{\equal{#1}{\dummystring}}{\mat{#2}_{#3}}{\mat{#2}_{#3}^{(#1)}}}
\newcommand{\tvci}[3][\dummystr]{\ifthenelse{\equal{#1}{\dummystring}}{\tvect{#2}_{#3}}{\tvect{#2}_{#3}^{(#1)}}}
\newcommand{\tmx}[3][\dummystr]{\ifthenelse{\equal{#1}{\dummystring}}{\tmat{#2}_{#3}}{\tmat{#2}_{#3}^{(#1)}}}
\newcommand{\hvci}[3][\dummystr]{\ifthenelse{\equal{#1}{\dummystring}}{\hvect{#2}_{#3}}{\hvect{#2}_{#3}^{(#1)}}}
\newcommand{\hmx}[3][\dummystr]{\ifthenelse{\equal{#1}{\dummystring}}{\hmat{#2}_{#3}}{\hmat{#2}_{#3}^{(#1)}}}

\DeclareMathOperator{\trace}{tr}
\DeclareMathOperator{\diag}{diag}
\DeclareMathOperator*{\argmax}{arg max}
\DeclareMathOperator*{\argmin}{arg min}

\DeclareMathOperator{\rank}{rk}

\newcommand{\Ex}{\mathrm{E}}
\newcommand{\Var}{\mathrm{Var}}
\newcommand{\Cov}{\mathrm{Cov}}

\newcommand{\Ind}[1]{\mathrm{I}_{\{#1\}}}

\newcommand{\Ent}{\mathrm{H}}

\newcommand{\Id}{\mat{I}}

\newcommand{\eps}{\mathrm{\varepsilon}}

\newcommand{\figref}[1]{Figure~\ref{fig:#1}}

\newcommand{\secref}[1]{Section~\ref{sec:#1}}
\renewcommand{\eqref}[1]{Eq.~\ref{eq:#1}}
\newcommand{\eqp}[1]{(\ref{eq:#1})}
\newcommand{\algref}[1]{Algorithm~\ref{alg:#1}}
\newcommand{\appref}[1]{Appendix~\ref{app:#1}}
\newcommand{\thref}[1]{Theorem~\ref{th:#1}}


%% file: mymacros2.tex
\newcommand{\vb}[2][\dummystring]{\vci[#1]{b}{#2}}

\newcommand{\vd}[2][\dummystring]{\vci[#1]{d}{#2}}

\newcommand{\vf}[2][\dummystring]{\vci[#1]{f}{#2}}

\newcommand{\vk}[2][\dummystring]{\vci[#1]{k}{#2}}

\newcommand{\vq}[2][\dummystring]{\vci[#1]{q}{#2}}
\newcommand{\vr}[2][\dummystring]{\vci[#1]{r}{#2}}
\newcommand{\vs}[2][\dummystring]{\vci[#1]{s}{#2}}

\newcommand{\vu}[2][\dummystring]{\vci[#1]{u}{#2}}
\newcommand{\vv}[2][\dummystring]{\vci[#1]{v}{#2}}
\newcommand{\vw}[2][\dummystring]{\vci[#1]{w}{#2}}
\newcommand{\vx}[2][\dummystring]{\vci[#1]{x}{#2}}
\newcommand{\vy}[2][\dummystring]{\vci[#1]{y}{#2}}
\newcommand{\vz}[2][\dummystring]{\vci[#1]{z}{#2}}

\newcommand{\vgamma}[2][\dummystring]{\vci[#1]{\gamma}{#2}}
\newcommand{\vdelta}[2][\dummystring]{\vci[#1]{\delta}{#2}}
\newcommand{\veps}[2][\dummystring]{\vci[#1]{\eps}{#2}}
\newcommand{\vth}[2][\dummystring]{\vci[#1]{\theta}{#2}}
\newcommand{\vmu}[2][\dummystring]{\vci[#1]{\mu}{#2}}

\newcommand{\vpi}[2][\dummystring]{\vci[#1]{\pi}{#2}}

\newcommand{\vkappa}[2][\dummystring]{\vci[#1]{\kappa}{#2}}

\newcommand{\vrho}[2][\dummystring]{\vci[#1]{\rho}{#2}}

\newcommand{\tvb}[2][\dummystring]{\tvci[#1]{b}{#2}}

\newcommand{\tvz}[2][\dummystring]{\tvci[#1]{z}{#2}}

\newcommand{\tvgamma}[2][\dummystring]{\tvci[#1]{\gamma}{#2}}

\newcommand{\tvpi}[2][\dummystring]{\tvci[#1]{\pi}{#2}}

\newcommand{\hvz}[2][\dummystring]{\hvci[#1]{z}{#2}}

\newcommand{\mxa}[2][\dummystring]{\mx[#1]{A}{#2}}
\newcommand{\mxb}[2][\dummystring]{\mx[#1]{B}{#2}}

\newcommand{\mxe}[2][\dummystring]{\mx[#1]{E}{#2}}

\newcommand{\mxh}[2][\dummystring]{\mx[#1]{H}{#2}}

\newcommand{\mxl}[2][\dummystring]{\mx[#1]{L}{#2}}
\newcommand{\mxm}[2][\dummystring]{\mx[#1]{M}{#2}}
\newcommand{\mxn}[2][\dummystring]{\mx[#1]{N}{#2}}

\newcommand{\mxq}[2][\dummystring]{\mx[#1]{Q}{#2}}

\newcommand{\mxs}[2][\dummystring]{\mx[#1]{S}{#2}}
\newcommand{\mxt}[2][\dummystring]{\mx[#1]{T}{#2}}

\newcommand{\mxv}[2][\dummystring]{\mx[#1]{V}{#2}}

\newcommand{\mxx}[2][\dummystring]{\mx[#1]{X}{#2}}

\newcommand{\mxgamma}[2][\dummystring]{\mx[#1]{\Gamma}{#2}}

\newcommand{\mxsigma}[2][\dummystring]{\mx[#1]{\Sigma}{#2}}

\newcommand{\tmxa}[2][\dummystring]{\tmx[#1]{A}{#2}}

\newcommand{\rng}[2][1]{{#1},\dots,{#2}}

\newcommand{\srng}[2][1]{\{{#1},\dots,{#2}\}}